\documentclass[twocolumn]{article}

\usepackage{hyperref}
\usepackage{jfrExamplee}
\usepackage{graphicx}
\usepackage{abstract}
\usepackage{setspace}
\doublespace
\usepackage{gensymb}
\usepackage{amsmath}
\usepackage{tabularx}
\usepackage{subcaption}
\usepackage{float}
\usepackage[table]{xcolor}% http://ctan.org/pkg/xcolor
\usepackage{diagbox}
\usepackage{tikz}
\usepackage{wrapfig}
\usepackage{multicol}
\usepackage{float}
\newcommand{\hyperfootnote}[1][]{\def\ArgI\hyperfootnoteRelay}
\def\checkmark{\tikz\fill[scale=0.4](0,.35) -- (.25,0) -- (1,.7) -- (.25,.15) -- cycle;}
\title{Fusion of neural networks, for LIDAR-based evidential road mapping}%road detection in LIDAR scans, and evidential road mapping}

\author{
\'{E}douard Capellier\thanks{Edouard CAPELLIER is also a research engineer in computer vision at Renault S.A.S, France} \\
HeuDiaSyc\\
Universit\'{e} de technologie de Compi\`{e}gne\\
CNRS, Heudiasyc, Alliance Sorbonne Université \\ CS 60 319, 60 203 Compiegne Cedex, France\\
\texttt{edouard.capellier@hds.utc.fr} \\
\And
Franck Davoine \\
HeuDiaSyc\\
Universit\'{e} de technologie de Compi\`{e}gne\\
CNRS, Heudiasyc, Alliance Sorbonne Université \\ CS 60 319, 60 203 Compiegne Cedex, France\\
\texttt{franck.davoine@hds.utc.fr} \\
\And
V\'{e}ronique Cherfaoui\\
HeuDiaSyc\\
Universit\'{e} de technologie de Compi\`{e}gne\\
CNRS, Heudiasyc, Alliance Sorbonne Université \\ CS 60 319, 60 203 Compiegne Cedex, France\\
\texttt{veronique.cherfaoui@hds.utc.fr} \\
\And
You Li \\
Renault S.A.S \\
Research department (DEA-IR)\\
1 av. du Golf \\
78288 Guyancourt, France\\
\texttt{you.li@renault.com} \\
}
\begin{document}
\singlespacing
\twocolumn[
\begin{@twocolumnfalse}
\maketitle
\raggedbottom
\begin{abstract}
\smallskip
LIDAR sensors are usually used to provide autonomous vehicles with 3D representations of their environment. In ideal conditions, geometrical models could detect the road in LIDAR scans, at the cost of a manual tuning of numerical constraints, and a lack of flexibility. We instead propose an evidential pipeline, to accumulate road detection results obtained from neural networks. First, we introduce RoadSeg, a new convolutional architecture that is optimized for road detection in LIDAR scans. RoadSeg is used to classify individual LIDAR points as either belonging to the road, or not. Yet, such point-level classification results need to be converted into a dense representation, that can be used by an autonomous vehicle. We thus secondly present an evidential road mapping algorithm, that fuses consecutive road detection results. We benefitted from a reinterpretation of logistic classifiers, which can be seen as generating a collection of simple evidential mass functions. An evidential grid map that depicts the road can then be obtained, by projecting the classification results from RoadSeg into grid cells, and by handling moving objects via conflict analysis. The system was trained and evaluated on real-life data. A python implementation maintains a 10 Hz framerate. Since road labels were needed for training, a soft labelling procedure, relying lane-level HD maps, was used to generate coarse training and validation sets. An additional test set was manually labelled for evaluation purposes. So as to reach satisfactory results, the system fuses road detection results obtained from three variants of RoadSeg, processing different LIDAR features. 
\end{abstract}
\end{@twocolumnfalse}
]
\saythanks 
\section{Introduction}
\paragraph{}
Grid mapping algorithms are traditionally deployed, within robotic systems, to infer the traversability of discretized areas of the environment. In particular, evidential grid mapping algorithms are commonly used to fuse sensor inputs over time~\cite{lacroix2002autonomous,rauskolb2008caroline,tanzmeister2017evidential,nuss2018random}. The evidential framework better represents uncertainties, and the fact of not knowing, than regular Bayesian frameworks~\cite{mullane2006evidential}. Indeed, the evidential framework allows for an explicit classification of unobserved, or ambiguous, areas as being in an \textit{Unknown} state.
\paragraph{}
Specifically, LIDAR-only evidential grid mapping systems tend to rely on a first ground detection step, strong geometrical assumptions, and manually tuned thresholds, to generate evidential mass functions at the cell level~\cite{yu2014evidential,capellier2018evidential,wirges2018evidential}. However, such systems might prove to have a limited applicability, for autonomous vehicles that are intended to evolve in urban or peri-urban areas. The first reason is that autonomous ground vehicles can only, typically, drive on roads. Yet, the road does not necessarily cover the whole ground, especially in urban areas. Moreover, the need for manual finetuning, and the use of geometrical assumptions, lead to grid mapping systems that lack of flexibility, and might fail when used in uncontrolled environments. For instance, the use of ray tracing, to densify evidential grid maps as in~\cite{rauskolb2008caroline}, supposes the absence of negative obstacles in unobserved areas, which cannot be guaranteed in practice when only using LIDAR sensors~\cite{shang2016lidar}.  We consider that a representation of the drivable area should only be generated by fusing actual observations, without any upsampling nor downsampling. Similarly, a grid mapping system relying on the flat-ground assumption, and altitude thresholds to detect obstacles~\cite{yu2014evidential}, might lead to false detections when encountering slopes or speed bumps.
\begin{figure*}[t]
\includegraphics[width=\textwidth]{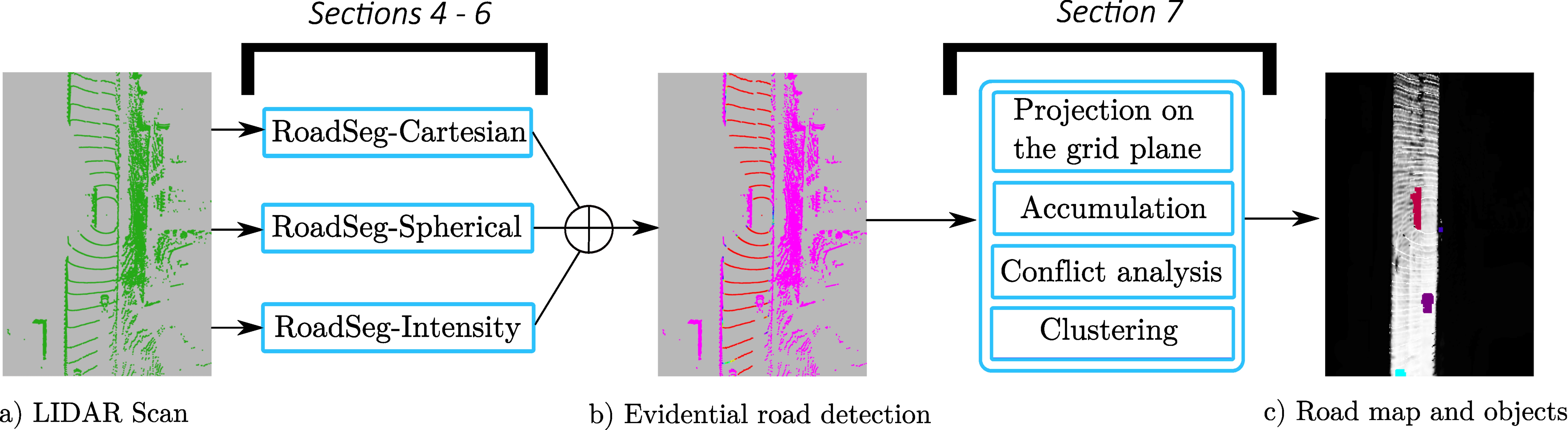}
\caption{Overview of the proposed system}
\label{overview}
\end{figure*}
\paragraph{}
To tackle those limitations, we propose to classify individual LIDAR points as either belonging to the \textit{road}, or to an \textit{obstacle}.  The use of machine-learning, for classification, would have the benefit of removing the need for manually tuned geometrical assumptions and thresholds, during the inference. The use of dense LIDAR scans, for road detection, has been a common practice in aerial data analysis for several years~\cite{clode2005improving}, and several approaches relying on machine learning techniques have recently been proposed to address automotive use cases~\cite{caltagirone2017fast,lyu2018chipnet}. However, those systems never generate a point-level classification, and always output an upsampled road area. Indeed, those approaches are designed to process data from the KITTI road dataset~\cite{Fritsch2013ITSC}, and are evaluated with regards to a dense ground-truth in the image domain. As such, the fact that LIDAR scans are sparse is not considered when evaluating those systems. Recent machine learning architectures, such as SqueezeSeg~\cite{wu2018squeezeseg} and PointNet~\cite{qi2017pointnet}, have been proposed to process raw point-clouds. In particular, SqueezeSeg relies on a very efficient convolutional architecture that processes LIDAR scans, which are organized as dense range images by simply using a spherical projection. This representation actually corresponds to the raw outputs of most state-of-the-art LIDAR sensors, that only measure ranges at fixed angular positions. We propose to adapt SqueezeSegV2~\cite{wu2019squeezesegv2}, and make it able to classify each LIDAR point as \textit{road} or \textit{obstacle}. Those classification results an then be accumulated into a 2D road map, thanks to an evidential grid mapping algorithm. Figure~\ref{overview} depicts the whole system.
\paragraph{}
To generate evidential mass functions from the classification results, we propose to reinterpret the output of neural networks as a collection of simple evidential mass functions, as proposed in~\cite{denoeux2019logistic}. The system was trained on a coarse training set, which was generated by automatically labelling LIDAR scans from re-existing lane-level HD Maps, thanks to a simple localization error model. An additional test set was also manually labelled, in order to reliably verify the classification performances. The best classification performances are achieved by fusing three distinct neural networks, that process different sets of LIDAR features: the Cartesian coordinates, the spherical coordinates, and the returned intensity and elevation angles. The classification results are then accumulated on the XY-plane, in order to generate an evidential grid that depicts the road surface. Moving objects are detected by analyzing of the evidential conflict during the fusion, and a grid-level clustering pipeline. Those objects are then excluded from the final road surface estimation.
\paragraph{}
Our main contributions are:
\begin{itemize}
\item RoadSeg, a refined version of SqueezeSegV2 for road detection in LIDAR scans, and optimized for the generation of evidential mass functions
\item An automatic labelling procedure of the road in LIDAR scans, from HD Maps
\item A road mapping and object detection algorithm relying on road detections results, obtained from a fusion of independent RoadSeg networks
\end{itemize}
\paragraph{}
The remainder of the paper is organized as follows. Section~\ref{relw} reviews some related works on road detection in LIDAR scans, machine learning architectures for LIDAR scan processing, and evidential grid mapping. Section~\ref{glrtoevi} describes how evidential mass functions can be generated from a neural network. Section~\ref{roasegsection} describes RoadSeg, our refined version of SqueezeSegV2 for road detection, and optimized for the generation of evidential mass functions. Section~\ref{softlabels} presents our training, validation and test sets, alongside our Map-based automatic labelling pipeline. In section~\ref{evalkittirenault}, we evaluate the performances of RoadSeg, and a fusion of RoadSeg networks that process independent inputs, both on the KITTI dataset, and our manually labelled test. Section~\ref{gridmapalgo} finally presents a road mapping and moving object detection algorithm, that relies on evidential mass functions generated from several RoadSeg networks. A python implementation of this algorithm processes LIDAR scans at a 10Hz rate.

\section{Related Works}
\label{relw}

\subsection{Evidential grid mapping from LIDAR scans}

\paragraph{}
Evidential occupancy grids associate evidential mass functions to individual grid cells. Each mass value represents evidence towards the fact that the corresponding cell is \textit{Free}, \textit{Occupied}, or \textit{Unknown}, and their sum is equal to 1. The first evidential occupancy grid mapping system that relied on LIDAR sensors was proposed in~\cite{yang2005evidential}. Evidential mass functions were constructed at the cell level from several stationary-beam LIDAR sensors, via ray tracing and ad-hoc false alarm and missed detection rates. The use of the evidential framework was shown to better represent uncertainty and the fact of not knowing, when compared to a traditional Bayesian fusion scheme. The Caroline vehicle already used evidential occupancy grid maps during the 2007 DARPA Urban Challenge~\cite{rauskolb2008caroline}, but the reasoning was done globally from both LIDAR scans and point-clouds generated by stereo-vision, without considering the specificity of LIDAR sensors. 
\paragraph{}
It was proposed, in~\cite{moras2011moving}, to extend the work in \cite{yang2005evidential} to four-layer LIDAR scanners. Based on ~\cite{smets2000data}, a discount factor was applied to an evidential polar grid, before fusing it with new sensor observations. Evidential mass functions were again generated from ray tracing, and ad-hoc false alarm and missed detection rates. This model was generalized, in~\cite{yu2014evidential}, to a more complex 360\degree Velodyne HDL-64 LIDAR scanner. A first ground-detection step, based on a simple thresholding, is used to classify LIDAR points as either belonging to an obstacle or to the drivable area. Then, evidential mass functions were created at the cell level, from the classification results and from, again, ad-hoc false-alarm and missed-detection rates.
\paragraph{}
More recent works aim at tackling some intrinsic limitations of LIDAR-based evidential occupancy grid mapping. It was proposed, in~\cite{nuss2018random}, to couple an evidential model with particle-filtering, in order to estimate the speed of each grid cell, and detect moving objects from a four-layer LIDAR scan. This comes at the cost of various hyper-parameters to manually tune, and a computational complexity, as virtually any occupied cell could be associated with a set of particles. Another recent work aims at predicting dense evidential occupancy grid maps from individual LIDAR scans~\cite{wirges2018evidential}. Consecutive LIDAR scans are registered, and dense grids are obtained from the resulting pointcloud thanks to a ground removal step, and manually defined false-alarm and missed detection rates. A convolutional neural network is then trained to recreate the dense evidential grids from a feature grid generated from only one of the original scans.
\paragraph{}
All those approaches rely on several strong assumptions, like the absence of negative obstacles that justifies the use of ray tracing, or the fact that the ground is flat. As such, they might lack flexibility, when used in uncontrolled environments. We thus consider that in complex and uncontrolled environments, evidential LIDAR grids should only be built from raw observations. The use of ad-hoc parameters should also be limited, so as to create general models. Moreover, autonomous urban vehicles are expected to drive on roads. Therefore, current evidential occupancy grid mapping algorithms cannot be used alone, in autonomous driving scenarios, as unoccupied grid cells are not necessarily drivable. An explicit road detection step is thus needed, when generating evidential grid maps for autonomous driving.

\subsection{Road detection in LIDAR scans}

\paragraph{}
Road detection from LIDAR scans can be addressed by focusing on the detection of road borders. One of the first road detection approaches that only relies on LIDAR data was proposed in the context of the DARPA Urban Challenge 2007~\cite{zhang2010lidar}. Elevation information was obtained from a single-layer LIDAR mounted on a prototype vehicle, and oriented towards the ground. The road was then considered as the lowest smooth surface within the LIDAR scan, and a comparison with detected road edges was used to confirm the results. A similar sensor setup was used in~\cite{han2012enhanced}, to detect breakpoints among LIDAR points, so that smooth road segments and obstacles could be jointly detected.  More recently,~\cite{he2018automatic} coupled a ground detection and a probabilistic fusion framework, to accumulate and vectorize road edge candidates. In~\cite{jung2018real}, road markings are detected among LIDAR points, from the reflectance information, and the geometry of the lane in inferred from the detected markings. Simply detecting road borders from LIDAR scans is nevertheless limited, as the provided information is extremely coarse by nature. Those approaches also heavily rely on assumptions about the sensor set-up and geometry of the scene, such as the absence of negative obstacles, the flatness of the ground, and the presence of road markings. In complex urban scenarios, those assumptions are not necessarily verified, which validates the need for algorithms that explicitely detect the road, instead of simple road borders.

\paragraph{}
State-of-the-art road detection algorithms usually rely on the fusion of LIDAR scans and RGB images~\cite{xiao2018hybrid,chen2019progressive}. Yet, the release of the KITTI dataset allowed for the emergence of LIDAR-only road detection algorithms. Most of those approaches actually readapt image processing techniques, to detect a road area instead of road edges. In ~\cite{fernandes2014road}, LIDAR points are projected and upsampled into a 2D image plane. The road area is then detected in this plane, via a histogram similarity measure. Another proposition was to project LIDAR points into a 2D sparse feature grid corresponding to a bird's eye view, and to train a convolutional neural network to predict a dense road region from this representation~\cite{caltagirone2017fast}. Another proposal was to train a neural network on range images, obtained from the spherical projection of LIDAR scans, and to fit a polygon on the predicted road points to obtain a dense prediction~\cite{lyu2018chipnet}. Yet, some LIDAR points are lost on the projection process, as each pixel of the range image could represent several LIDAR points.

\paragraph{}
Although those approaches are currently the best performing LIDAR-only road-detection approaches on the KITTI dataset, they share the same limitation: they aim at predicting a dense road area although LIDAR scans are sparse, since they are evaluated with regards to dense, image-level road labels. All those approaches then predict the presence of road on locations where no actual LIDAR measurements are actually available, which is an incorrect behavior for a LIDAR-only road detection algorithm. Indeed, gaps or small obstacles could be present but unobserved, because of the sparsity of LIDAR scans. Moreover, due to the limitations of the KITTI dataset, which only have road labels for the front camera view, those systems are not designed to detect the road at 360\degree, and might require to be significantly modified to detect the road on full LIDAR scans, without any guarantee of success. However, machine learning approaches~\cite{caltagirone2017fast,lyu2018chipnet} have the advantage of being able to detect roads, after their training, without any explicit geometrical model, nor manually tuned thresholds. A machine-learning algorithm that could process raw LIDAR scans, and classify each individual LIDAR point as either belonging to the road or to an obstacle, would then be valuable when generating evidential grid maps in the context of autonomous driving.

\subsection{Machine learning on raw LIDAR scans}

\paragraph{}
The Machine Learning algorithms that can process raw LIDAR scans, and output per-point results, can be split into two main categories. The first category of approaches consider LIDAR scans as unorganized point-clouds, and usually rely on PointNet-like architectures~\cite{qi2017pointnet}. The seconc category of approaches are strongly inspired by SqueezeSeg~\cite{wu2018squeezeseg}, and consider that LIDAR scans are organized point-clouds, which can be processed as range images obtained by spherical projections, similarly to what is done in~\cite{lyu2018chipnet}.

\paragraph{}
PointNet applies a common multi-layer perceptron to the features of each point of a point-cloud. A global max-pooling operation is then used to extract a cloud-level feature. Such an architecture was mathematically demonstrated to be able to approximate any set function. Further layers can then be trained to perform object classification or semantic segmentation. However, PointNet expects normalized, and relatively constrained inputs, which involves several pre-processing steps. Several PointNet-like architectures, optimized  for the semantic segmentation of LIDAR scans, are proposed in ~\cite{engelmann2017exploring}. They require an initial pre-processing step, as they process independent subsets of the overall scan, and none of them reach fully satisfactory results. PointNet was also tested for object detection in LIDAR scans. Voxelnet~\cite{zhou2018voxelnet} applies PointNet network on individual voxel partitions of a LIDAR scan. It then uses additional convolutions, and a region proposal network, to perform bounding-box regression. PointNet can also be used to perform bounding-box regression on LIDAR frustrums, corresponding to a pre-computed 2D bounding-box obtained from an RGB image~\cite{qi2018frustum}. Given the limitations observed in~\cite{engelmann2017exploring} for semantic segmentation, and the need for pre-processing, we chose to rely on another type of approach to perform road detection. Indeed, as the results are intended to be fused in evidential grid maps in real-time, a solution that could directly process complete LIDAR scans would be more relevant.

\paragraph{}
SqueezeSeg~\cite{wu2018squeezeseg} was introduced as a refined version of SqueezeNet~\cite{iandola2016squeezenet}, a highly efficient convolutional neural network reaching AlexNet-level accuracy, with a limited number of parameters and a low memory footprint. SqueezeNet heavily on Fire layers, that require less computations and use fewer parameters than traditional convolution layers. SqueezeSeg adapts SqueezeNet for semantic segmentation of LIDAR scans, by processing organized LIDAR scans. Spherical projection can indeed be used to generate dense range images from LIDAR scans. SqueezeSeg was initially designed to perform semantic segmentation. The labels were obtained from the KITTI dataset, and ground-truth bounding-boxes: the LIDAR points that were falling into those boxes were classified according the class of the related object. Again, as only the front camera view is labelled in KITTI, the labels do not cover complete LIDAR scans. A conditional random field (CRF), reinterpreted as a recurrent neural network, was also trained alongside SqueezeSeg, to further improve the segmentation results. Recently, SqueezeSegV2 introduced a context aggregation module, to cope with the fact that LIDAR scans usually include missing points, due to sensor noise. The input range image used in SqueezeSegV2 also includes an additional channel that indicates whether a valid sensor reading was available, at each angular position. Finally, SqueezeSegV2 extensively uses batch-normalization, contrary to SqueezeSeg. Both SqueezeSeg and SqueezeSegV2 are highly efficient networks, as their original implementation could process up to 100 frames per second. More recently, RangeNet++~\cite{milioto2019rangenet++} reused a similar approach, to perform semantic segmentation on the SemanticKitti dataset~\cite{behley2019semantickitti}. As the SemanticKitti dataset is a finally labelled dataset, with numerous labelled classes, RangeNet++ has more parameters than SqueezeSegV2. It is also significantly slower, so that its original implementation does not match the 10Hz original frame rate of the Velodyne HDL64, which was used to collect the raw data of the SemanticKitti dataset. 
\raggedbottom
\paragraph{}
As SqueezeSegV2 is highly efficient, and processes full LIDAR scans that are easy to organize, refining it for road detection seems to be a natural option. Such a network, thanks to its fast ineference time, can also be coupled with an evidential road mapping algorithm, to convert the segmentation results into a representation that can be directly used by an autonomous vehicle.  A straightforward way to do it would be to create a dataset with evidential labels, train the network on them, and fuse the detection results over time. However, such labels are hard to obtain, due to the presence of an \textit{unknown} class. We instead propose to extend the model described in~\cite{denoeux2019logistic}, which offers a way to generate evidential mass functions from pre-trained binary generalized logistic regression (GLR) classifiers. 

\section{Generation of evidential mass functions from a binary GLR classifier}
\label{glrtoevi}

\subsection{Definition of a binary GLR classifier for road detection in LIDAR scans}
\paragraph{}
Let $\Omega = \{R, \neg R\}$ be a binary frame of discernment, with $R$ corresponding to the fact that a given LIDAR point belongs to the road, and $\neg R$ to the fact that it does not. Following the evidential framework, the reasoning in done on the power set $2^\Omega=\{\emptyset, R, \neg R, \Omega\}$. $\Omega$ indicates ignorance on the class of the point. $\emptyset$ indicates that the point does not fit the frame of discernment, which is not possible in our case, since each LIDAR point either belongs, or does not belong, to the road. 
\paragraph{}
Let the binary classification problem with $X=(x_1,...,x_d)$, a d-dimensional input vector representing a LIDAR point, and $Y \in \Omega$ a class variable. Let $p_R(x)$ be the probability that $Y=R$ given that $X = x$. Then $1-p_R(x)$ is the corresponding probability that $Y=\neg R$. Let $w$ be the output of a binary logistic regression classifier, trained to predict the probability that a LIDAR point belongs to the road. Then, $p_R(x)$ is such that:
\begin{equation}
p_R(x) = S(\sum_{j=1}^d\beta_j\phi_j(x)+\beta_0)
\label{GLRC}
\end{equation}
where $S$ is the sigmoid function, and the $\beta$ parameters usually learnt alongside those of the $\phi_i$ mappings. Equation~\ref{GLRC} exactly corresponds to the behavior of a deep neural network trained as a binary GLR classifier, with $x$ being its input.

\subsection{Reinterpratation the GLR classifier in the evidential perspective}
\paragraph{}
Let $\oplus$ be Dempster's rule of combination, that can be used to fuse two independent evidential mass functions $m_1, m_2$ that follow the same frame of discernment. The resulting mass function $m_{1,2} = m_1 \oplus m_2$ is such that: 
\begin{subequations}
\begin{flalign}
&m_{1,2}(\emptyset)=0\\
&K = \sum_{B \cap C = \emptyset}m_1(B)m_2(C)\\
&\forall A \in 2^\Omega\setminus\{\emptyset\}: m_{1,2}(A) = \frac{\sum_{D \cap E = A \neq \emptyset}m_1(D)m_2(E)}{1-K}
\end{flalign}
\label{dsoperator}
\end{subequations}

A \textit{simple} evidential mass function $m$ on $\Omega$ is such that $\exists \theta \subset \Omega, m(\theta)=s, m(\Omega) = 1-s$. Let $w = -ln(1-s)$ be the \textit{weight of evidence} associated to $m$; $m$ can then be represented as $\{\theta_i\}^{w}$. The sigmoid function is strictly increasing. Then, in Equation~\ref{GLRC}, the larger the value generated by the classifier is, the larger $p_R(x)$ is and the smaller $1-p_R(x)$ is. Equation~\ref{GLRC} can be rewritten as follows:
\begin{equation}
p_R(x) = S(\sum_{j=1}^d (\beta_j\phi_j(x)+\alpha_j)) = S(\sum_{j=1}^dw_j)
\label{wjold}
\end{equation} 
with
\begin{equation}
\sum_{j=1}^d \alpha_j = \beta_0
\label{alpha}
\end{equation} 
\paragraph{}
Each $w_j$ can then be seen as piece of evidence towards $R$ or $\neg R$, depending on its sign. Let us assume that the $w_j$ values are weights of evidence of simple mass functions, denoted by $m_j$. Let $w_j^+ = max(0,w_j)$ be the positive part of $w_j$, and let $w_j^- = max(0,-w_j)$ be its negative part. Whatever the sign of $w_j$, the corresponding evidential mass function $m_j$ can be written as:
\begin{equation}
m_j = \{R\}^{w_j^+}\oplus\{\neg R\}^{w_j^-}
\label{naivemj}
\end{equation}
\paragraph{}
Under the assumption that all the $m_j$ mass functions are independent, the Dempster-Shafer operator can be used to fuse them together.The resulting mass function obtained from the output of the binary logistic regression classifier, noted $m_{LR}$ is as follows:
\begin{subequations}
\begin{flalign}
 &m_{LR}(\{R\})=\frac{\left[1-\exp(-w^+)\right](\exp(-w^-))}{1-K} \\
 &m_{LR}(\{\neg R\})=\frac{\left[1-\exp(-w^-)\right](\exp(-w^+))}{1-K} \\
 &m_{LR}(\Omega)=\frac{\exp(-\sum_{j=1}^d|w_j|)}{1-K} \\
\textnormal{with}\\
 &K = \left[1-\exp(-w^+)\right]\left[1-exp(-w^-)\right]
\end{flalign}
\label{mlrfull}
\end{subequations}
\paragraph{}
By applying the plausibility transformation described in~\cite{cobb2006plausibility} to the evidential mass function in Equation~\ref{mlrfull}, the expression of $p_R(x)$ can be reconstructed. Indeed:
\begin{equation}
p_R(x) = \frac{m_{LR}(\{R\})+m_{LR}(\Omega)}{m_{LR}(\{R\})+m_{LR}(\{\neg R\})+2m_{LR}(\Omega)}
\label{plaustransfo}
\end{equation}
This means that any binary GLR classifier can be seen as a fusion of simple mass functions, that can be derived from the parameters of the final linear layer of the classifier. The previous assumptions are thus justified. However, the $\alpha_j$ values introduced in Equation~\ref{alpha} have to be estimated. 

\subsection{Optimization of the parameters of the evidential mass functions}
\paragraph{}
Let $\alpha = (\alpha_1,...\alpha_d)$. A cautious approach would be to select them so as to maximize the mass values on $\Omega$. A solution to the resulting minimization problem over the training set was proposed in~\cite{denoeux2019logistic}. Alongside this approach, we propose to explore two other possibilities.

\subsubsection{Optimization over the training set as a post-processing}
\paragraph{}
The original approach proposed in~\cite{denoeux2019logistic} was to select the $\alpha$ vector that maximizes the sum of the $m_{LR}(\Omega)$ mass values over the training set, so as to get the most uncertain evidential mass functions. This leads to the following minimization problem:
\begin{equation}
min f(\alpha) = \sum\limits_{i=1}^n\sum\limits_{j=1}^d(\beta_j\phi_j(x_i)+\alpha_j)^2
\label{min}
\end{equation}
with $\left\lbrace(x_i,y_i)\right\rbrace_{i=1}^n$ being the training dataset. Let $\overline{\phi_k}$ be the mean of $\phi_k(x_i)$ on the training set. The optimal value for $\alpha_j$ is then:
\begin{equation}
\alpha_j = \frac{\beta_0}{d} + \frac{1}{d}\sum\limits_{q=1}^d\beta_q\overline{\phi_q} - \beta_j\overline{\phi_j}
\label{best_a}
\end{equation}
This solution thus relies on a post-processing step, and is dependent on the parameters that were initially learnt from the network. Typically, if either $d$, the $\beta_j$ values or the $\phi_j(x_i)$ values in Equation~\ref{wjold} are very large, the resulting optimal $\alpha_j$ values might remain negligible with regards to the corresponding $w_j$ values. Given that binary GLR classifiers are usually trained from binary labels, and thus to saturate the Sigmoid function by predicting probabilities close to 0 or 1, this case is likely to happen, especially for deep GLR classifiers for which the $d$ value can be extremely high.

\subsection{Optimization over abundant unlabelled data as a post-processing}
\paragraph{}
There is actually no practical reason to optimize the $\alpha$ vector over the training set, appart from the lack of data. Indeed, the predicted probabilities are not dependent on the $\alpha$ vector, since the sum of its elements is always equal to $\beta_0$. Moreover, the minimization problem in Equation~\ref{min} does not depend on the original labels. As we intend to use this model for road detection in LIDAR scans, abundant and various unlabelled data can be acquired easily by simply recording LIDAR scans. Then, the minimization problem in Equation~\ref{min} can be optimized on these abundant LIDAR scans. 

\subsubsection{Optimization over the training set during the training}
\paragraph{}
It was observed in~\cite{denoeux2019logistic} that if the $\phi_k(x_i)$ features in Equation~\ref{min} were centered around their mean, the minimization problem might be partially solved by applying L2-regularization during the training. Indeed, Equation~\ref{min} then becomes:
\begin{equation}
min f(\alpha) = \sum_{j=1}^d\beta_j^2(\sum_{i=1}^n\phi_j(x_i)^2)+\frac{n}{d}\beta_0^2
\end{equation}
However, nothing would then prevent the parameters of the $\phi_k$ functions to overcompensate for the L2-regularization during the training, especially if non-linearities are used, and if the $\phi_k$ mappings are modelled by a very deep network. This means that the total sum might actually not be minimized. Moreover, centering the $\phi_k(x_i)$ is not a straightforward operation. If the classifier only used linear operations, this could be easily done by centering the training set. However, performant classifiers are usually non-linear. The centering thus has to be done as an internal operation of the network.
\paragraph{}
We previously observed that using Instance-Normalization~\cite{ulyanov2016instance} in the final layer of the classifier would be an easy way to solve those problems~\cite{capellier2019evidential}. Let $\upsilon(x) = (\upsilon_1(x),...,\upsilon_d(x))$ be the mapping modelled by all the consecutive layers of the classifier but the last one ; let $\overline{\upsilon_j}$ be the mean value of the $\upsilon_j$ function on the training set, and $\sigma(\upsilon_j)^2$ its corresponding variance. Then, if it is assumed that the final layer of the network relies on  an Instance-Normalization and a feature-wise sum, Equation~\ref{min} becomes:

\begin{equation}
min f(\alpha) = \sum\limits_{i=1}^n\sum\limits_{j=1}^d((\beta_j\frac{\upsilon_j(x_c) - \overline{\upsilon_j}}{\sqrt{\sigma(\upsilon_j)^2+\epsilon}})+\sum\limits_{j=1}^d\alpha_j)^2
\label{minnorm}
\end{equation}
The $\alpha$ vector would actually correspond to the biases of the Instance-Normalization layer. After development, if $\epsilon$ is assumed to be neglectible, the following expression is obtained:

\begin{equation}
minf(\alpha) = n\sum_{j=1}^d\beta_j^2 + n\sum_{j=1}^d\alpha_j^2
\label{wdres}
\end{equation}
Again, this expression would be minimized during the training by simply applying L2-regularization. Instance-Normalization is thus an interesting way of centering the penultimate features of the network, since it inhibates the influence of the intermediate layers of the network. 
\paragraph{}
It should be noted that the use of Instance-Normalization is not incompatible with a post-processing step to further optimize the $\alpha$ vector over either the training set, or abundant unlabelled data. The best strategy will thus have to be defined with regards to the road detection and grid mapping tasks. 
%%%%%%%%%%%%%%%%%%
\section{RoadSeg: a deep learning architecture for road detection in LIDAR scans}
\label{roasegsection}
\paragraph{}
To properly detect the road in LIDAR scans, we propose an architecture that is heavily inspired by Squeezeseg~\cite{wu2018squeezeseg} and SqueezeSegV2~\cite{wu2019squeezesegv2}, as those network are particularly efficient to process LIDAR scans. However, we introduce several refinements to the original architectures and training schemes, so as to better fit the task of road detection, and allow for evidential fusion of segmented scans. We name the resulting architecture "RoadSeg".
\subsection{Dense range image generation from LIDAR scans}
\begin{figure}[H]
\centering
\includegraphics[width=0.5\textwidth]{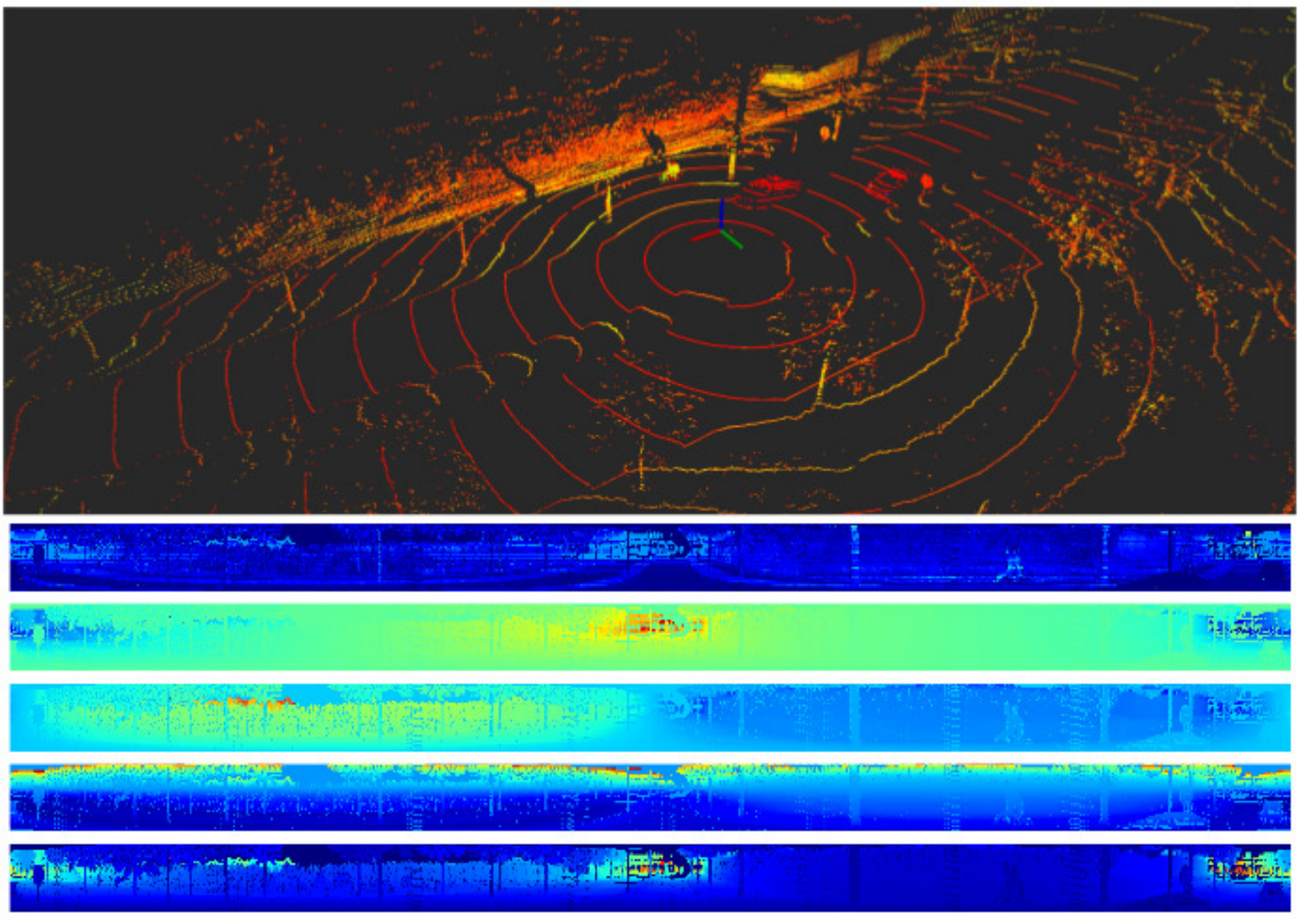}
\caption{Conversion of a LIDAR scan into a multi-channel range image by spherical projection. The channels in the lower part correspond to the returned intensity, the Cartesian coordinates, and the measured range.}
\label{spher_proj}
\end{figure}
\paragraph{}
Similarly to SqueezeSeg, RoadSeg processes LIDAR scans that are converted into dense range images. Such a representation corresponds to raw sensor measurements, as LIDAR scanners actually return distances at fixed angular positions. The dimensions of those range images then depend on the specifications of the LIDAR sensor. In this work, RoadSeg is assumed to process scans obtained from Velodyne VLP32C sensors, operating at 10Hz, and scanning at 360\degree. These sensors rely on 32 stacked lasers, each scanning at a different zenith angle, with a horizontal angular resolution of 0.2\degree when operating at 10Hz. As such, only 32*360/0.2 angular positions can be observed by the sensor. Each LIDAR scan obtained from a VLP32C can then be represented by a 32*1800*C grid, with C being the number of features available for each point. This grid can be considered as an image with an height of 32 pixels, a width of 1800 pixels, and C channels. Let (x,y,z) be the Cartesian coordinates of a LIDAR point. Let $\alpha$, $\beta$ be the indexes of the pixel that correspond to this point, and RingId the index of the laser that measured the point. A RingId of 0 (respectively 31) indicates that the point is measured by the topmost (respectively lowest) laser. Then $\alpha$ and $\beta$ are such that:
\[
\alpha = arcsin(\frac{z}{\sqrt{x^2+y^2+z^2}})*5
\]
\[
\beta = RingId
\]
\paragraph{}
The channels of the pixel at the ($\alpha$,$\beta$) position can then be filled with the features of the corresponding point. Figure~\ref{spher_proj} presents the resulting ranges images obtained after spherical projection of LIDAR points. In practice, the VLP32C scanner returns ethernet packets containing measured ranges that are already ordered by laser and azimuth position. The range images can then be obtained directly from the sensor, without having to pre-process a cartesian LIDAR scan. This projection was only used to manually label data, in practice. But when run on the vehicle, RoadSeg processes range images that are directly obtained from the ethernet packets.
\paragraph{} 
In total, we chose to represent a LIDAR point by its Cartesian coordinates, its spherical coordinates, and its returned intensity. Additionally, we also use a validity channel, as done for SqueezeSegV2. A point is valid, and the corresponding validity is equal to 1, if the range that was measured at its angular position is strictly positive. Otherwise, the point is not valid, which can correspond to missed detections, or lost ethernet LIDAR packets. 
\begin{figure*}[!h]
\begin{subfigure}{0.5\linewidth}
\includegraphics[width=\textwidth]{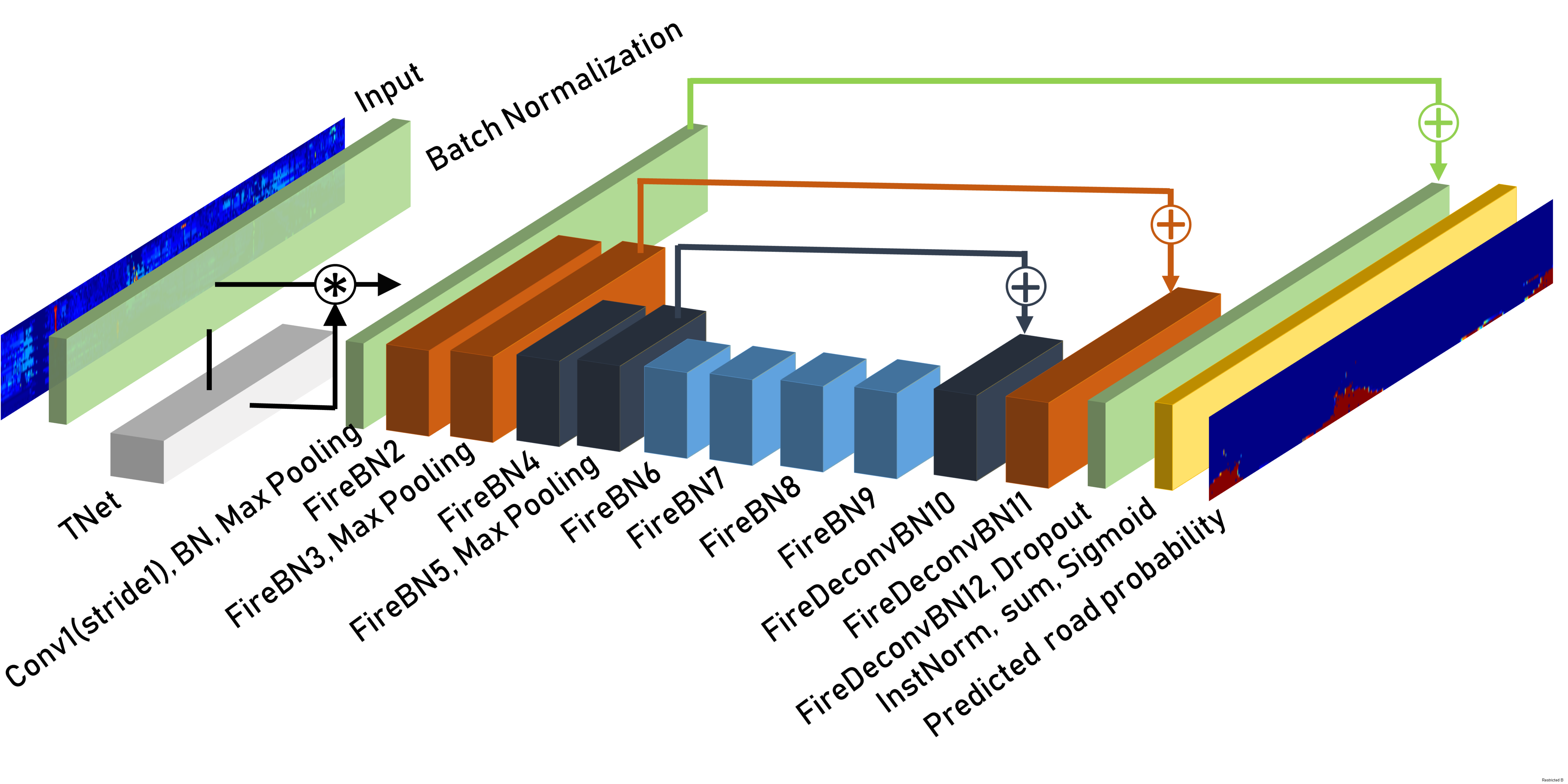}
\caption{General road classification architecture}
\end{subfigure}
\begin{subfigure}{0.5\linewidth}
\centering
\includegraphics[width=\textwidth]{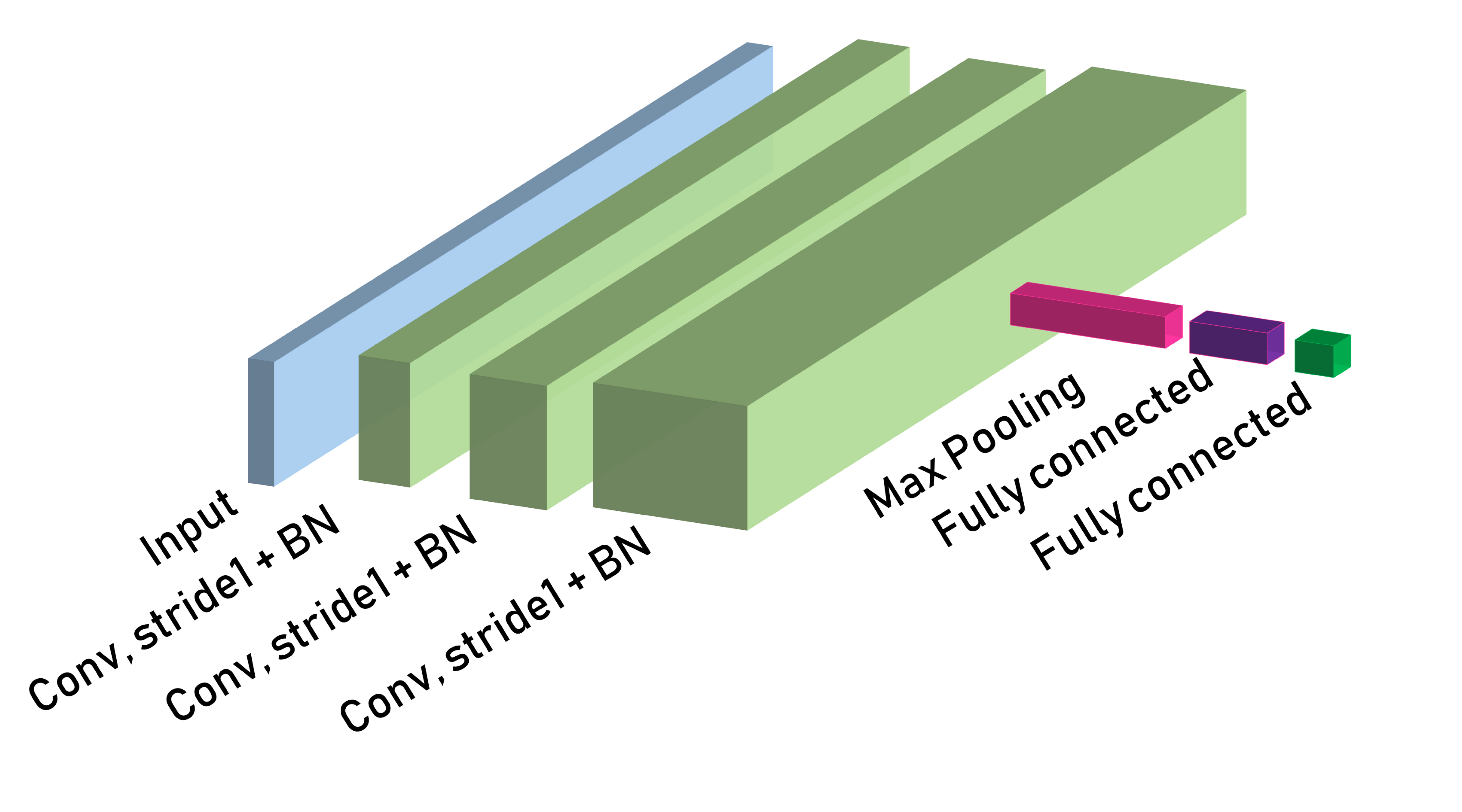}
\caption{T-Net of RoadSeg}
\end{subfigure}
\caption{RoadSeg, a refined SqueezeSeg-like architecture for road detection and evidential fusion}
\label{roadseg}
\end{figure*}
\subsection{From SqueezeSeg to RoadSeg}
\paragraph{}
Figure~\ref{roadseg} depicts the RoadSeg architecture, which is very close to the original SqueezeSeg architecture, as it also relies on the use of Fire and Fire-deconvolutional layers, and skip connections. RoadSeg also only downsamples horizontally, as SqueezeSeg and SqueezeSegV2. RoadSeg uses one downsampling step less than SqueezeSegV2. Indeed, the first convolutional layer of RoadSeg has a stride of 1 in both direction, while SqueezeSeg has a horizontal stride of 2. The kernel size is however still equal to 3. This is justified by the fact that, in the case of road detection, downsampling might make road hard to distinguish at long-range. SqueezeSeg benefitted from this downsampling because it was originally designed for semantic segmentation of road users, which are often relatively close to the sensor. RoadSeg then also has one deconvolutional layer less than SqueezeSeg. Additionnally, as done in SqueezeSegV2, RoadSeg uses Batch Normalization after each convolution, both inside and outside the Fire layers. 
\paragraph{}
An initial Batch Normalization is also applied on the input of the network. This mechanism was proposed to replace the manual normalization step, that is used in both SqueezeSeg and SqueezeSegV2. As the normalization parameters initially used in SqueezeSeg and SqueezeSegV2 are computed from the train set, manual normalization assumes that the sensor setup is the same in the train, validation and test sets. As our labelled LIDAR scans were not acquired by the same vehicles, this assumption does not hold anymore. Applying Batch Normalization to the inputs of RoadSeg is thus a way to train the network on inputs that are not exactly normalized. A T-Net, inspired by PointNet~\cite{qi2017pointnet}, can also be used to realign the LIDAR scans, by predicting a $3\times3$ affine transformation matrix that is applied on the Cartesian coordinates of the LIDAR points. The TNet is composed of three $1\times1$ convolutional layers, with respective output sizes of 32, 64 and 512 ; a channel-wise max-pooling; and three linear layers having output sizes of 256, 128 and 9. Finally, according to~Equations~\ref{minnorm} and~\ref{wdres}, the final convolutional layer is replaced, in RoadSeg, by an Instance-Normalization layer and a sum over the output channels.
\paragraph{}
In order to limit the total number of weights in RoadSeg, the number of output channels produced by the Fire layers was reduced, with regards to their counterparts in SqueezeSeg and SqueezeSegV2. Table~\ref{vsfire} compares the Fire layers in RoadSeg and in SqueezeSeg/SqueezeSegV2. The Fire-deconvolutional layers were left unchanged. The Conditional Random Field (CRF) layer that was, originally, refining the outputs of SqueezeSeg and SqueezeSegV2, was removed for several reasons. First, as it was originally implemented as a recurrent neural network, generating evidential mass functions from this layer is more complex than from a regular convolutional layer, or an Instance-Normalization layer. Secondly, removing this layer reduces the inference time of RoadSeg. Finally, experiments showed than the CRF layer was degrading the performances of SqueezeSegV2, for the road detection task, and was thus assumed not to be suitable for RoadSeg. LIDAR scans  were collected in several urban and peri-urban areas in France, to do those experiments. Coarse training and validation sets were automatically labelled from HD Maps. Moreover, a test set was manually labelled, to obtain reliable performance indicators.
\begin{table}
\centering
\begin{tabular}{|l|c|c|}
   \hline
    \cellcolor{black} & \multicolumn{2}{c|}{Output channels} \\
   \hline
   Layer & SqueezeSegV1/V2 & RoadSeg \\
   \hline
   Fire2 & 128 & 96 \\
   Fire3 & 128 & 128 \\
   Fire4 & 256 & 192 \\
   Fire5 & 256 & 256 \\
   Fire6 & 384 & 256 \\
   Fire7 & 384 & 256 \\
   Fire8 & 384 & 256 \\
   Fire9 & 512 & 256 \\     
   \hline
\end{tabular}
\caption{Comparison of the fire layers in RoadSeg and SqueezeSeg}
\label{vsfire}
\end{table}
\newpage
\section{Automatic labellisation procedure of LIDAR scans from lane-level HD Maps}
\label{softlabels}

\paragraph{}
Reliable semantic segmentation labels are usually generated manually, by expert annotators. This however proves to be expensive and time consuming. Automatic labelling can instead be used to generate labelled LIDAR scans, on which a road detection algorithm could be trained. Ideally, an error model should be associated to an automatic labelling procedure, as the resulting data is likely to include errors. We thus chose to softly label the road in LIDAR scans from pre-existing lane-level maps, and according to a localization error model. 

\subsection{Soft-labelling procedure}
\label{subsec:softlabelprocedure}
\begin{figure*}[!h]
\centering
\includegraphics[width=\textwidth]{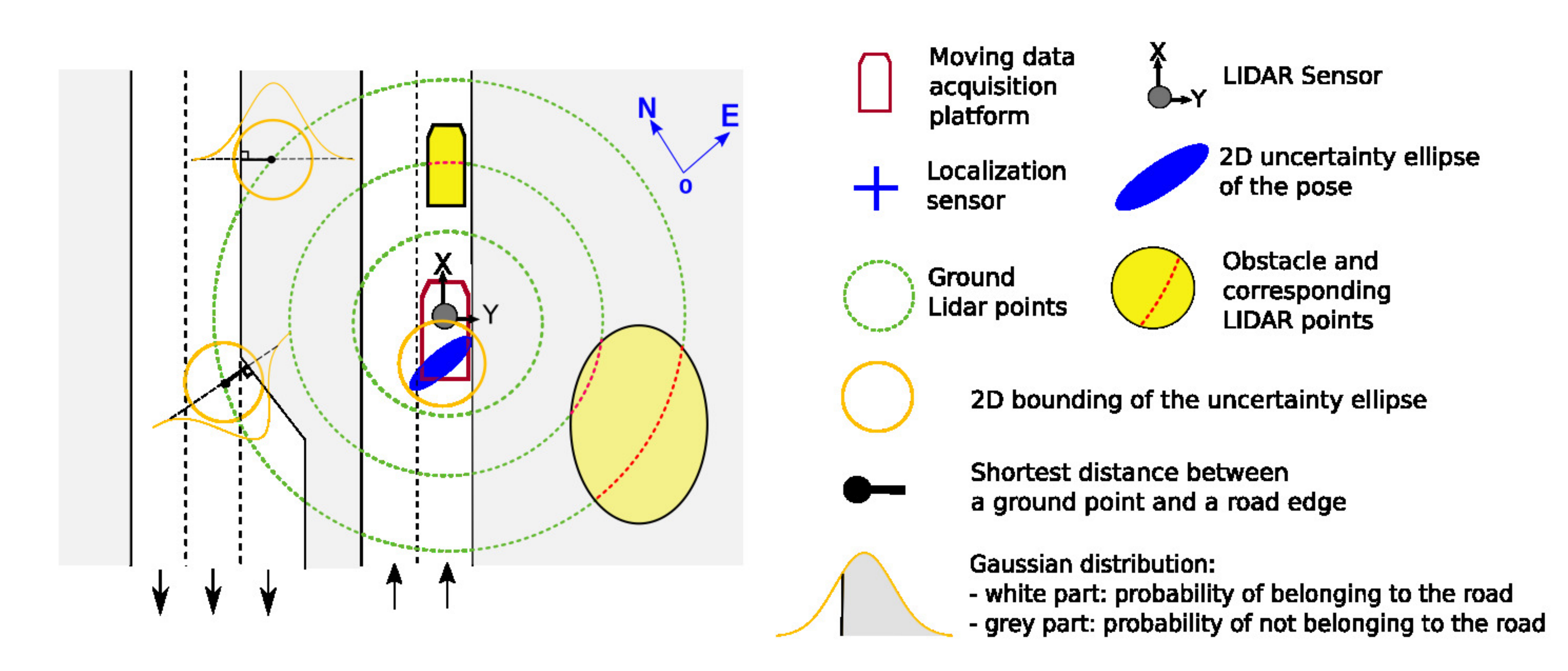}
\caption{Automatic soft-labelling of LIDAR scans from road maps}
\label{automaticlabelling}
\end{figure*}
\paragraph{}
The automatic soft-labelling procedure used in the context of this work is presented in Figure~\ref{automaticlabelling}. It was assumed that the LIDAR scans can be acquired in areas where reliable georeferenced maps, with correct positions and road dimensions, are available. Moreover, the LIDAR scans were supposed to be acquired from a moving data acquisition platform, which includes a LIDAR scanner that can perceive the ground and obstacles, and a GNSS localization system which returns an estimation of its pose in the map frame, and a corresponding covariance matrix. Those sensors were considered to be rigidly attached to the acquisition platform, and calibrated together. The coordinates in the map frame were expressed in terms of northing and easting offsets with regards to an origin. This origin was close to the data acquisition platform. Under the classical assumption that the localization error follows a zero-mean Gaussian model~\cite{ahmad2014characterization}, an uncertainty ellipse could then be obtained, from the covariance matrix associated with the current pose. A probability of belonging to a mapped road could then be estimated, for each LIDAR point. 
\paragraph{}
Let $X^l_i(x_i^l,y_i^l,z_i^l)$ represent the a LIDAR point, and $X^m_i(x_i^m,y_i^m,z_i^m)$ the corresponding point after projection in the map frame. Let $R_i$ and $T_i$ be the rotation matrix, and translation vector, to projet $X^l_i$ into the map frame. We were using a scanning LIDAR with a parameterized sweeping speed that we note $S$, and assumed that this rotation speed remained constant over time. The vehicle motion thus had to be accounted for, before projection into the map frame. We neglected the rotation speed of the vehicle during the scan. We also assumed that the longitudinal speed of the vehicle was constant during the scan, and that the sensor was positionned so that its X-axis was parallel with the vehicle's forward direction.  Let $\Delta t_i$ be the estimated time offset between the beginning of the scan, and the moment when $X^l_i$ was acquired. A spinning LIDAR records points at pre-defined azimuth positions. Let $\theta_i$ be the azimuth position, in degrees, at which $X^l_i$ was acquired. Then, $\Delta t_i$ can be estimated by:

\begin{equation}
\Delta t_i \approx S*\frac{\theta_i}{360}
\end{equation}

We also had access to the closest instantaneous measure of the vehicle's speed at the end of the scan, via its CAN network. We note this speed $v_s$. The projection of $X^l_i$ into the map frame is then done as follows:

\begin{align}
X^{m}_{i} &= \begin{bmatrix}
           x_i^m \\
           y_i^m \\
           z_i^m
         \end{bmatrix} 
         = \begin{bmatrix}
         R & T \\
         0 & 1
         \end{bmatrix}
         \begin{bmatrix}
         x_i^l  + \Delta t_i * v_s\\
         y_i^l \\
         z_i^l \\
         1       
         \end{bmatrix}
         \label{motioncomp}
\end{align}

\begin{figure*}[h!]
\centering
\begin{subfigure}{.5\textwidth}
  \centering
  \includegraphics[width=0.8\linewidth]{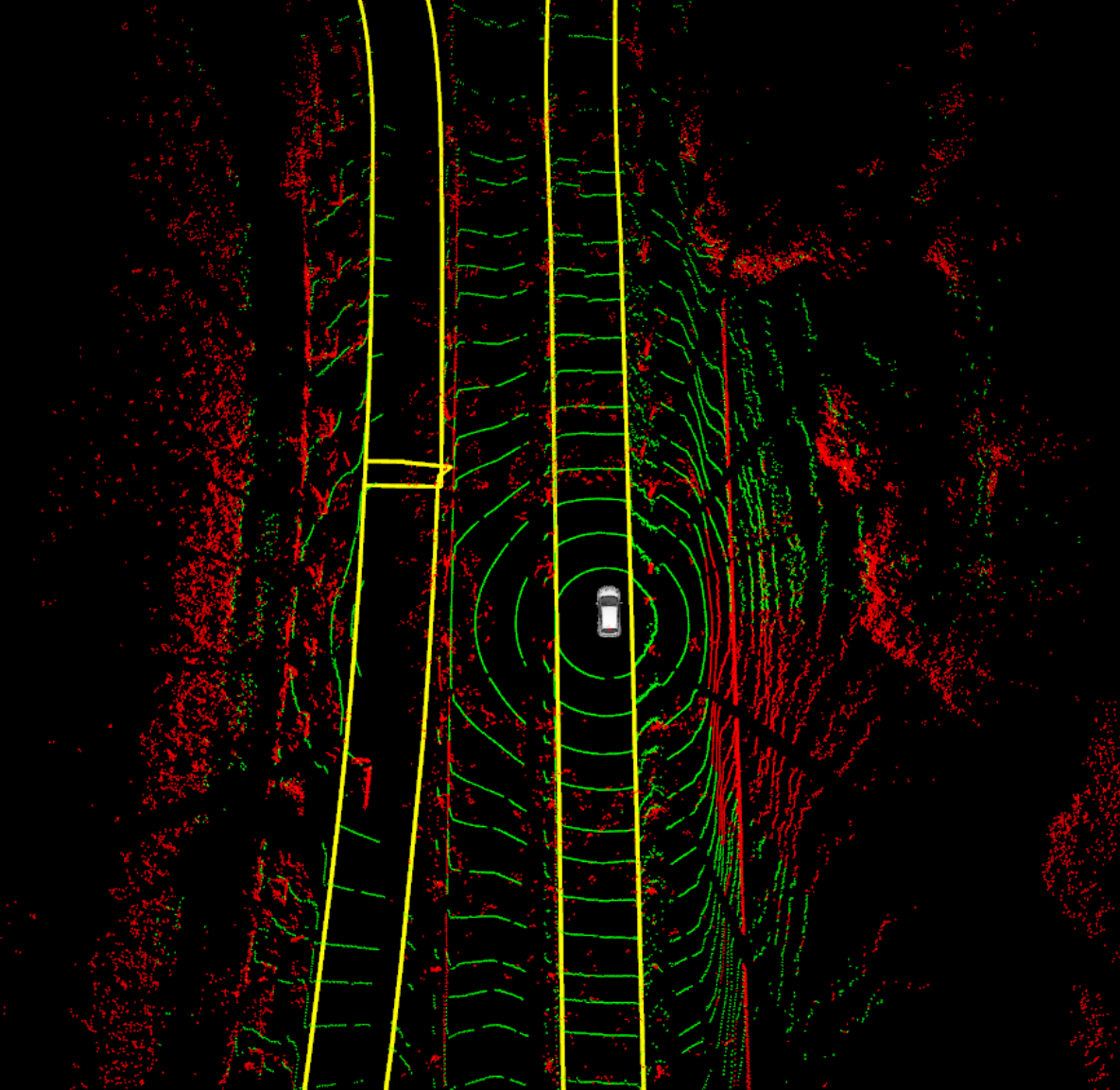}
  \caption{Use case 1: two opposite lanes}
  %\label{fig:sub1}
\end{subfigure}%
\begin{subfigure}{.5\textwidth}
  \centering
  \includegraphics[width=0.8\linewidth]{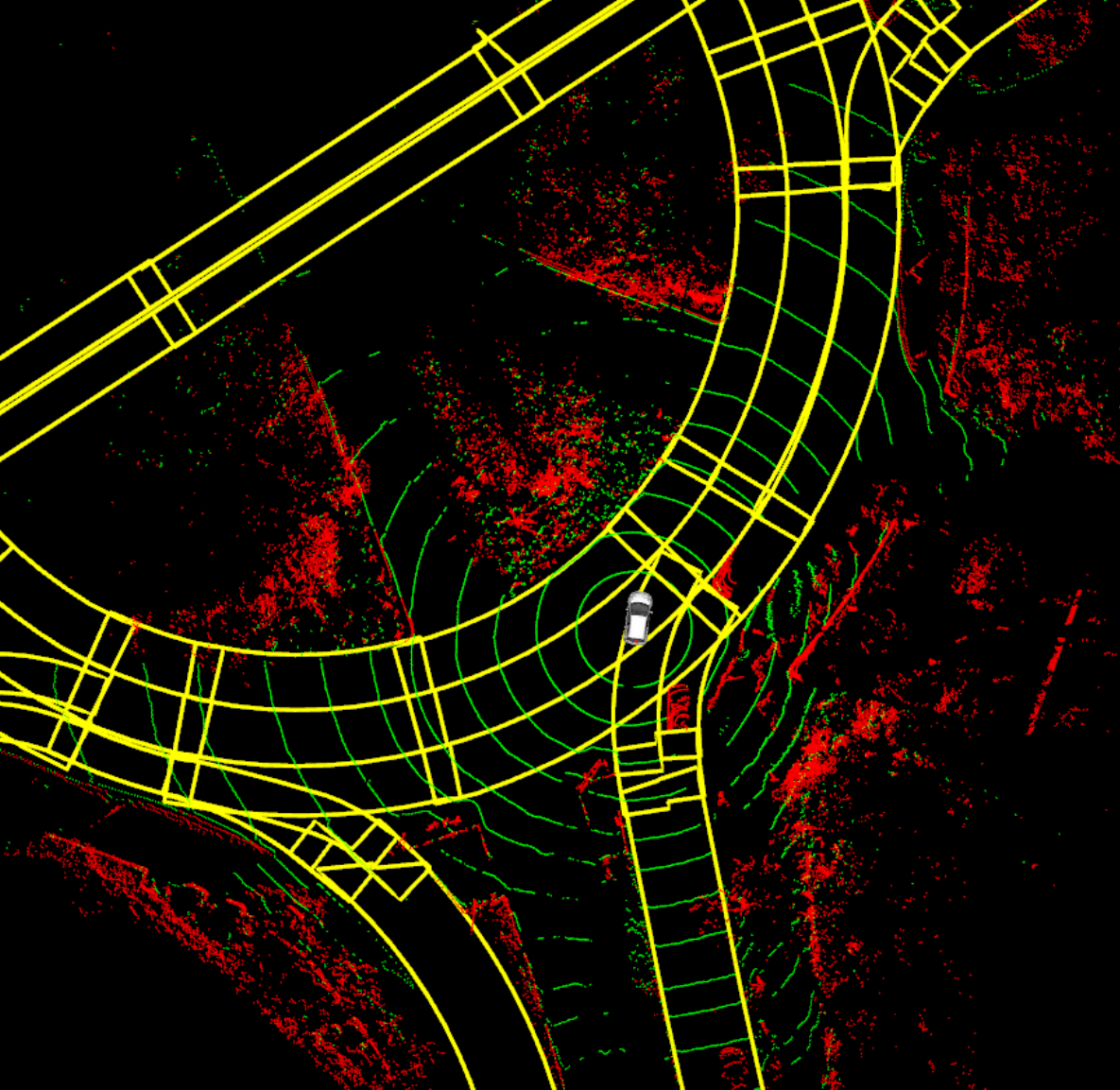}
  \caption{Use case 2: Roundabout}
  %\label{fig:sub2}
\end{subfigure}
\begin{subfigure}{.5\textwidth}
  \centering
  \includegraphics[width=0.8\linewidth]{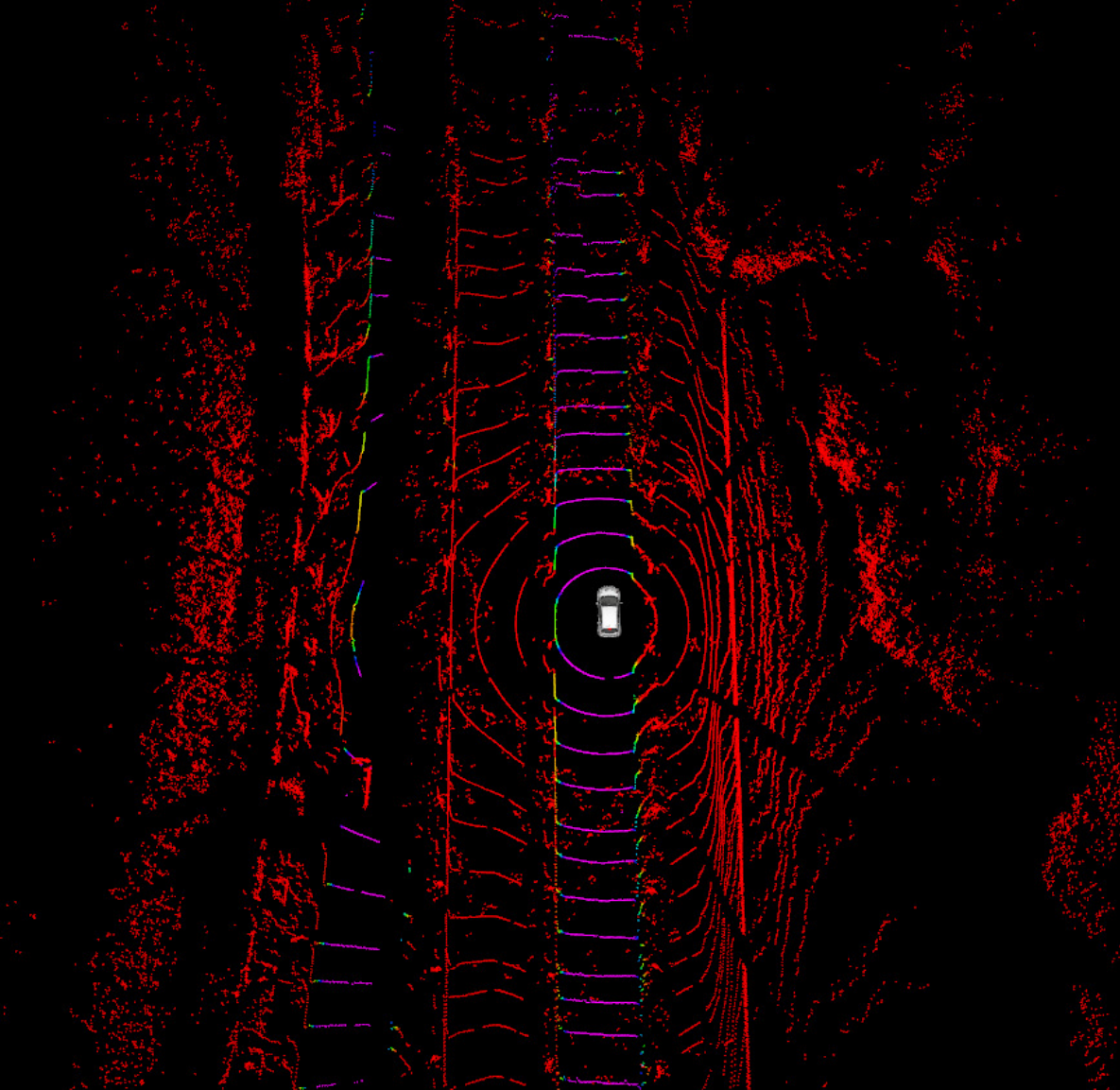}
  \caption{Use case 1: resulting labelled LIDAR can}
  %\label{fig:sub1}
\end{subfigure}%
\begin{subfigure}{.5\textwidth}
  \centering
  \includegraphics[width=0.8\linewidth]{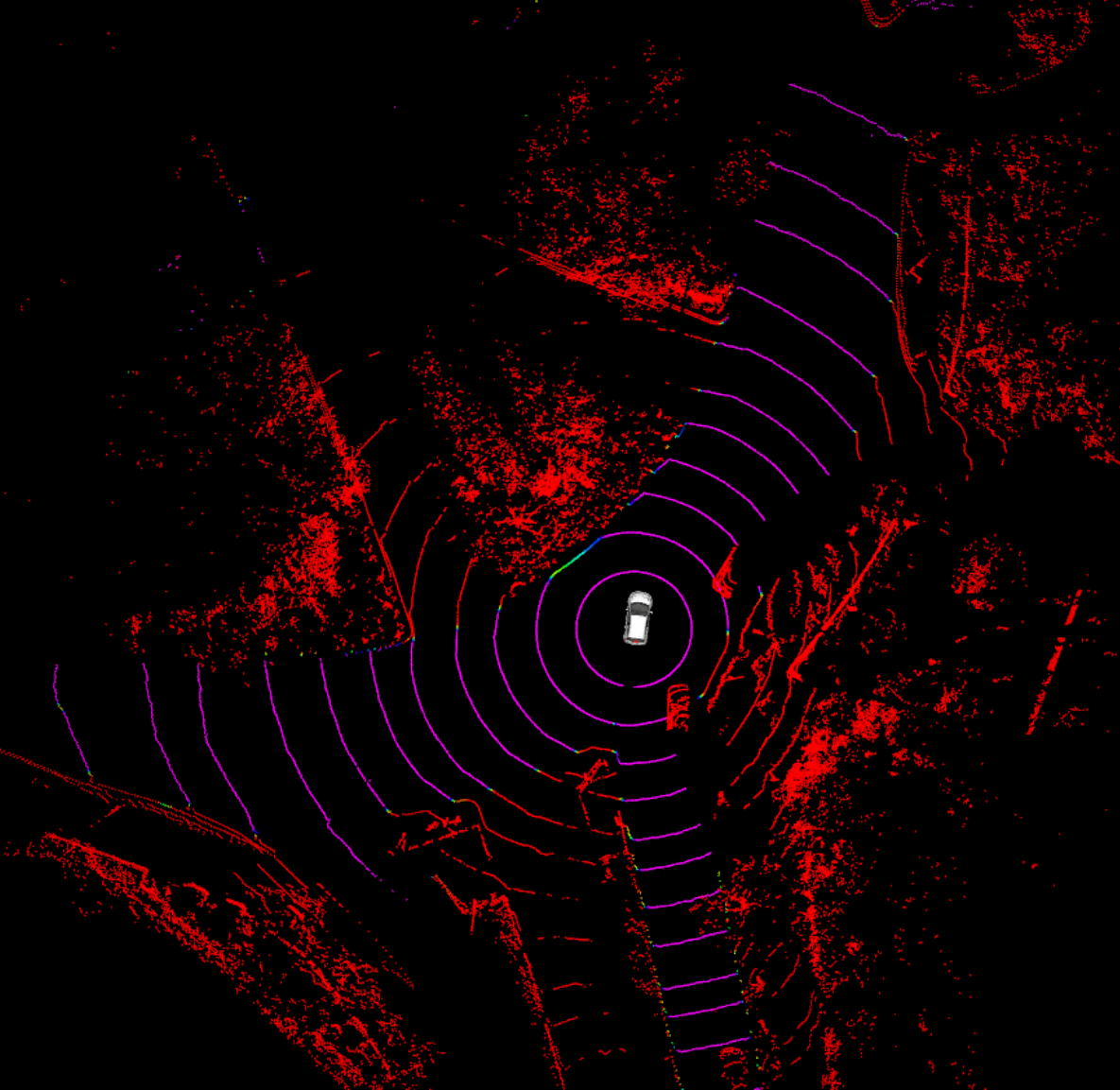}
  \caption{Use case 2: Automatically labelled LIDAR scan}
  %\label{fig:sub2}
\end{subfigure}
\centering
\caption{Use cases and examples of automatically labelled LIDAR scans. The data acquisition platform is depicted in white, and the map in yellow. Green points are pre-classified as ground points ; the purpler a point is, the higher its probability of belonging to the road is; the redder a point is, the lower its probability of belonging to the road is.}
\label{fig:usecases}
\end{figure*}

Let $P_R(X^l_i)$, the labelled probability that $X^l_i$ belonged to the road. First, a ground detection step was used to segregate obvious obstacles from points that, potentially, belonged to the road. It was assumed that the algorithm does not falsely classify ground points as obstacles. If $X^l_i$ did not belong to the detected ground, then $P_R(X^l_i)=0$, since it was considered to belong to an obstacle. Otherwise, $X^l_i$ was projected into the map frame. Given that the projection into the map frame relied on a rigid transformation, the localization uncertainty of the resulting $X^m_i$ was equal to the uncertainty of the pose measured by the GNSS localization system. The closer from a road edge $X^m_i$ was, the lower the probability that it belonged to the road is. Let $d_i$ be the minimum distance between $X^m_i$ and a mapped road edge. The covariance matrix given by the GNSS localization system could, ideally, have been used to estimate the standard-deviation of the Gaussian distribution that models the error on $d_i$, by having considered the heading of the road in the map frame~\cite{tao2015road}. However, the value of this heading can be ambiguous in curbs or roundabouts. To account for those use cases and facilitate the computations, a bounding of this standard-deviation, noted $\sigma_b$, was thus estimated instead. Let $\sigma_n$ and $\sigma_e$ be, respectively, the standard-deviation in the northing and easting directions; then $\sigma_b$ was estimated as follows:
\begin{equation}
\sigma_b = Max(\sigma_n,\sigma_e)+\gamma
\label{sigmab}
\end{equation}
The $\gamma$ term is a hyperparameter used to account for the uncertainty in the measures of the LIDAR sensor, and possible errors in motion compensation. It also covers inevitable errors in the extrinsic calibration parameters, that we used to register the LIDAR and the localization. Indeed, such parameters are only accurate up to a certain point~\cite{d2018registration}. Then, if $X^l_i$ belonged to the ground, $P_R(X^l_i)$ was estimated as follows:
If $X^m_i$ falls into a mapped road: 
\begin{equation}
P_R(X^l_i) = \int_{-\infty}^{d_i} \frac{1}{\sigma_b\sqrt{2\pi}}\exp^{-\frac{1}{2}(\frac{x}{\sigma_b})^2}dx
\end{equation}
Otherwise:
\begin{equation}
P_R(X^l_i) = 1-\int_{-\infty}^{d_i} \frac{1}{\sigma_b\sqrt{2\pi}}\exp^{-\frac{1}{2}(\frac{x}{\sigma_b})^2}dx
\end{equation}
It has to be noted that, even though the motion was compensated in the forward direction via Equation~\ref{motioncomp}, this probability is associated to the non-compensated LIDAR point. The reason for that was that, even though our compensation procedure empirically seemed sufficient to build our training set, it was far from being very accurate, as many parameters were not considered (among others: the variations in the longitudinal speed of the vehicle over the scanning process, the variations in the rotation speed of the LIDAR over the scanning process, the rotation speed of the vehicle, the noises in speed measurements, the differences in framerates among the odometry and LIDAR sensors). In actual autonomous driving situations, those parameters would have to be accounted for, probably via approaches inspired by Kalman filters, which rely on manualy tuned hyper-parameters. So as not to have a road detection system trained on data with a specific compensation process, which would potentially have to be tuned manually and could thus have to be modified for every experiment, we decicided to build our training set from uncompensated LIDAR scans, and to only compensate roughly via Equation~\ref{motioncomp} when labelling automatically from the map. This way, the training set, and the road detection systems trained on it, are independant from any compensation method. This allows for the use of any compensation method after road detection, without having to potentially generate training scans for every possible compensation method. Figure~\ref{fig:usecases} presents two use cases and the resulting softly labelled LIDAR scans.

\subsection{Data collection and resulting dataset}
\begin{figure}
\centering
\includegraphics[width=0.9\linewidth]{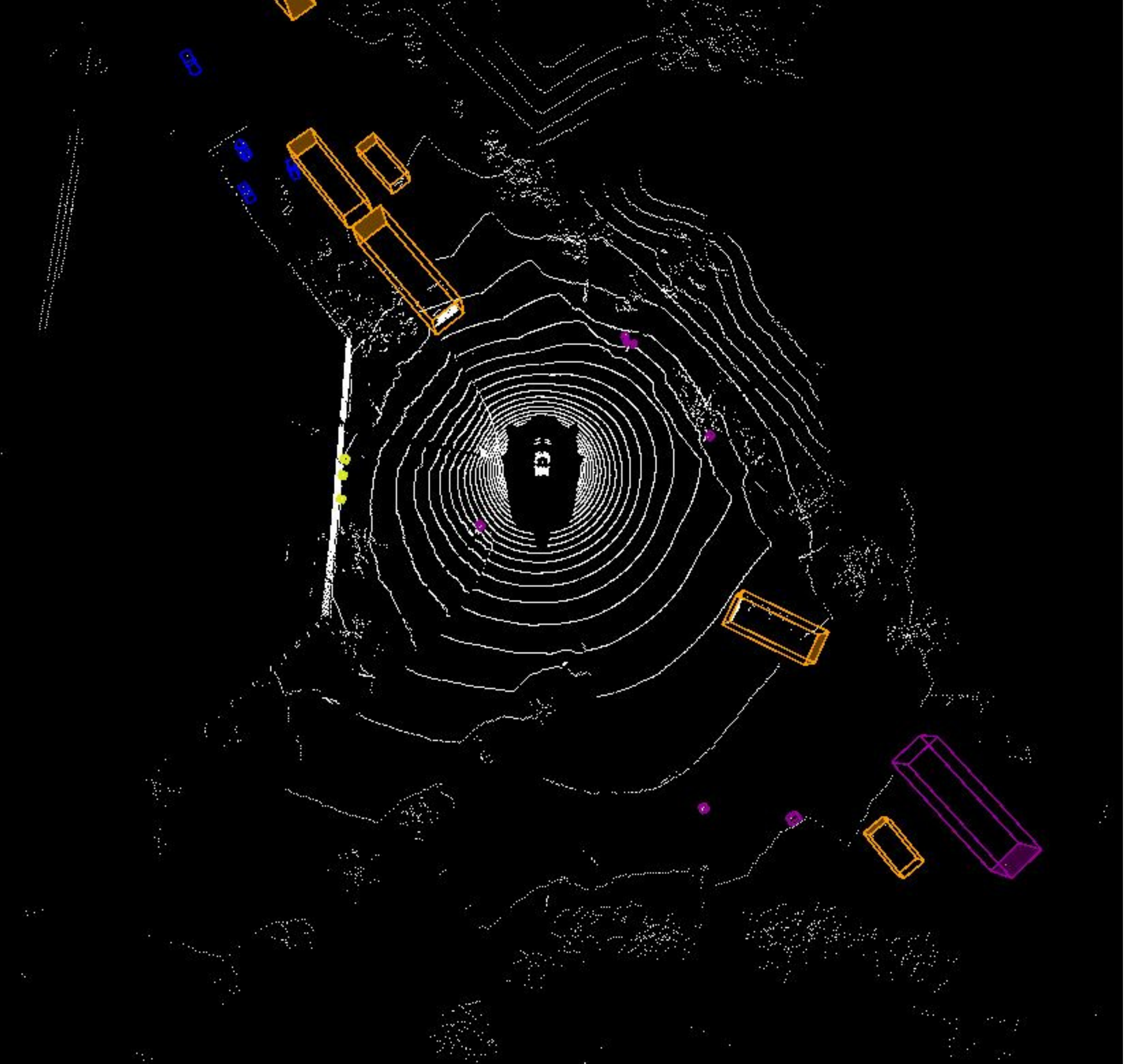}
  \caption{Example of labelled scan from the nuScenes dataset. The horizontal resolution of the scan is significant, and most of the points belonging to the road are, actually, very close to the data acquisition platform.}
\label{fig:nuscenesfail}
\end{figure}

Appropriate data was needed to apply this automatic labelling procedure. Accurate maps and localization were needed. Open source maps, such as OpenStreetMap, were thus not considered, as the road dimensions, and especially their width, are rarely available and accurate. The NuScenes dataset~\cite{caesar2019nuscenes} includes LIDAR scans and an accurate map ; however, the localization of the vehicle, though extremely accurate, is not provided with uncertainty indicators, which prevented us from using this dataset. Moreover, we observed that the Velodyne HDL32E LIDAR sensors, that were used to collect the nuScenes dataset, seemed improper for road detection. Indeed, this sensor has a vertical resolution of $1.33\deg$. As such, most of the road points in nuScenes were actually acquired very close to the data collection platform, as presented in Figure~\ref{fig:nuscenesfail}, limiting the use of this dataset for long-range road detection. Data was thus collected  via a LIDAR scanner that has a better vertical resolution, in several areas where Renault S.A.S has access to lane-level HD-map. The framework of those maps is described in~\cite{li2017lane}.

\subsubsection{Data acquisition platform}

\begin{figure*}[h!]
\centering
\begin{subfigure}{.5\textwidth}
  \centering
  \includegraphics[width=0.7\linewidth]{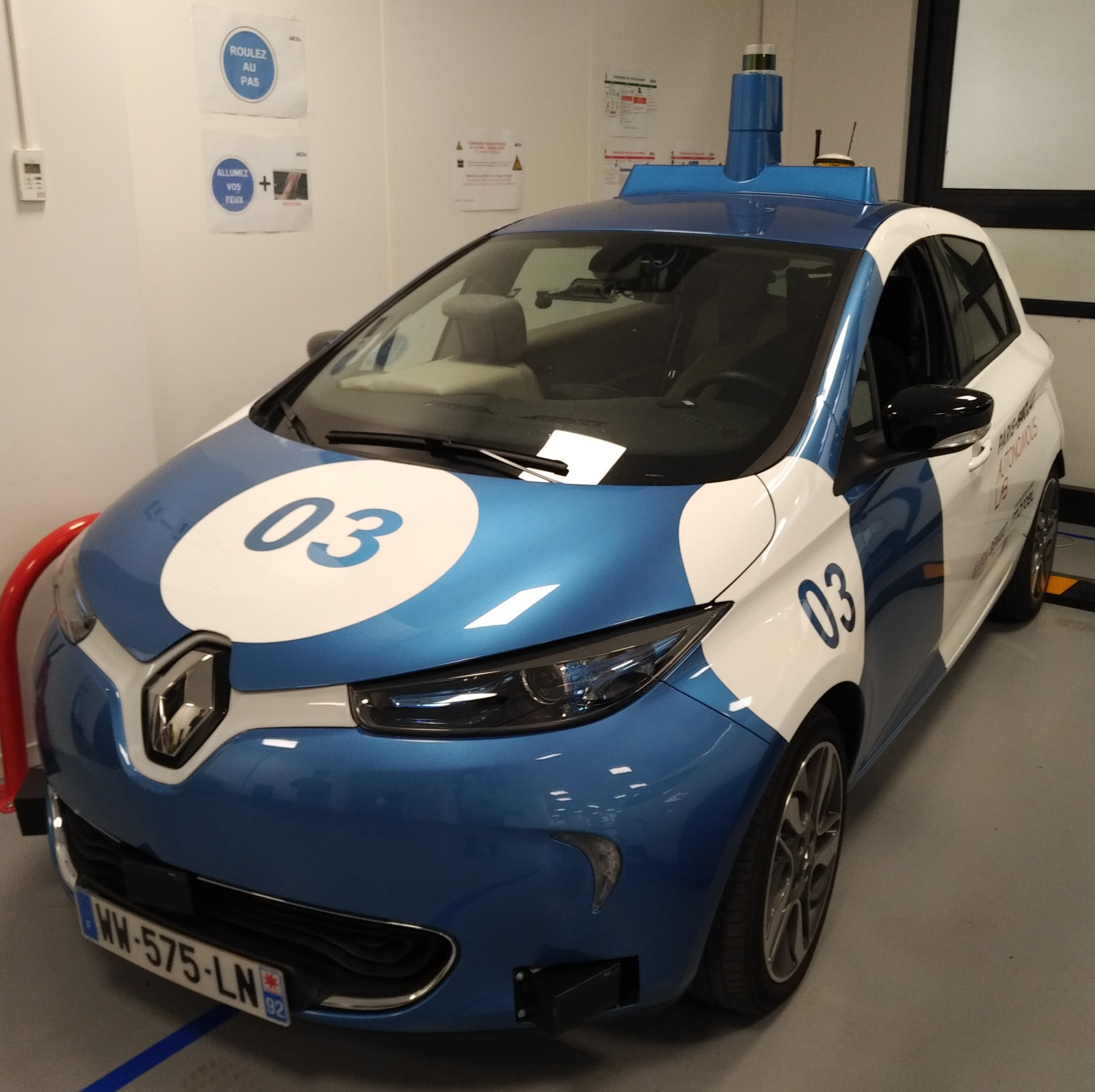}
  \caption{A ZoeCab platform}
\end{subfigure}%
\begin{subfigure}{.5\textwidth}
  \centering
  \includegraphics[width=0.7\linewidth]{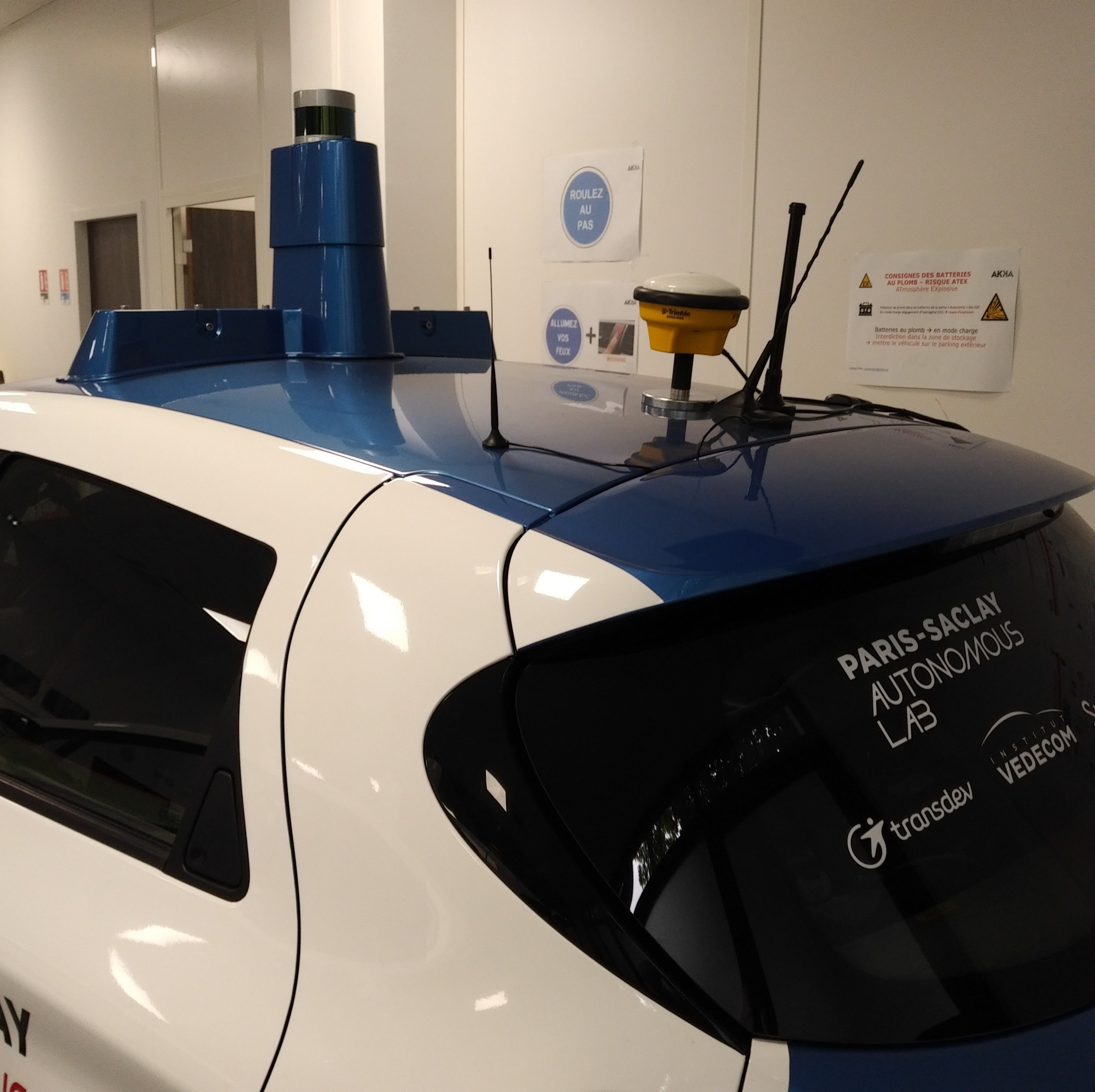}
  \caption{Velodyne VLP32C and GNSS antenna}
  %\label{fig:sub2}
\end{subfigure}
\caption{Data collection platform and sensor setup}
\label{zoecab}
\end{figure*}

\paragraph{}
ZoeCab platforms were used to collect data. ZoeCabs are robotized electric vehicles, based on Renault Zoes, that are augmented with perception and localization sensors, and intended to be deployed as autonomous shuttles in urban and peri-urban areas. 
Each vehicle embedded a VLP32C Velodyne LIDAR running at 10 Hz, and a Trimble BX940 inertial positioning system, which was coupled with an RTK Satinfo, so that centimeter-accurate localization could be achieved when RTK corrections were available. The PPS signal provided by the GPS receiver was used to synchronize the computers and sensors together. Figure~\ref{zoecab} presents one of the ZoeCab systems that were used for this work.
\subsubsection{Coarse training dataset}

\paragraph{}
In order to get diverse training samples, LIDAR and localization data was acquired in four different cities and locations in France, with two different ZoeCab systems: Compi\`{e}gne, Rambouillet, Renault Techocentre and the Paris-Saclay campus.
One of the vehicles was used to collect the data in Compiègne and Rambouillet, and the other one was used in Renault Technocentre and the Paris-Saclay campus.

\begin{figure*}[h]
\centering
\begin{subfigure}{0.45\textwidth}
  \centering
  \includegraphics[width=0.8\linewidth]{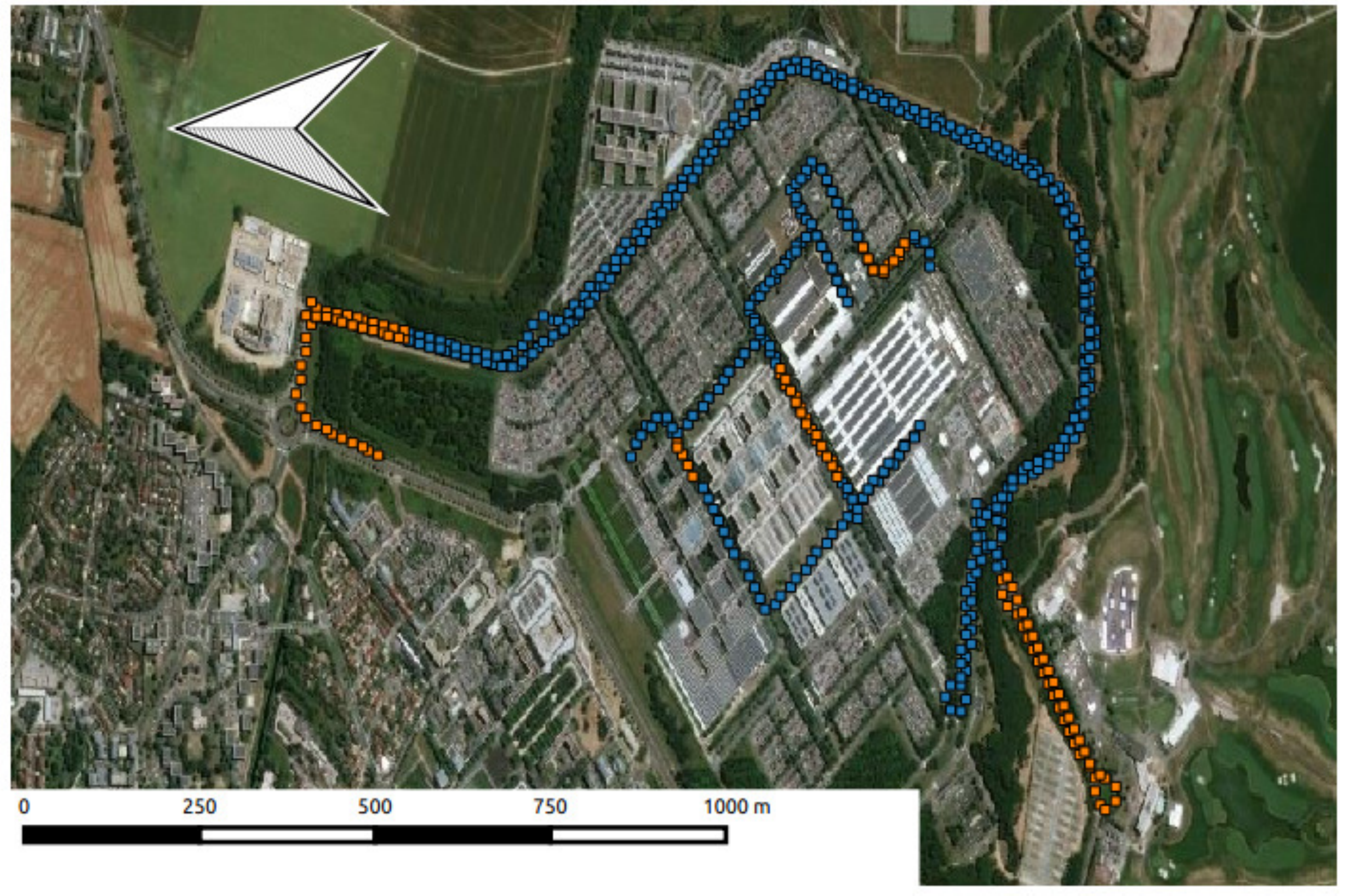}
  \caption{Technocentre}
  %\label{fig:sub1}
\end{subfigure}
\begin{subfigure}{0.45\textwidth}
  \centering
  \includegraphics[width=0.8\linewidth]{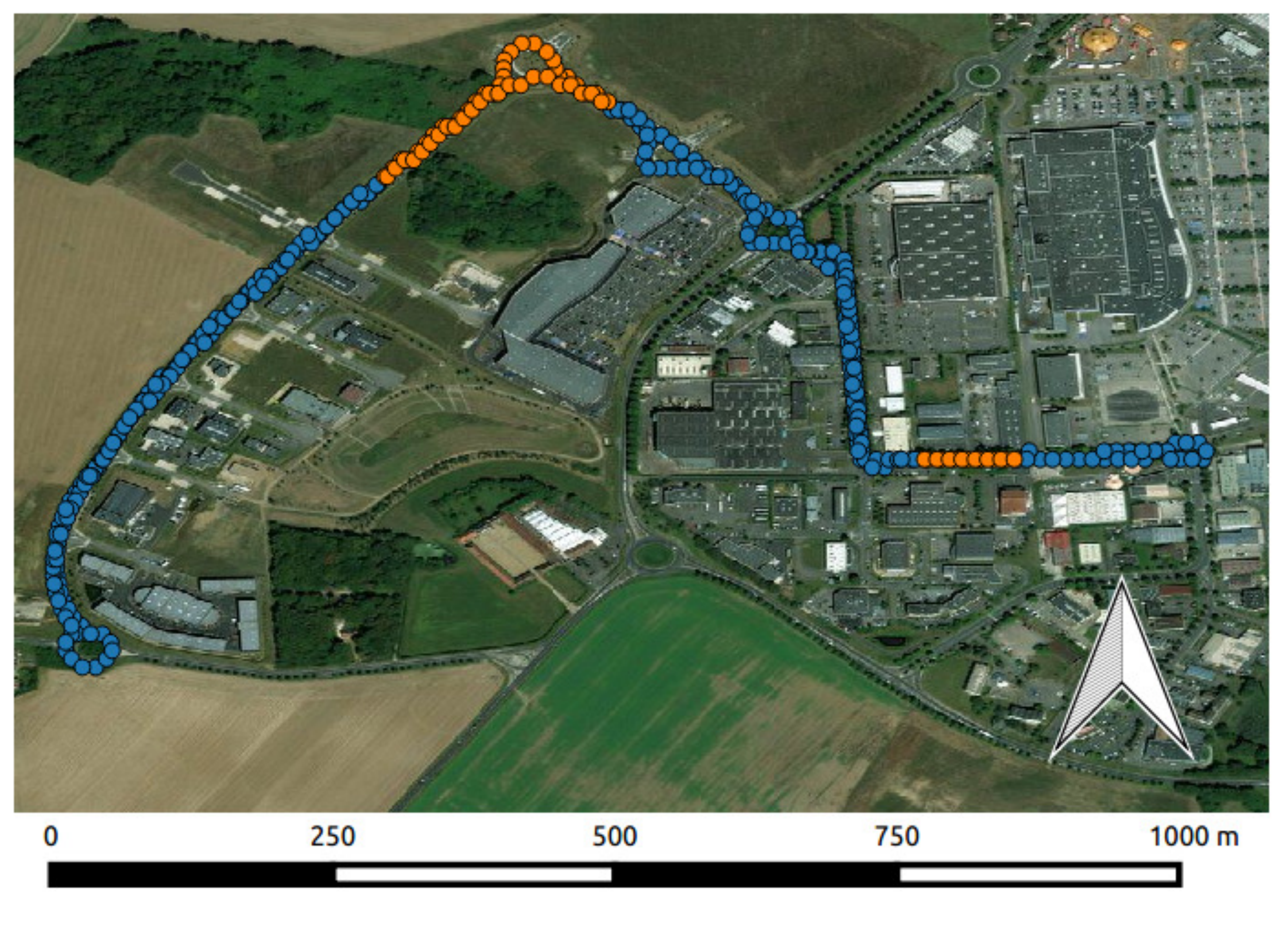}
  \caption{Rambouillet}
  %\label{fig:sub2}
\end{subfigure}\\
\begin{subfigure}{0.45\textwidth}
  \centering
  \includegraphics[width=0.7\linewidth]{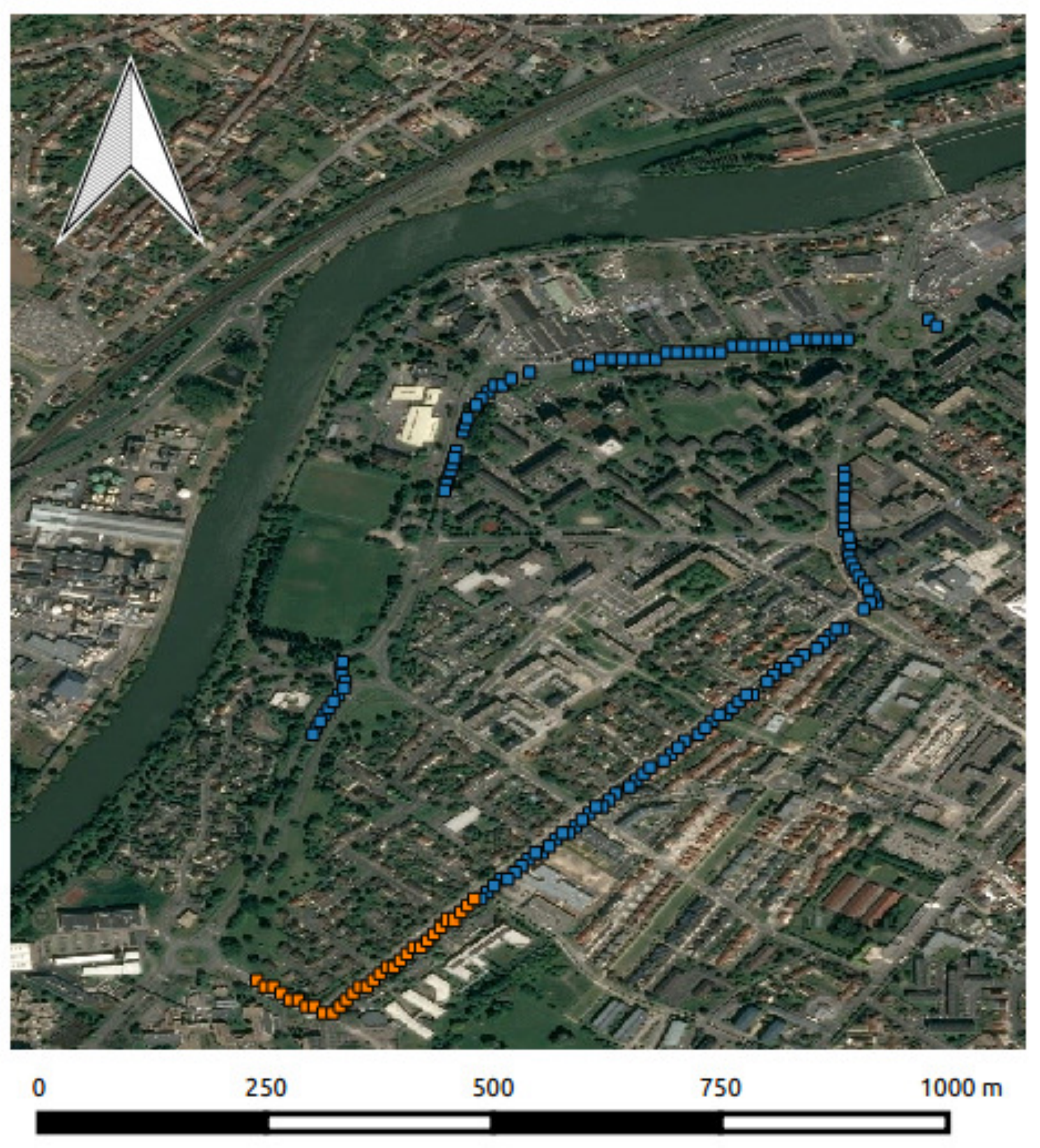}
  \caption{Compi\`{e}gne}
  %\label{fig:sub1}
\end{subfigure}
\begin{subfigure}{0.45\textwidth}
  \centering
  \includegraphics[width=\linewidth]{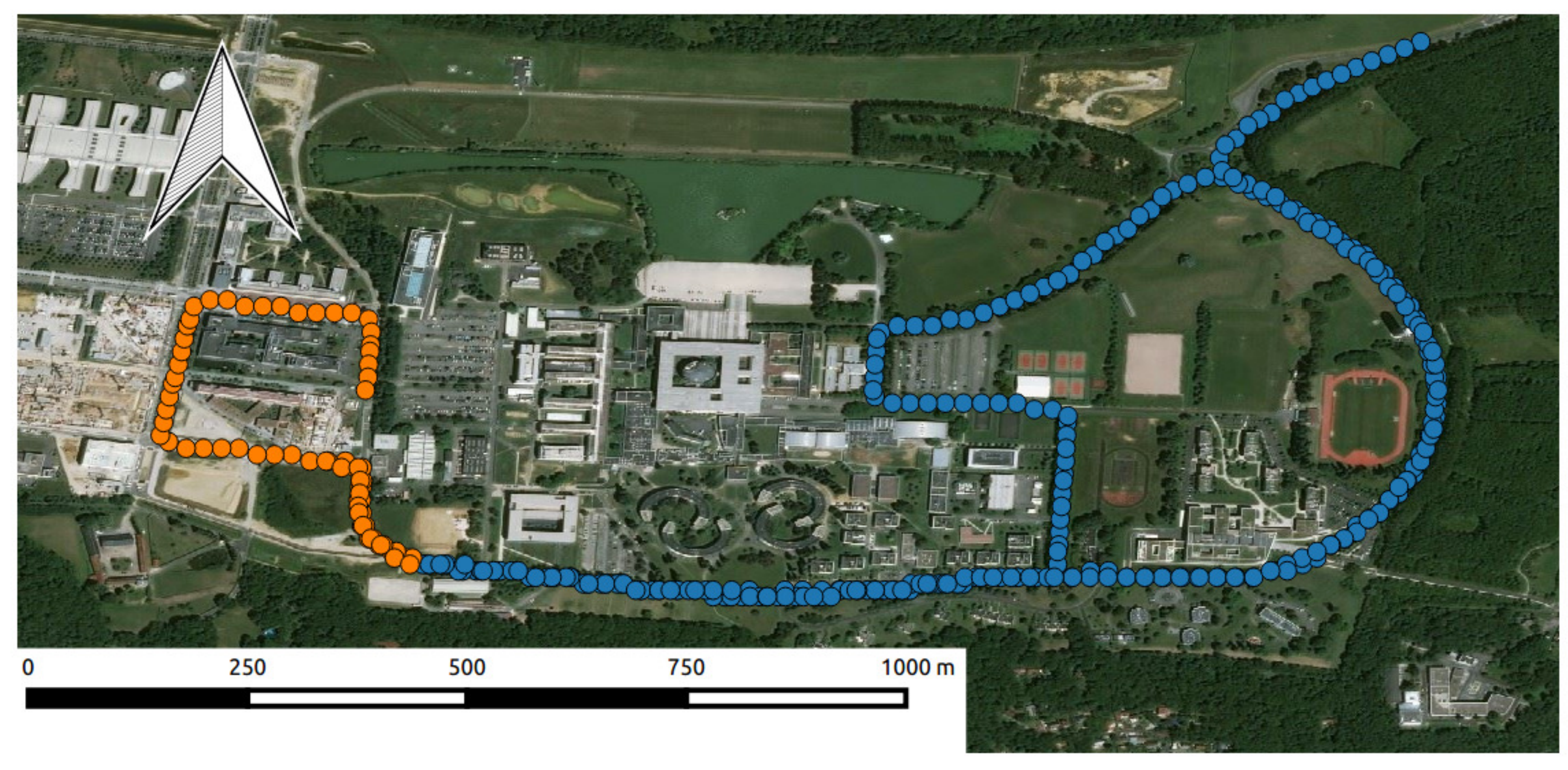}
  \caption{Paris-Saclay campus}
  %\label{fig:sub2}
\end{subfigure}
\caption{Automatically labelled dataset, in four locations. Each point indicates a position where a LIDAR scan was automatically labelled, and the white arrows are oriented towards the north direction. The blue points correspond to the training set, the orange points correspond to the validation set.}
\label{trainset}
\end{figure*}

\paragraph{}
All those locations are urban or peri-urban areas. The sensor setup was similar, but not identical, between those vehicles, with a a VLP32C LIDAR and GNSS antenna on top of the vehicle and the Trimble localization system in the trunk. LIDAR scans were only automatically labelled every ten meters, to limit the redundancy in the training set. As an accurate localization was needed to label the LIDAR point-clouds from the map, the LIDAR scans were only labelled when the variance in the easting and northing direction, associated with the pose estimated by the localization system, were lower than 0.5m. The $\gamma$ parameter in Equation~\ref{sigmab} was empirically set to 10 cm. The initial ground detection was realized thanks to a reimplementation of the algorithm described in~\cite{chu2017fast}. The vehicles were driven at typical speed of 30 km/h, in order to be able to be able to rely on our compensation procedure. Figure~\ref{trainset} presents the areas in which data was collected. The total number of collected and labelled samples for each location is reported in Table~\ref{train-table}.

\begin{table*}[h]
\begin{center}
\begin{tabular}{|c|c|c|c|c|c|}
  \hline
   \cellcolor{black} & Technocentre & Rambouillet & Compi\`{e}gne & Paris-Saclay Campus & Total \\
  \hline\hline
  Number of samples & 647 & 337 & 160 & 356 & 1500 \\ 
  \hline
\end{tabular}
\caption{Number of automatically labelled LIDAR scans} \label{train-table}
\end{center}
\end{table*}
\subsubsection{Test dataset}
\begin{figure*}[h]
\centering
\includegraphics[width=0.8\linewidth]{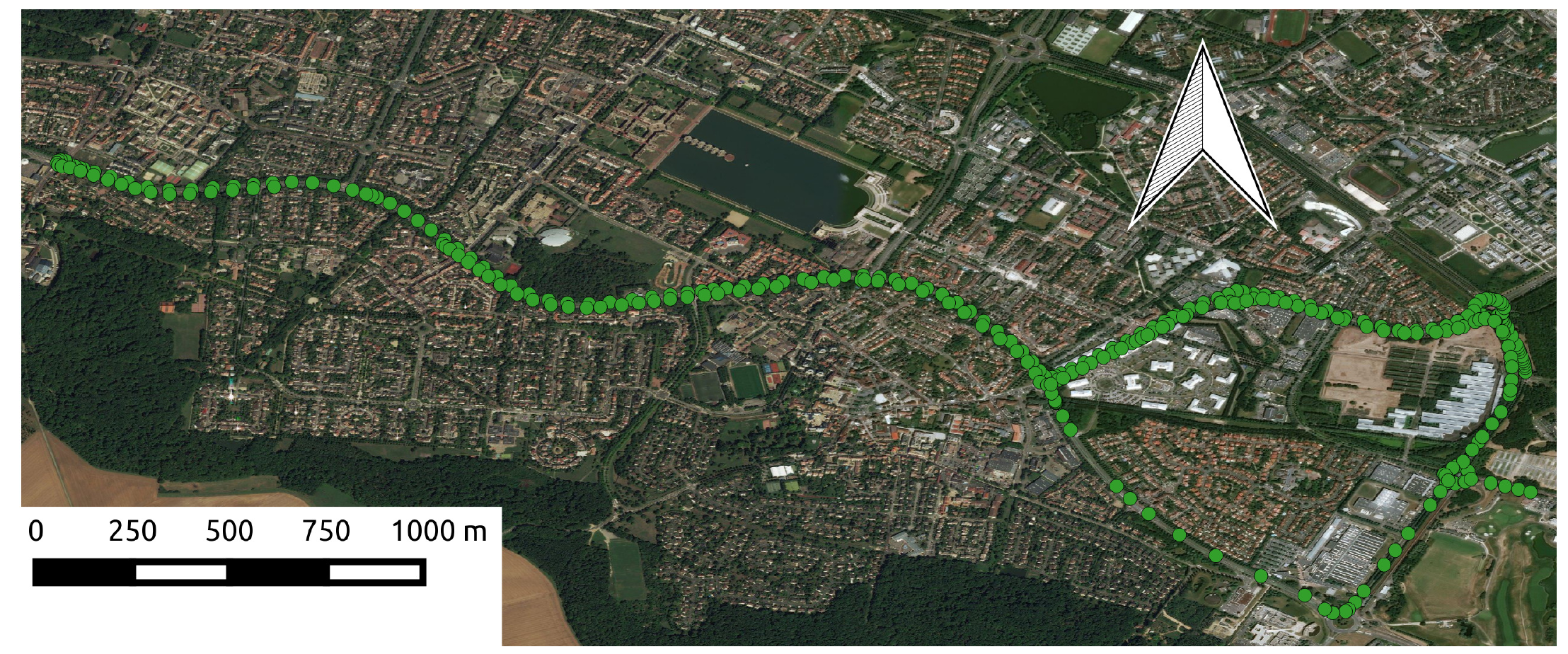}
\caption{Generation of the test dataset in Guyancourt. Each green point indicates a location where a manually labelled LIDAR scan was recorded.}
\label{test_dataset}
\end{figure*}

\begin{figure*}[h!]
\centering
\begin{subfigure}{0.40\linewidth}
\includegraphics[width=\textwidth]{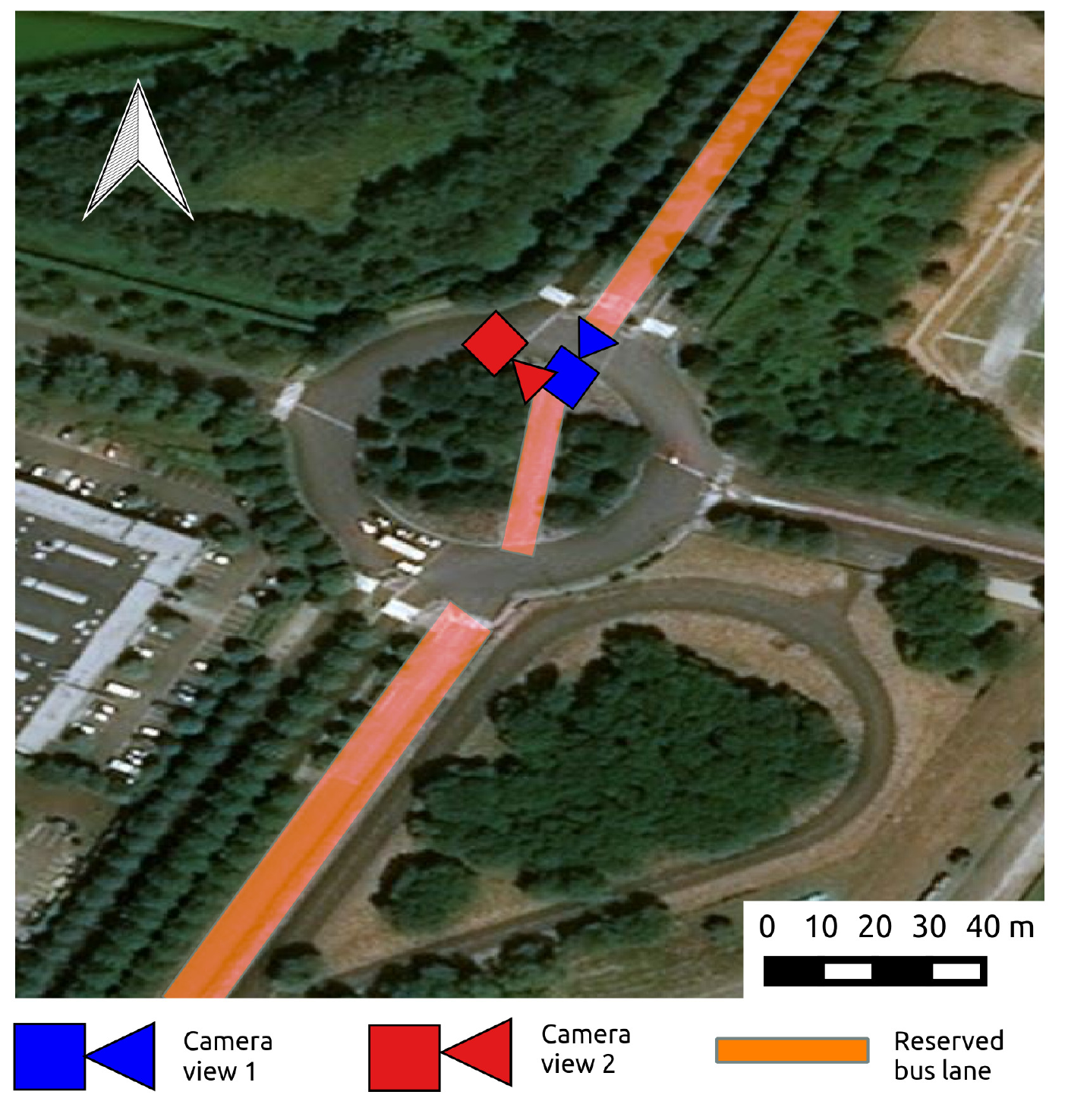}
\caption{Exemple of situation in the test set, with a reserved bus lane that goes through a roundabout}
\end{subfigure}
\begin{subfigure}{0.55\linewidth}
\centering
\includegraphics[width=\textwidth]{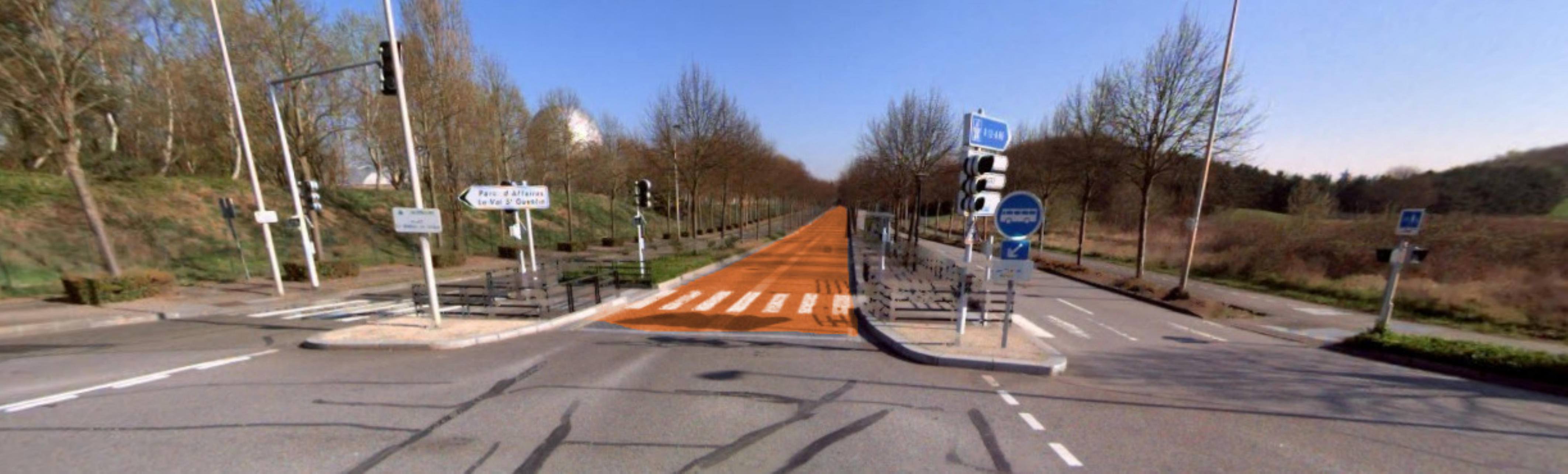}
\caption{Camera view 1 ; the orange mask represents the reserved the bus lane}
\label{view1}
\includegraphics[width=\textwidth]{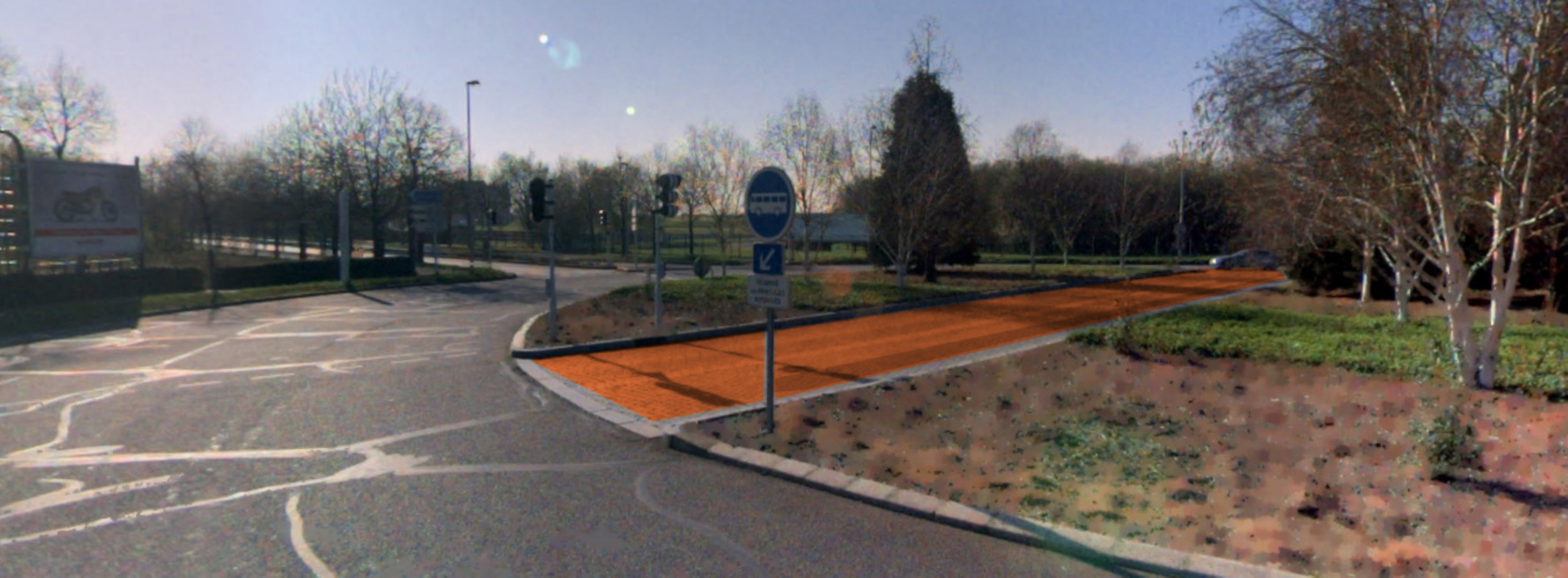}
\caption{Camera view 2 ; the orange mask represents the reserved the bus lane}
\label{view2}
\end{subfigure}
\begin{subfigure}{\textwidth}
\centering
\includegraphics[width=0.7\textwidth]{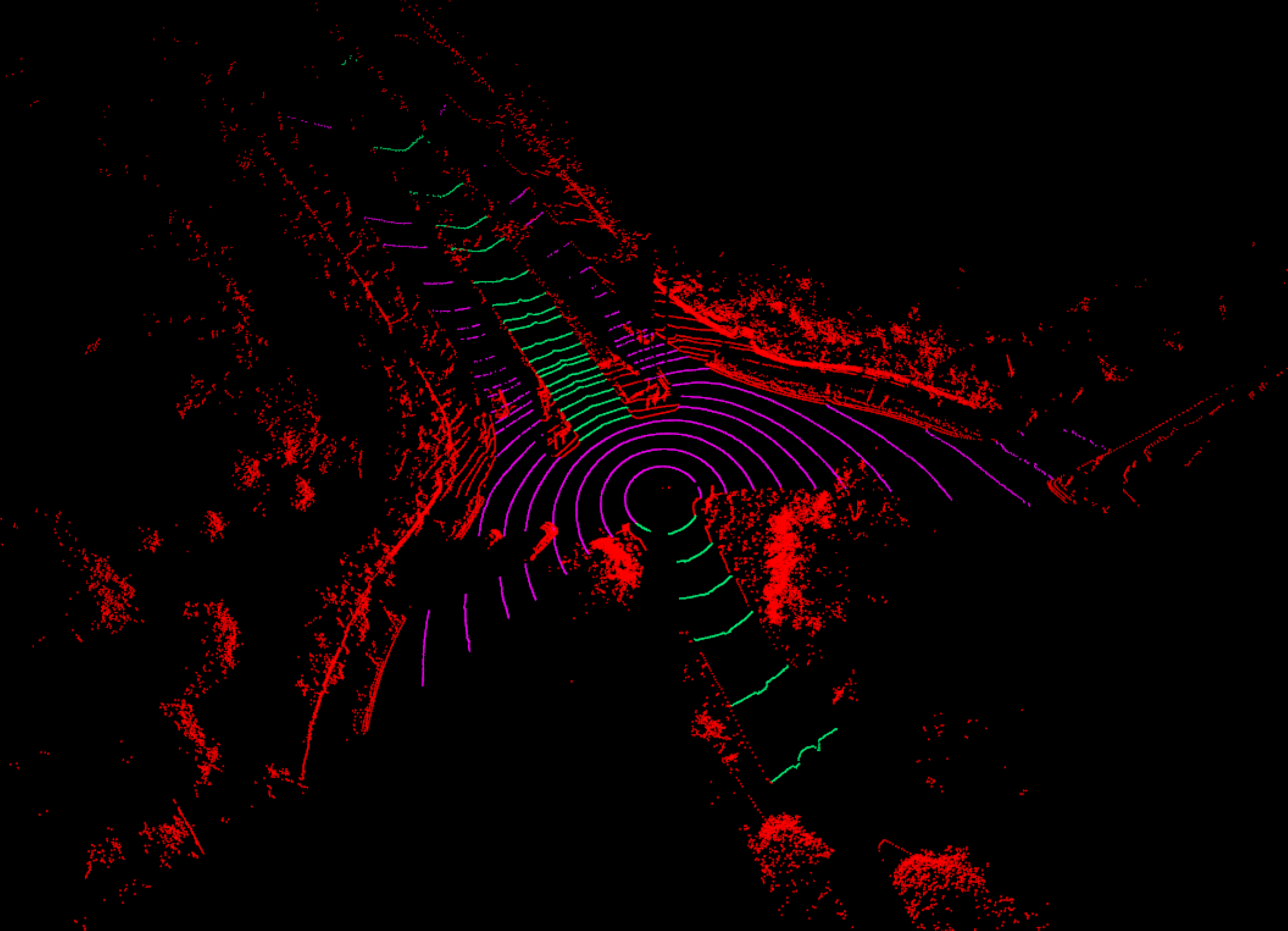}
\caption{Corresponding manually labelled LIDAR scan. Purple points are labelled as \textit{road}, red points as \textit{obstacle}, and green points as \textit{do not care}, since they belong to a reserved bus lane.}
\end{subfigure}
\caption{Example of reserved bus lane in Guyancourt that go through a roundabout}
\label{guyround}
\end{figure*}

\paragraph{}
Though a training and validation dataset were obtained, they would not enable road detection algorithms to be properly evaluated, as the obtained labels are not binary, and may still include errors. An additional test set, that would only be used to evaluate the performances of the road detection system, was thus also created. As no open-source dataset uses a VLP32C LIDAR scanner, this test dataset was specifically labelled by hand for our work. The platform used to collect the data was the ZoeCab system that was used to collect the Renault Technocentre and Paris-Saclay datasets, and was driven in Guyancourt, France. Yet, the LIDAR sensor that was used to collect the test data  was different from the one that was, previously, used to create the automatically labelled dataset. Indeed, we observed that some differences with regards to the returned intensity exist among different VLP32C LIDARs, even if the sensors are properly calibrated. Additionnally, the sensor was put five centimeters lower than the position that was used, previously, to collect the training and validation sets. This test dataset was pre-labelled thanks to the same automatic labelling procedure that was used for the training set, and manually refined afterwards. We chose the Guyancourt area for our test set because it contains road setups that are very different from our training scans: most of the roads in Guyancourt are two-lane one-way roads, while the training scans were mostly acquired on single-lane two-way roads. Moreover, this area numerous roundabouts with very varrying sizes. In the Guyancourt area, the vehicle was driven at up to 70km/h, following local driving rules, as the motion compensation for projection into the map did not need to be as accurate as for the training set. Indeed the automatic labels were refined. In total, the resulting test dataset includes 347 scans. Figure~\ref{test_dataset} presents the locations where the test dataset was collected.
\paragraph{}
A specificity of the Guyancourt area is that there exist reserved lanes for buses that are physically separated from the rest of the road. Those lanes might have unique and particular setups. Figure~\ref{guyround} presents for instance a bus lane that goes through a roundabout, while the other vehicles have to drive in the circle. In Figure~\ref{view1}, it can be observed that the bus lane was separated from the other lanes before the roundabout. In Figure~\ref{view2}, it can be seen that the part of the bus lane that goes through the roundabout does not have the same texture as the other parts of the road. As those bus lanes are very particular, they were labelled as belonging to a \textit{Do not care} class. We considered that classifying those bus lanes as roads was not relevant, but not an error per se, as they are still technically roads. They were not considered in the evaluations done on the Test set. Our training, validation and test sets are made publicly available\footnote{https://datasets.hds.utc.fr -- under the "Automatic and manual LIDAR road labels" project}, alongside metadata that include GPS locations and CAN measurements from the data acquisition platforms.
\raggedbottom

\section{Evaluation of the performances}
\label{evalkittirenault}
The performances of the RoadSeg are mainly evaluated on our manually labelled test set. The KITTI dataset is also used to give some preliminary, though limited, results, as it is one of the most commonly used dataset for road detection.

\subsection{Evaluation on the KITTI Road dataset}
\paragraph{}
Contrary to other approaches that are evaluated on the KITTI road dataset, we evaluate our system at the scan level, and do not aim at predicting the presence of road in areas that are not observed by the LIDAR sensor. No publication to the KITTI road benchmark was then possible. The labels of the KITTI road dataset are only given in an image plane corresponding to a camera whose intrinsic calibration parameters are available, alongside extrinsic calibration parameters with regards to a Velodyne HDL64 LIDAR, which was synchronized with the camera. The image road labels can then be projected into the LIDAR scans, to create ground-truth for road detection in LIDAR scans. Each point is represented by its Cartesian coordinates, its reflectance, the range measured by the LIDAR sensor, and its validity. The scans are then represented as $6\times64\times512$ images.
\begin{figure}[h!]
\centering
\begin{subfigure}{0.5\textwidth}
\includegraphics[width=\textwidth]{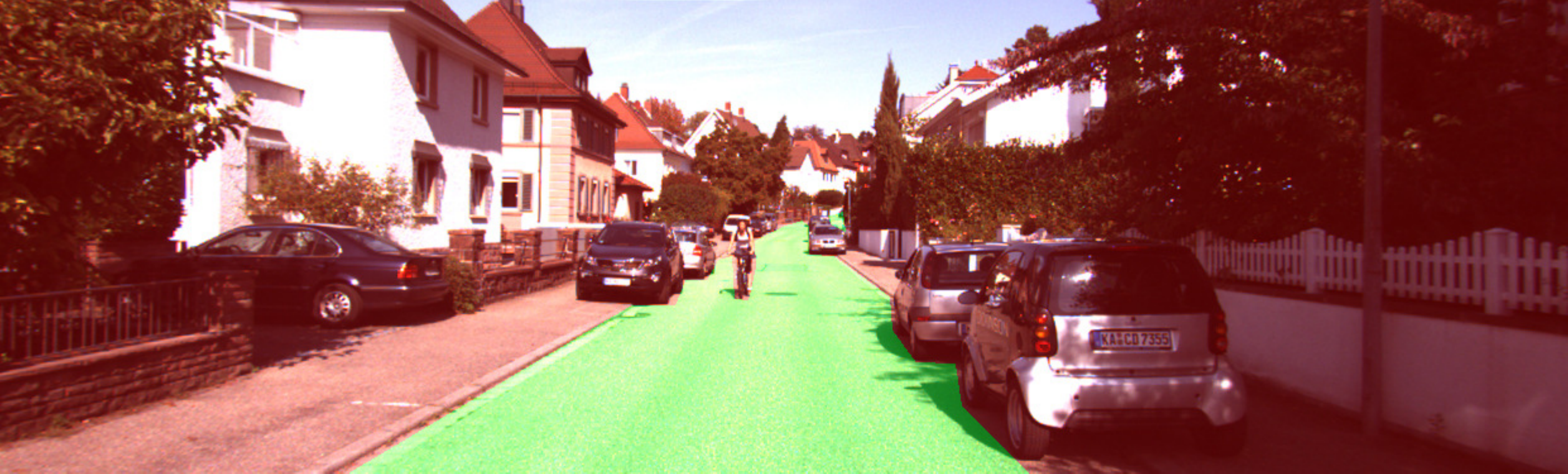}
\caption{Labels from the KITTI road dataset in the image plane}
\end{subfigure}
\begin{subfigure}{0.5\textwidth}
\includegraphics[width=\textwidth]{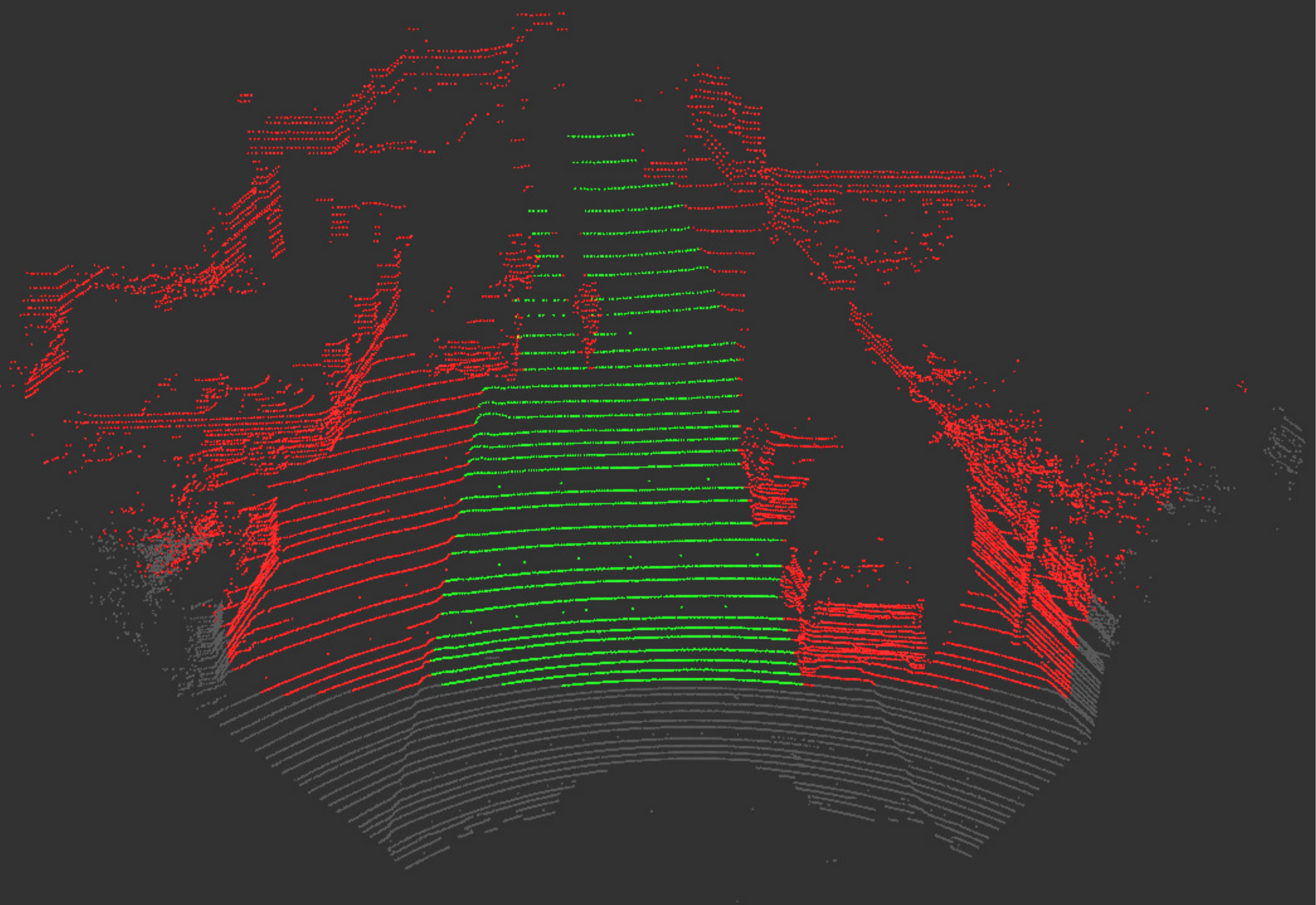}
\caption{Labelled LIDAR scan by projection of the image labels. Points that do not intersect the field-of-view of the camera are not displayed. Green points are labelled as \textit{road}, red points as \textit{not road}, and grey points are \textit{unlabelled}.}
\end{subfigure}
\caption{Generation of LIDAR labels from the KITTI road dataset. }
\label{kittigenlabels}
\end{figure}
\paragraph{}
The TNet of RoadSeg only predicts transformations for the Cartesian coordinates, and leaves the ranges and intensities unchanged. The LIDAR scans can only be labelled partially, because the camera view only covers a section of the corresponding LIDAR scan. Moreover, some LIDAR points can easily be mislabelled. Indeed, the LIDAR labels in the KITTI road dataset are not accurate, especially around road objects, and synchronization and calibration errors between the sensors exist in the KITTI dataset. Figure~\ref{kittigenlabels} displays an example of labels obtained from the KITTI dataset. Short range points are unlabelled, most of 360\degree scan is excluded from the labelling procedures, and the image labels around the cyclist and the parked vehicles are noticeably coarse. Public labels are only available for the 289 scans that are present in training set of the KITTI road dataset. 
\paragraph{}
Given the small number of labelled scans available in the KITTI dataset, a 5-fold cross-validation procedure was used to estimate the performances of RoadSeg. RoadSeg is compared with both SqueezeSegV2, and SqueezeSegV2 without its CRF layer. Each model was trained for 200 epochs on 4 of the folds. The training procedure was the same for the three models, and followed the one originally used by SqueezeSegV2, with a batch size of 16 for all the models. The training was repeated until each model was trained on every possible combination of folds. The 5 folds were the same for each model. After the 200 epochs, the selected parameters for each model correspond to those that maximize the F1-score over the 4 training folds. Only the points that are both valid and labelled are considered in the loss function and by those metrics. We consider that a point, for which a probability of belonging to the road is strictly higher that 0.5, is predicted as being a road point. Finally, average F1, Recall, Precision and IoU scores were computed from the results of each model on the corresponding test folds, and reported in Table~\ref{kittires}. We also report maximum and minimum scores over the 5 test folds, alongside the average execution time of each model on an NVidia TitanX GPU.

\begin{table*}[h]
\centering
\resizebox{\textwidth}{!}{%
\begin{tabular}{|c|ccc|ccc|ccc|ccc|c|}
\hline
    \cellcolor{black} & \multicolumn{3}{|c|}{Precision} & \multicolumn{3}{c|}{Recall} & \multicolumn{3}{c|}{F1-score} & \multicolumn{3}{c|}{IoU} & Inference time  \\
   \hline
   %\multicolumn{14}{c}{}\\
   \hline
   Model & Avg. & Max. & Min. & Avg. & Max. & Min. & Avg. & Max. & Min. & Avg. & Max. & Min. & Mean\\
   \hline
   SqueezeSegV2 without CRF & 0.9267 & 0.9573 & 0.8882 & 0.9108 & 0.9675 & 0.8345 & 0.9177 & 0.9426 & 0.8792 & 0.8487 & 0.8914 & 0.7844 & \textbf{16.3 ms}\\
   SqueezeSegV2 with CRF & 0.9277 & 0.9424 & \textbf{0.8993} & 0.9074 & 0.9635 & 0.8303 & 0.9167 & 0.9459 & 0.8730 & 0.8472 & 0.8974 & 0.7746 & 17.8 ms\\
   RoadSeg & \textbf{0.9411} & \textbf{0.9741} & 0.8253 & \textbf{0.9758} & \textbf{0.9942} & \textbf{0.9649} & \textbf{0.9572} & \textbf{0.9783} & \textbf{0.8909} & \textbf{0.9199} & \textbf{0.9575} & \textbf{0.8033} & 30.8 ms\\
    \hline
\end{tabular}}
\caption{Comparison of the fire layers in RoadSeg and SqueezeSeg}
\label{kittires}
\end{table*}

The amount of data that is available on the KITTI dataset is probably inadequate for proper training, and thus limits the the relevance of those evaluations. The limitations of the projection procedure also prevent us from considering that those results are completely reliable. However, they tend to show that RoadSeg vastly outperforms SqueezeSegV2 for road detection. This however comes with an increased mean inference time. Yet, RoadSeg still processes LIDAR scans at a high rate, with a mean inference time of 30.8 ms. We also observe that the CRF layer of SqueezeSegV2 seems not to improve the performances in road detection. This may mean that using a CRF alongside SqueezeSegV2 may require other hyperparameters, or kernels, for road detection. This could also mean that the CRF layer struggles with large objects, such as the road, as it aims at locally smoothing the segmentation results. More reliable results are given in the next section, as the models are evaluated on our manually labelled dataset, after having been trained on the automatically labelled dataset that was generated from our HD Maps.
%%%%%%%%%%%%%%%%%%

\subsection{Evaluation on the manually labelled Guyancourt dataset}
\subsubsection{Classification performances of single networks}
To give more significant results, we report metrics on our manually labelled test set. All the approaches were only trained thanks to the 1500 automatically labelled LIDAR scans that were recorded in Compiègne, Renault Technocentre, Rambouillet and the Paris-Saclay Campus, and thus have never been trained on the area where the manually labelled data was recorded. The training/validation split is exactly the one presented in Figure~\ref{trainset}, and all the approaches were trained with the same procedure, which is close to the one used for the KITTI dataset, except that batch size was reduced to 10 for all the approaches. This was needed because, while the scans from the KITTI dataset only cover the front view, our training set is composed of 360\degree scans. This also justified to modify the behavior regarding the padding of the feature maps processed by the networks. Indeed, the left and right sides of the inputs actually correspond to neighboring areas. The left (respectively right) padding is then obtained by mirroring the right (respectively left) side. Tests on several variants are presented. First of all, SqueezeSegV2, SqueezeSegV2 without CRF, RoadSeg, and RoadSeg without TNet have been trained on the whole automatically labelled dataset, and each point was represented by the eight available features (Cartesian coordinates, spherical coordinates, intensity and validity). We also trained four additional and lighter networks.

\begin{itemize}
\item RoadSeg-Intensity: the points are only represented by their intensity, elevation angle, and validity. 
\item RoadSeg-Spherical: the points are only represented by their spherical corrdinates, and their validity
\item RoadSeg-Cartesian: the points are only represented by their Cartesian corrdinates, and their validity
\item RoadSeg-Cartesian without TNet: like RoadSeg-Cartesian, but without the TNet
\end{itemize}

RoadSeg-Intensity and RoadSeg-Spherical do not include TNets. Indeed, TNets normally predict an affine transformation matrix for Cartesian coordinates. Yet, in the case of RoadSeg-Intensity and RoadSeg-Spherical, only spherical features are used. A spherical transformation could be predicted by a modified TNet. Yet, since the elevation and azimuth angles are unique for each position in the range image, such a transformation would be equivalent to an image deformation, which could complexify the further convolutions used in RoadSeg. RoadSeg-Intensity processes the elevation angle, alongside the intensity, to cope with the individual behavior of each LIDAR laser. Indeed, the intensity might be inconsistent among each LIDAR laser. Each model was trained ten times on the training set, to account for the possible variability in the results due to the random initialization of the network. At each training session, the parameters that were kept were those that maximize the F1-score on the validation set. We again report, in Table~\ref{testsetres}, average F1, Recall, Precision and IoU scores reached by all the models on the test set. We also report maximum and minimum scores alongside the average execution time of each model on an NVidia TitanX GPU. As the test set is manually labelled, and considered as reliable, it is a proper way to evaluate what was learnt by the networks from the automatically labelled dataset.

\begin{table*}[h]
\centering
\resizebox{\textwidth}{!}{%
\begin{tabular}{|c|ccc|ccc|ccc|ccc|c|}
\hline
    \cellcolor{black} & \multicolumn{3}{c|}{Precision} & \multicolumn{3}{c|}{Recall} & \multicolumn{3}{c|}{F1-score} & \multicolumn{3}{c|}{IoU} & Inference time  \\
   \hline
   %\multicolumn{14}{c}{}\\
   \hline
   Model & Avg. & Max. & Min. & Avg. & Max. & Min. & Avg. & Max. & Min. & Avg. & Max. & Min. & Mean\\
   \hline
   SqueezeSegV2 without CRF & \textbf{0.8799} & \textbf{0.9097} & \textbf{0.8597} & 0.7982 & 0.8269 & 0.7776 & 0.8363 & 0.8590 & 0.7952 & 0.7190 & 0.7312 & 0.6601 & 12,3 ms\\
   SqueezeSegV2 with CRF & 0.8606 & 0.8989 & 0.8319 & 0.7920 & 0.8059 & 0.7801 & 0.8247 & 0.8352 & 0.8052 & 0.7018 & 0.7171 & 0.6739 & 14,1 ms\\
   RoadSeg with TNet & 0.8561 & 0.8803 & 0.8199 & 0.8587 & 0.9097 & 0.7979 & 0.8561 & 0.8796 & 0.8355 & 0.7493 & 0.7851 & 0.6463 & 33,4 ms\\
   RoadSeg without TNet & 0.8268 & 0.8547 & 0.7944 & 0.8902 & \textbf{0.9316} & \textbf{0.8668} & 0.8570 & 0.8750 & 0.8426 & 0.7500 & 0.7777 & 0.7107 & 11,7 ms \\
   RoadSeg-Intensity & 0.7687 & 0.8014 & 0.7471 & 0.8076 & 0.8865 & 0.7653 & 0.7870 & 0.8108 & 0.7526 & 0.6491 & 0.6818 & 0.6033 & \textbf{11.5} ms\\
   RoadSeg-Spherical & 0.8565 & 0.8968 & 0.8253 & 0.8776 & 0.9045 & 0.8414 & 0.8665 & 0.8830 & 0.8497 & 0.7645 & 0.7409 & \textbf{0.7899} & 11.6 ms\\
   RoadSeg-Cartesian with TNet & 0.8528 & 0.8911 & 0.8160 & \textbf{0.8927} & 0.9159 & 0.8606 & 0.8718 & \textbf{0.8874} & 0.8594 & 0.7729 & \textbf{0.7976} & 0.7543 & 33,1 ms\\
   RoadSeg-Cartesian without TNet & 0.8674 & 0.8900 & 0.8317 & 0.8852 & 0.9178 & 0.8575 & \textbf{0.8758} & 0.8865 & \textbf{0.8659} & \textbf{0.7791} & 0.7961 & 0.7635 & 11,6 ms \\
    \hline
\end{tabular}}
\caption{Comparison variants of RoadSeg and SqueezeSegV2}
\label{testsetres}
\end{table*}

Interestingly, SqueezeSegV2 without CRF seems to have the best precision on the test set. Yet, this is compensated by its poor performances in terms of recall. We can assume that this is mainly due to the additional subsampling that it uses with regards to RoadSeg: SqueezeSegV2 cannot properly process points at long-range, and considers a significant amount of remote road points as obstacles. We again observe that the use of the CRF degrades the performances for road detection. RoadSeg, when used without the TNet and trained on the full set of features, outperforms SqueezeSegV2, while being slightly faster. The use of the TNet does not seem relevant when training on the full set of features, as it doubles the inference time while slightly degrading the performances, except for the precision scores. Another interesting result is that RoadSeg-Spherical and both version of RoadSeg-Cartesian outperform the networks that are trained on the full set of features. The best F1-score and IoU are even reached by a specific instance of RoadSeg-Cartesian that relies on a TNet. However, RoadSeg-Cartesian without TNet is faster, and easier to train since it has best F1-score and IoU in average. This network thus seems to be the best trade-off for road-detection. However, the performances reached so far are still relatively low. This can be explained by the automatic labelling procedure that we used to generate the train set, as it is very sensitive to unmapped roads, localization errors and improper obstacle filtering. These results also highlight the difficulties for RoadSeg and SqueezeSeg to process numerous features for a given point, as the best performing approaches only process a limited number of features. Fusing the results from several networks, that process different sets of features, however lead to significant improvements. 

\subsubsection{Fusion of neural networks}
\begin{figure*}
\begin{@twocolumnfalse}
\captionsetup{justification=centering}
\centering
\begin{subfigure}{0.32\linewidth}
\centering
\includegraphics[width=\linewidth]{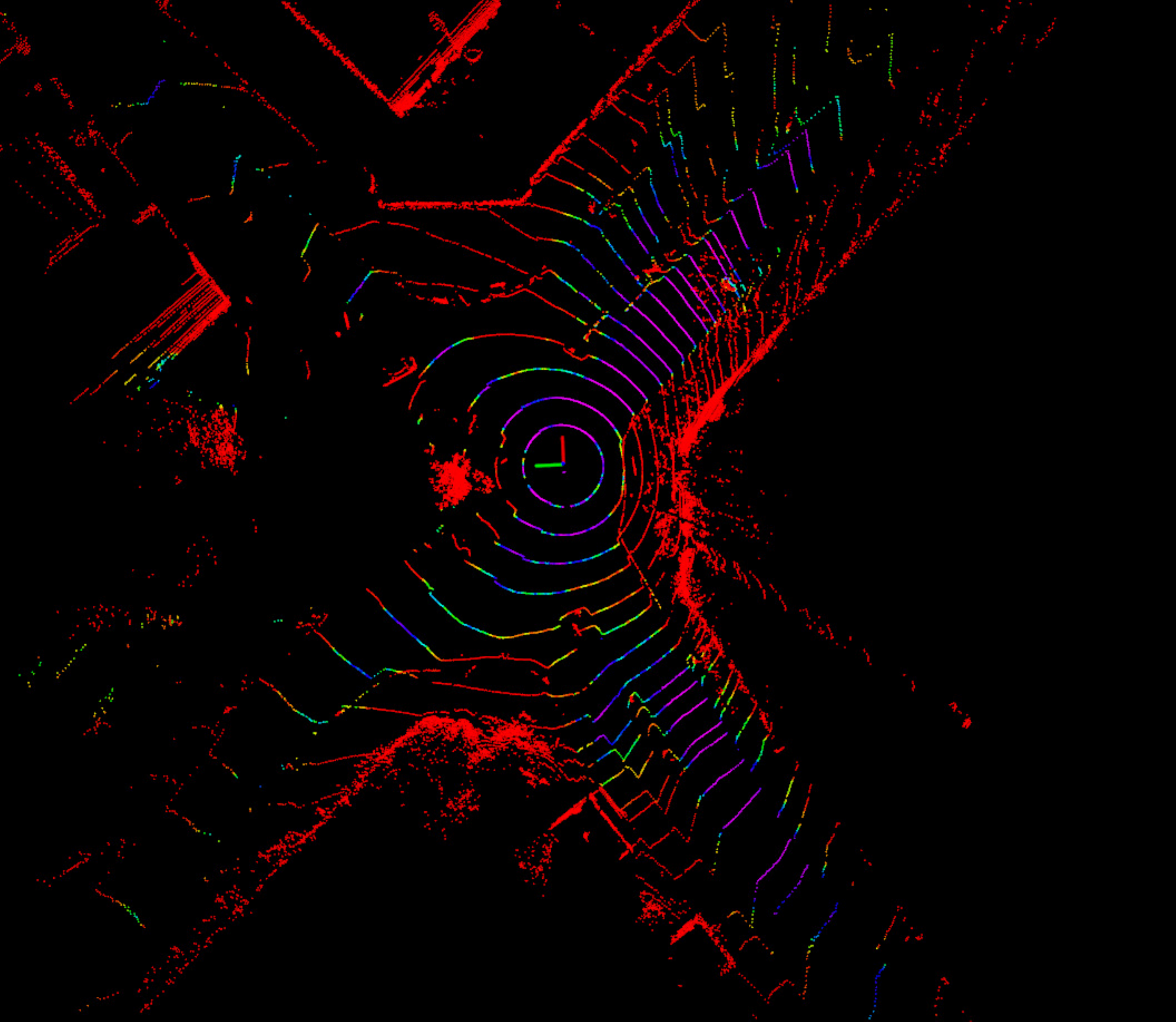}
\caption{$m(\{R\})$, RoadSeg-Intensity}
\end{subfigure}
\begin{subfigure}{0.32\linewidth}
\centering
\includegraphics[width=\linewidth]{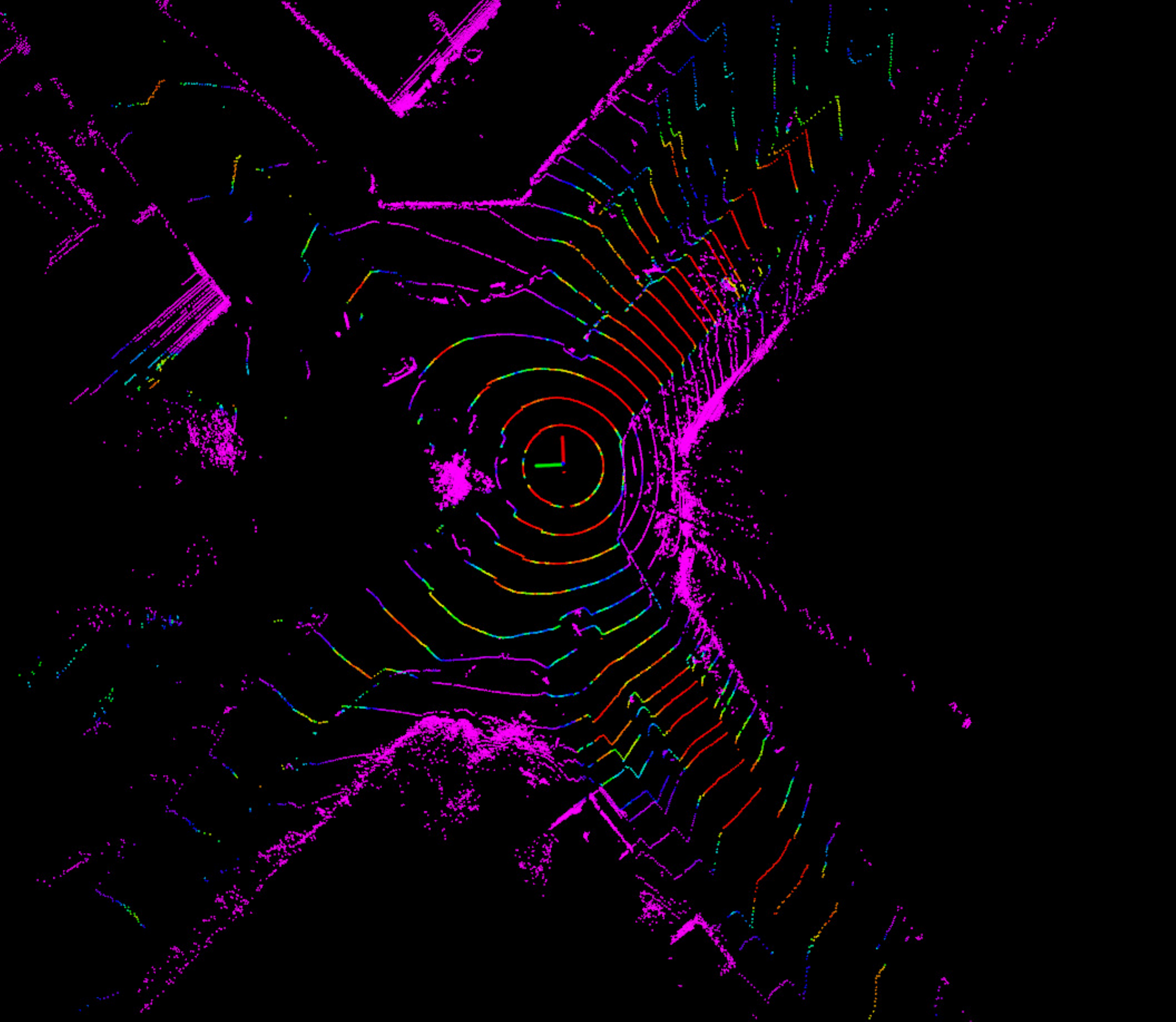}
\caption{$m(\{\neg R\})$, RoadSeg-Intensity}
\end{subfigure}
\begin{subfigure}{0.32\linewidth}
\centering
\includegraphics[width=\linewidth]{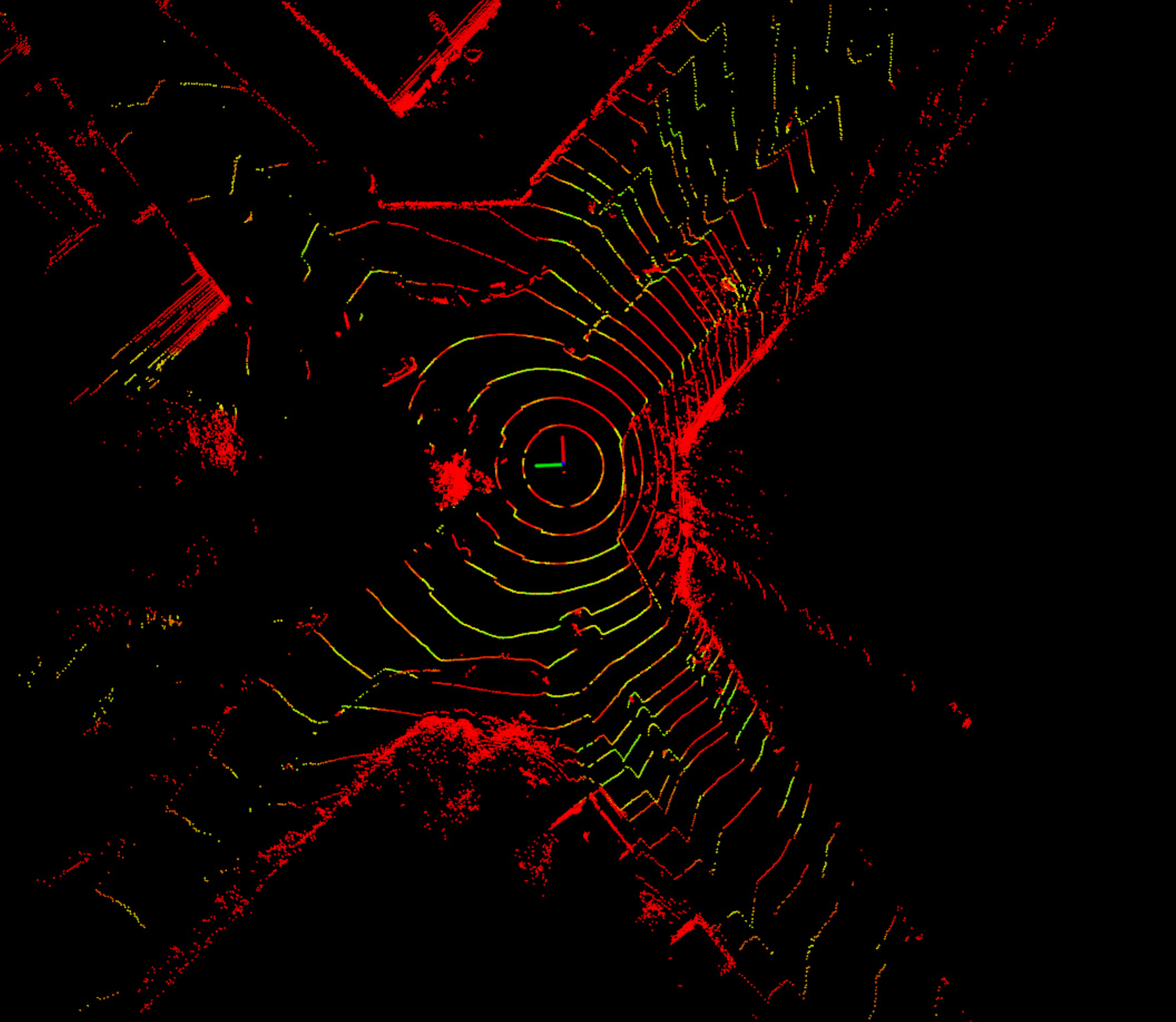}
\caption{$m(\{\Omega\})$, RoadSeg-Intensity}
\end{subfigure}
\begin{subfigure}{0.32\linewidth}
\centering
\includegraphics[width=\linewidth]{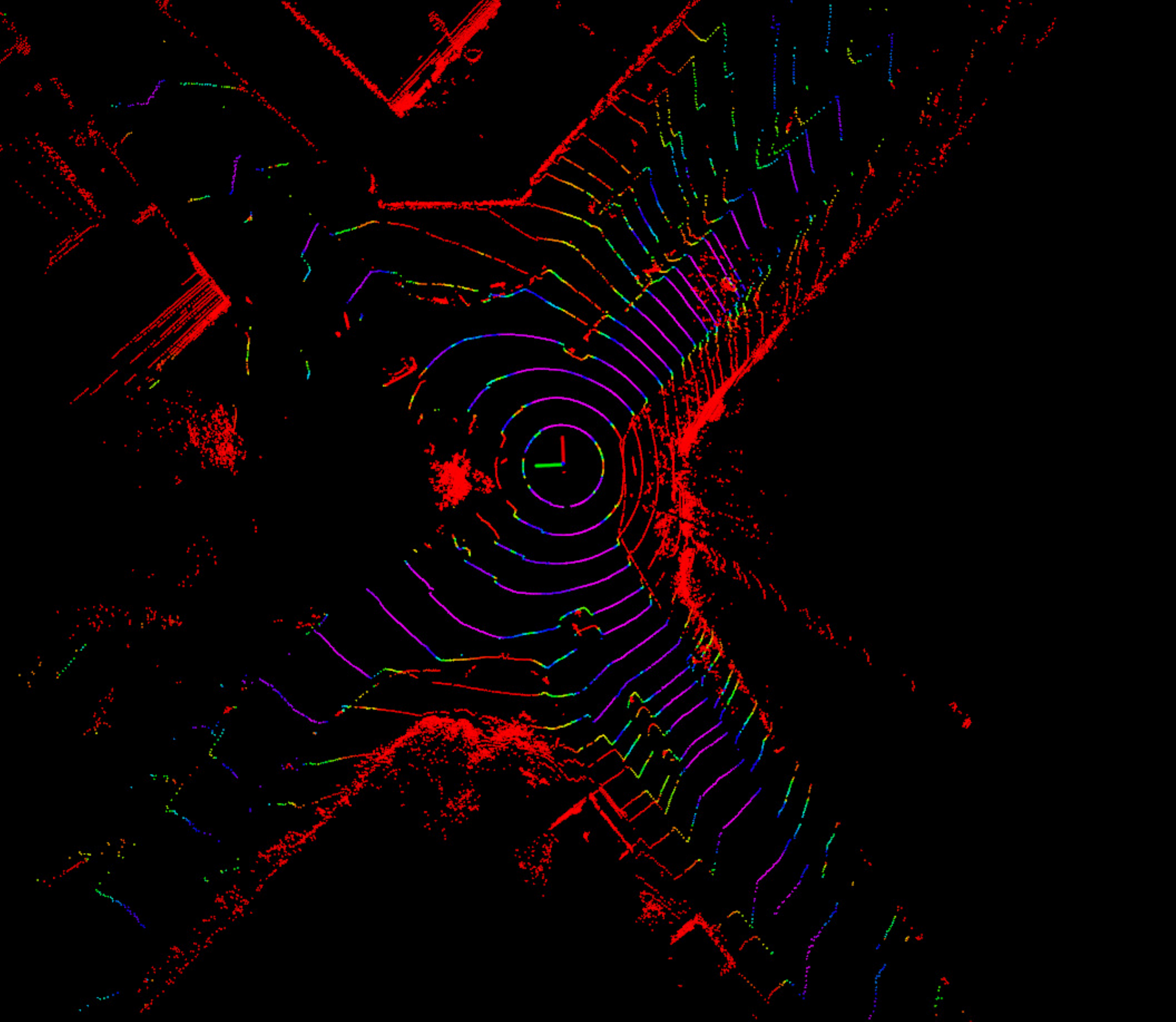}
\caption{$m(\{R\})$, RoadSeg-Spherical}
\end{subfigure}
\begin{subfigure}{0.32\linewidth}
\centering
\includegraphics[width=\linewidth]{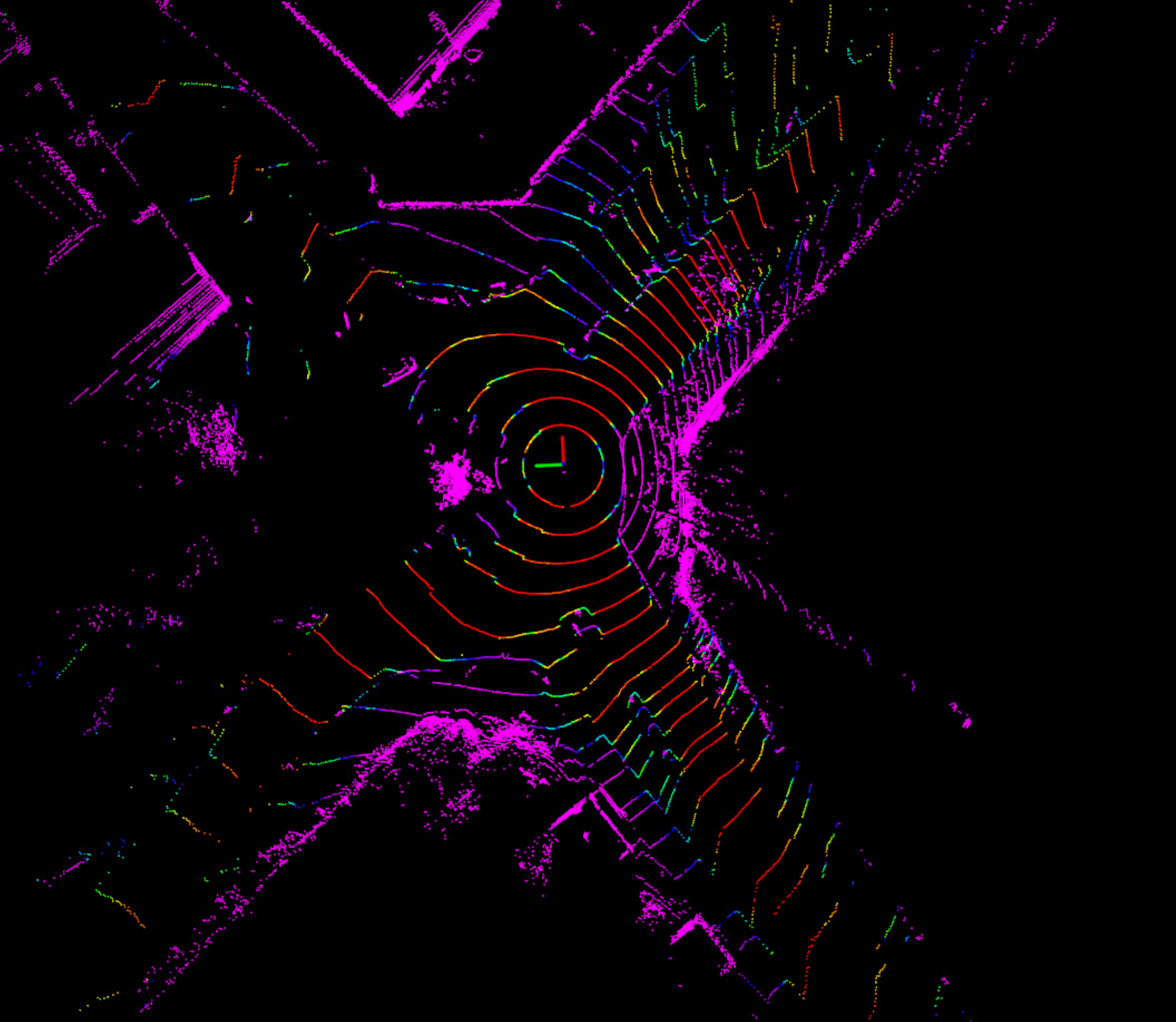}
\caption{$m(\{\neg R\})$, RoadSeg-Cartesian}
\end{subfigure}
\begin{subfigure}{0.32\linewidth}
\centering
\includegraphics[width=\linewidth]{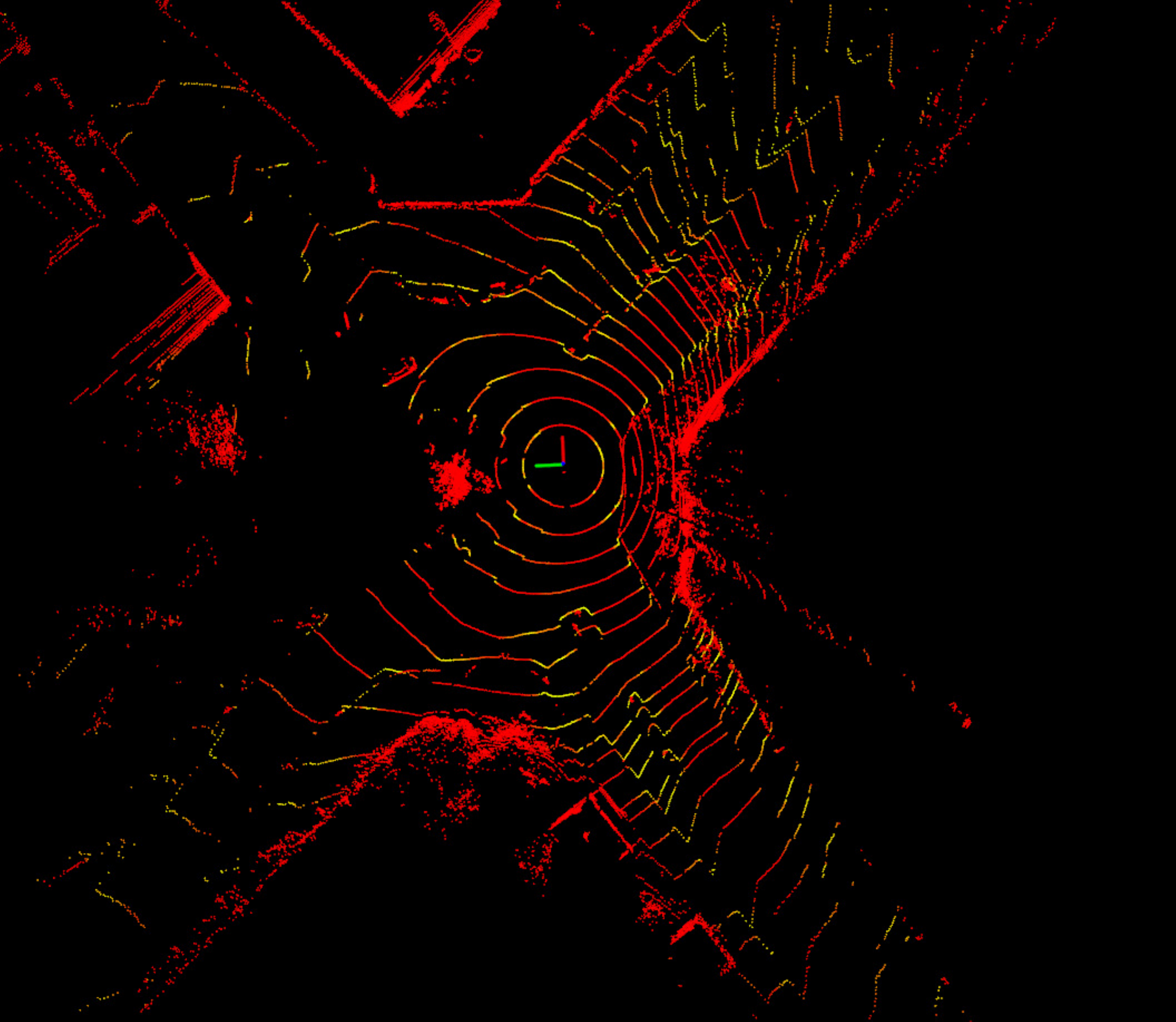}
\caption{$m(\{\Omega\})$, RoadSeg-Spherical}
\end{subfigure}
\begin{subfigure}{0.32\linewidth}
\centering
\includegraphics[width=\linewidth]{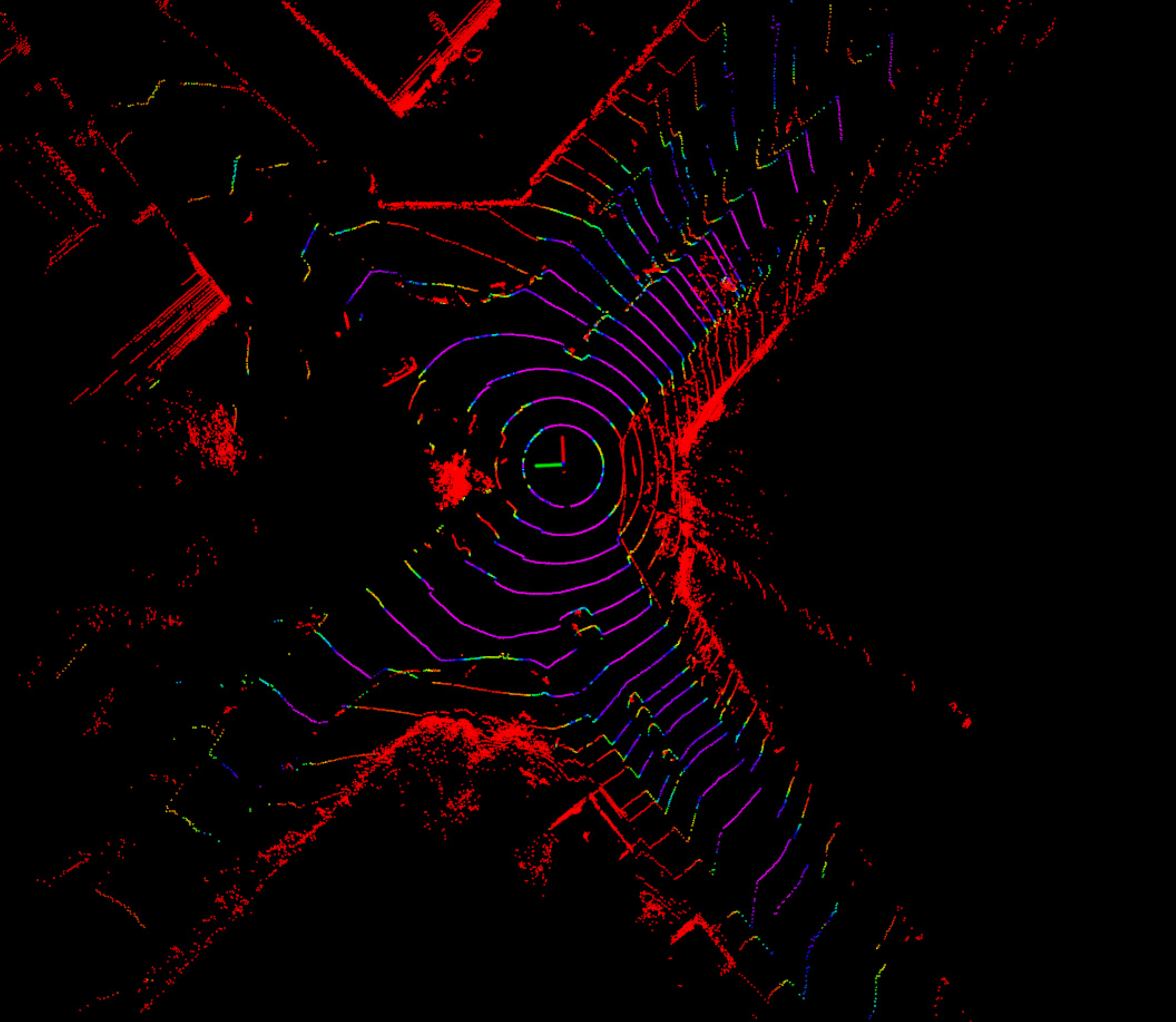}
\caption{$m(\{R\})$, RoadSeg-Cartesian with TNet}
\end{subfigure}
\begin{subfigure}{0.32\linewidth}
\includegraphics[width=\linewidth]{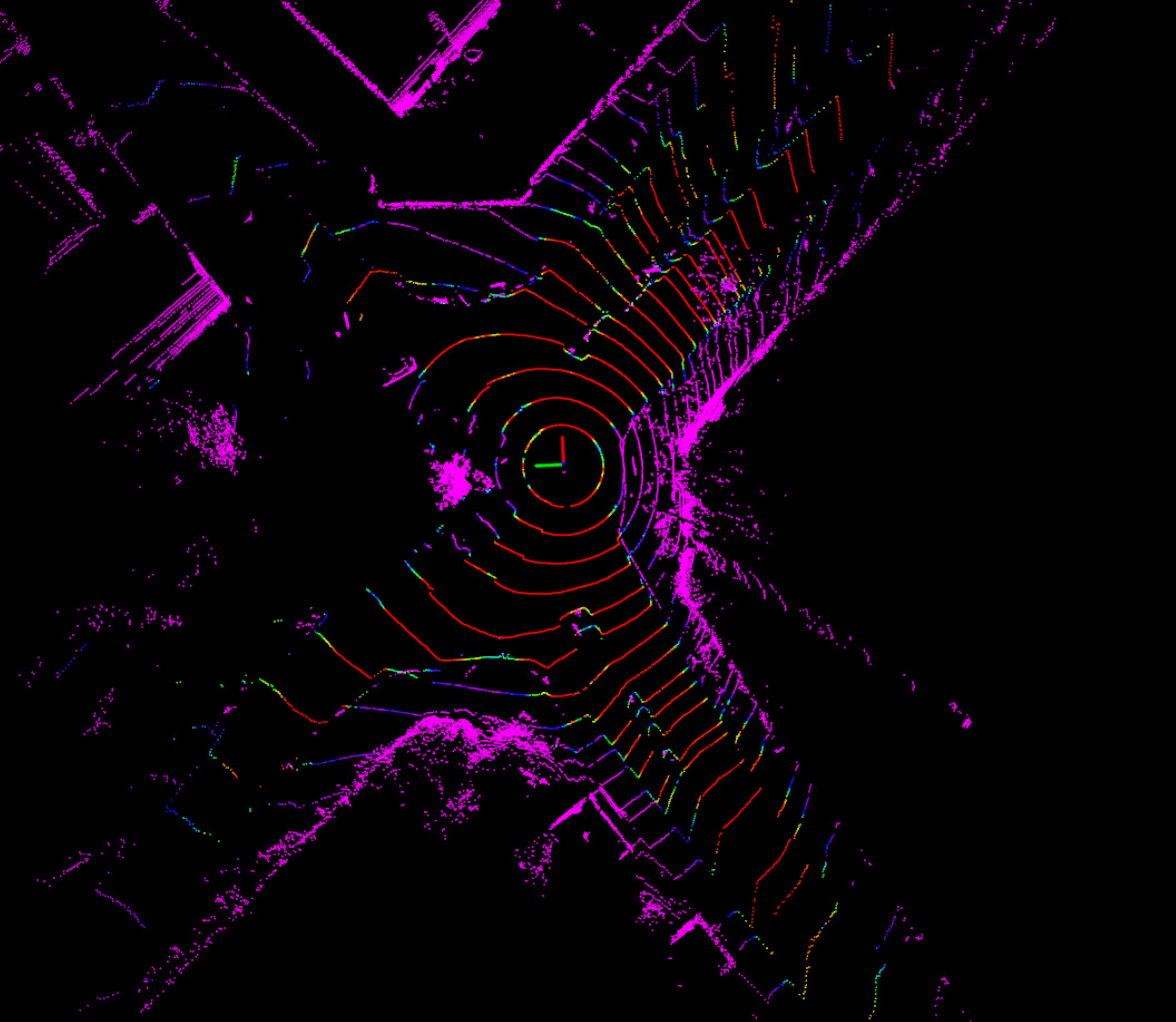}
\caption{$m(\{\neg R\})$, RoadSeg-Cartesian with TNet}
\end{subfigure}
\begin{subfigure}{0.32\linewidth}
\centering
\includegraphics[width=\linewidth]{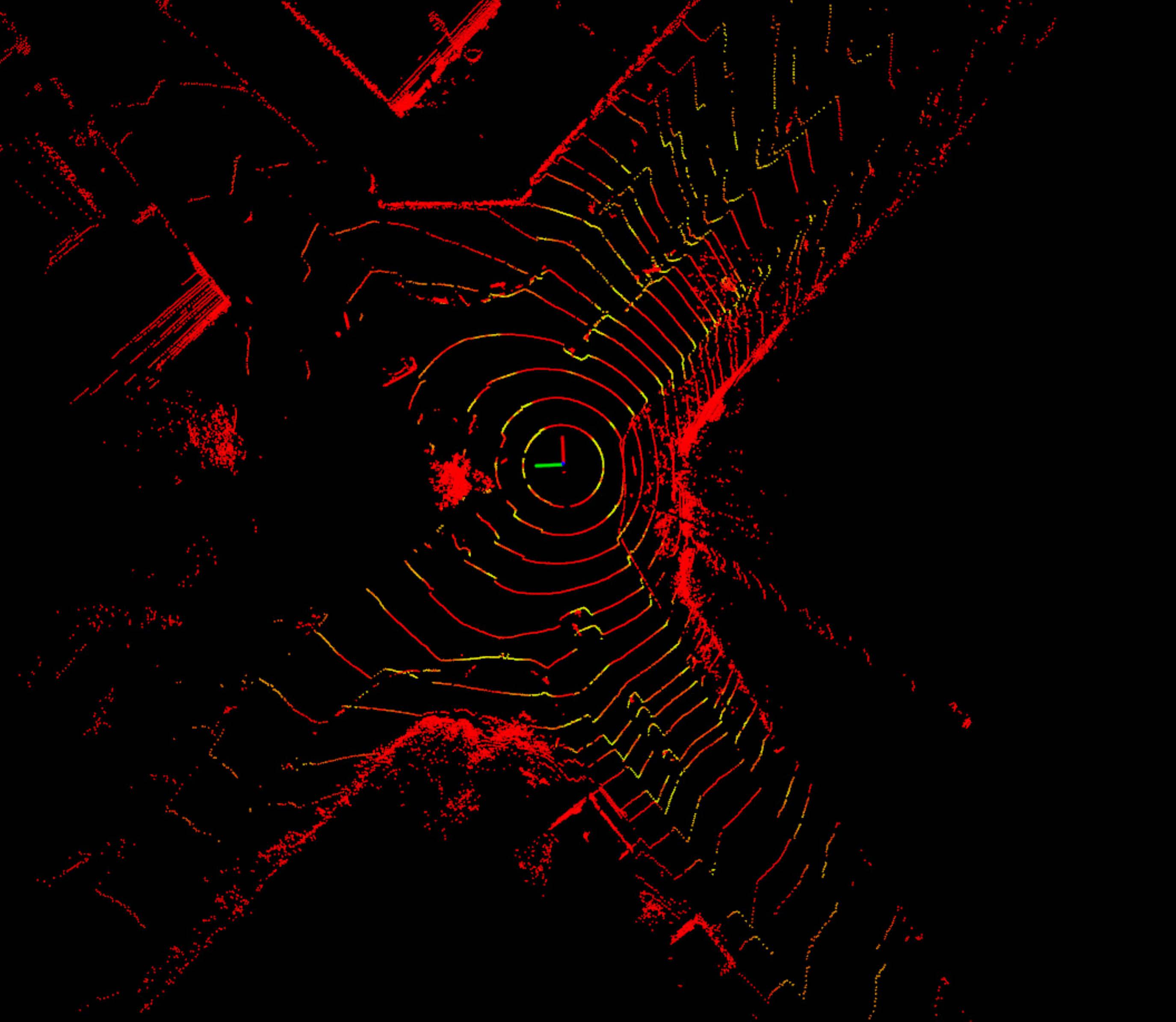}
\caption{$m(\{\Omega\})$, RoadSeg-Cartesian with TNet}
\end{subfigure}
\begin{subfigure}{0.32\linewidth}
\centering
\includegraphics[width=\linewidth]{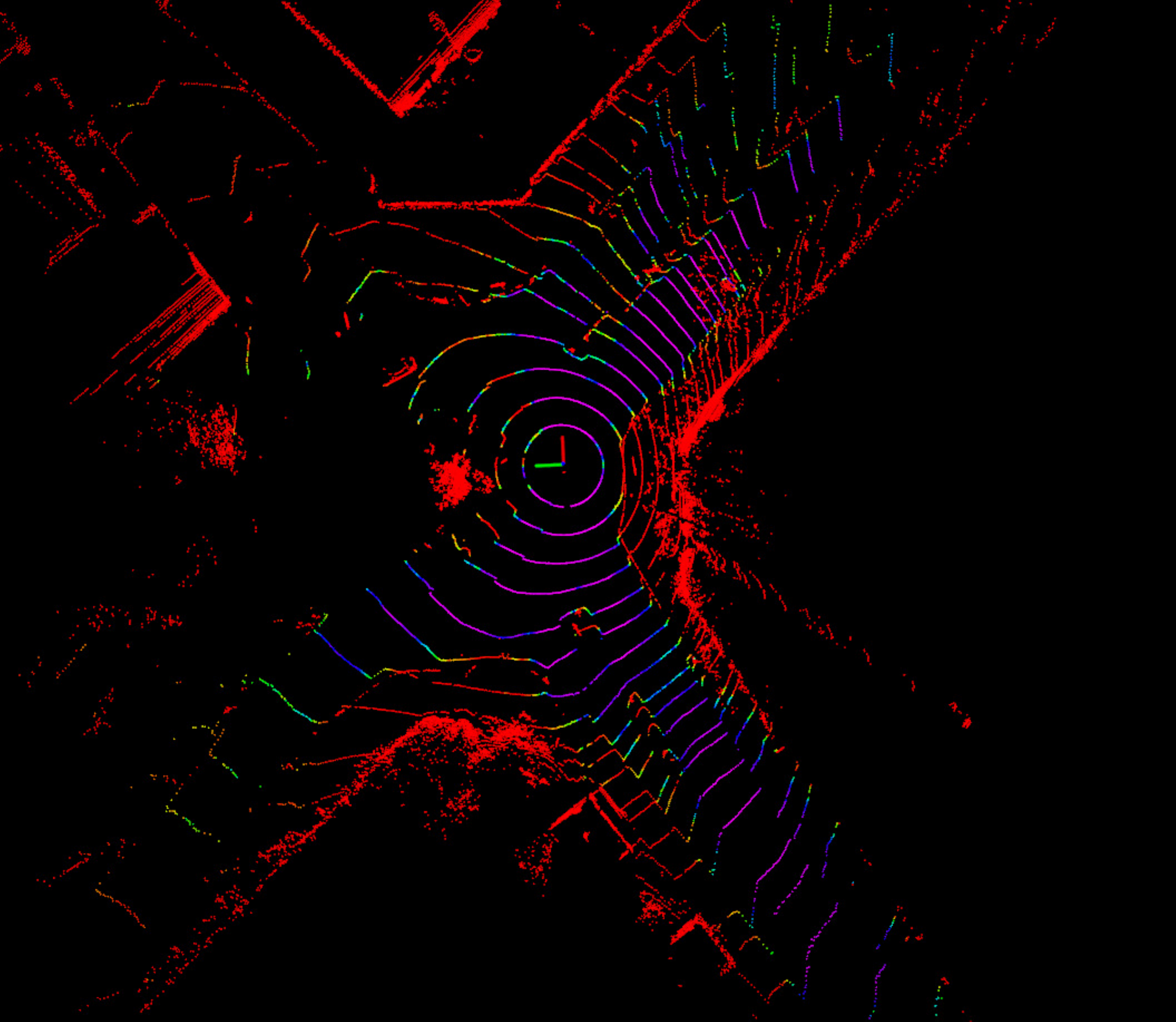}
\caption{$m(\{R\})$, RoadSeg-Cartesian without TNet}
\end{subfigure}
\begin{subfigure}{0.32\linewidth}
\centering
\includegraphics[width=\linewidth]{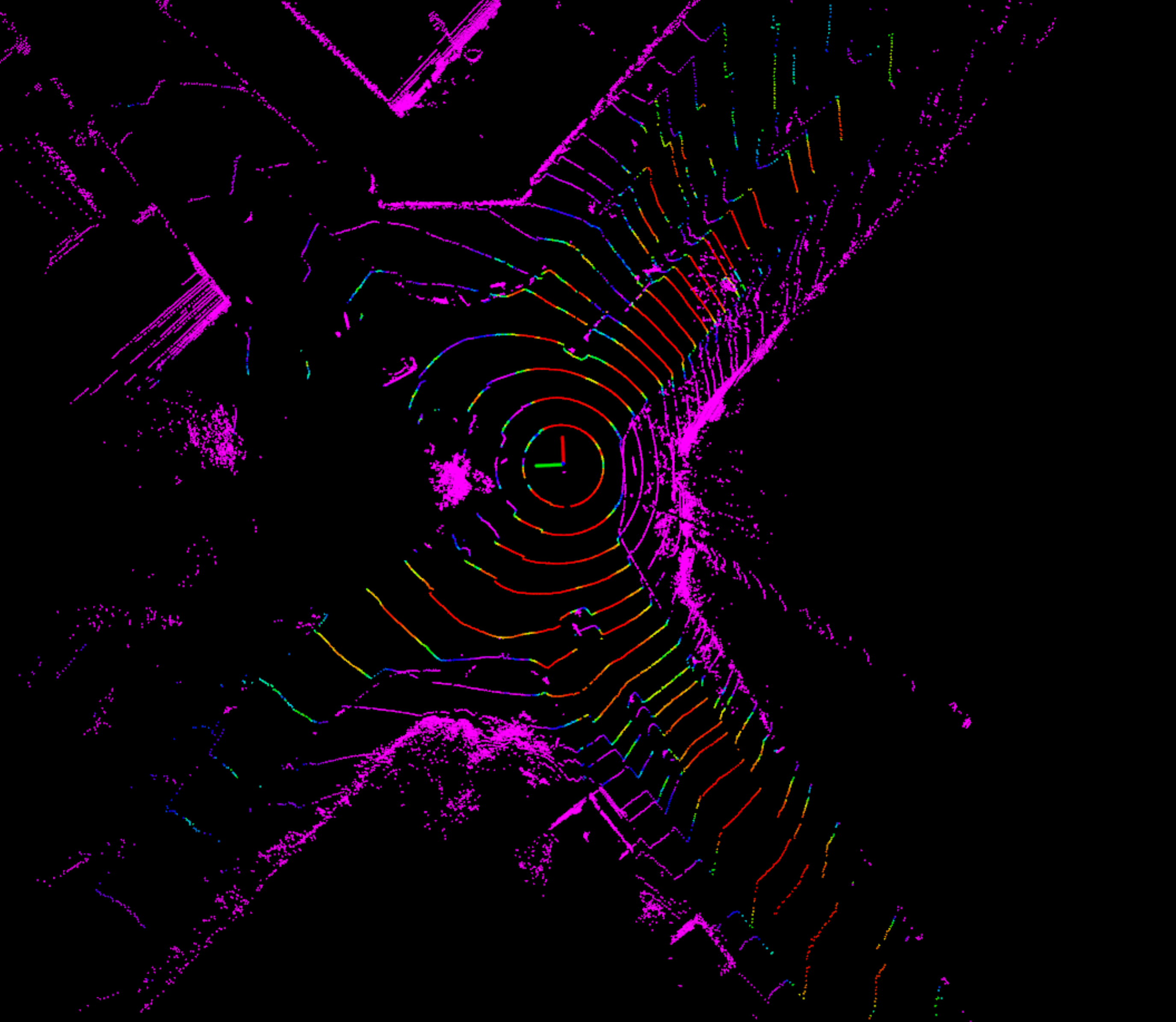}
\caption{$m(\{\Omega\})$, RoadSeg-Cartesian without TNet}
\end{subfigure}
\begin{subfigure}{0.32\linewidth}
\centering
\includegraphics[width=\linewidth]{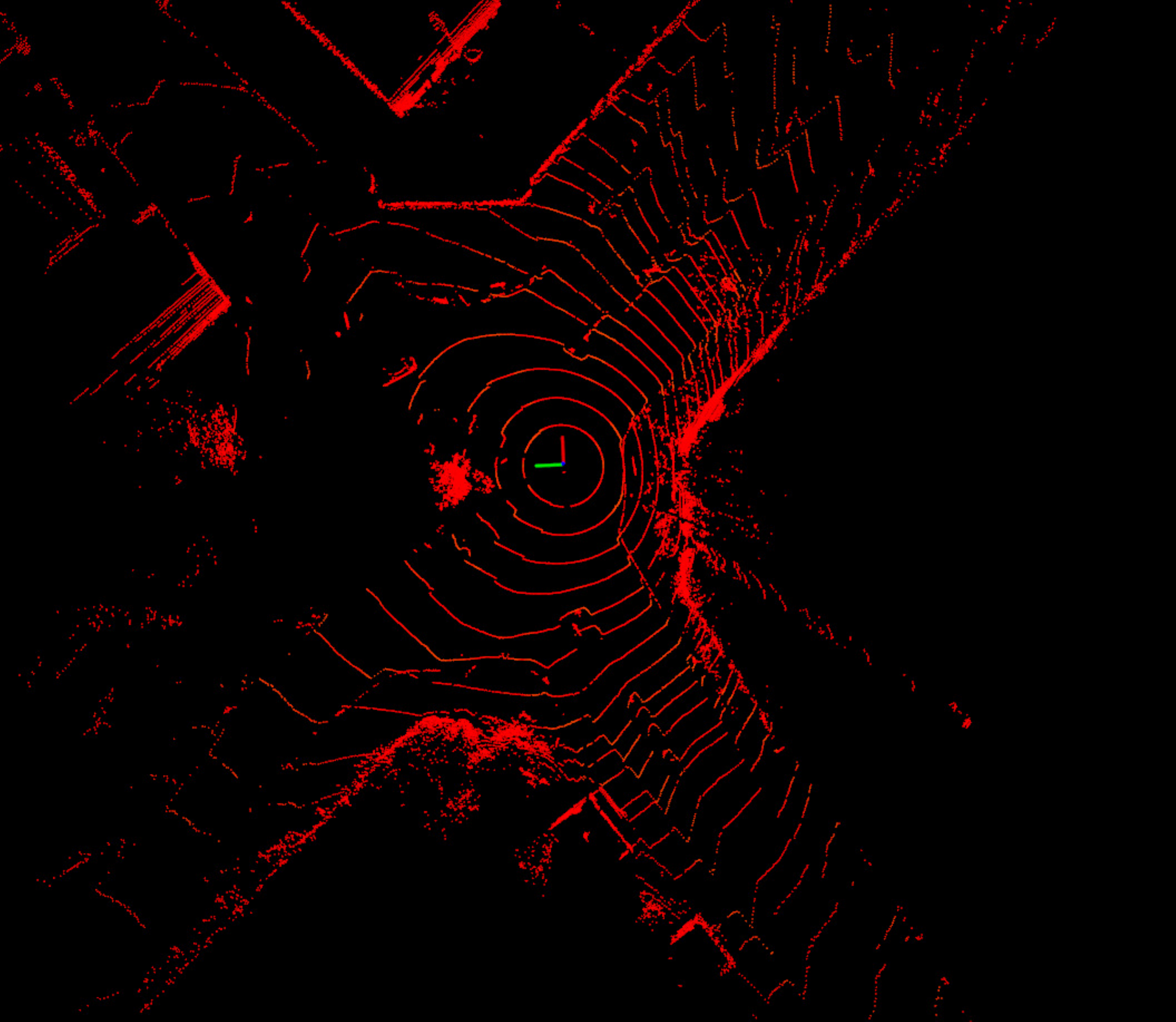}
\caption{$m(\{\Omega\})$, RoadSeg-Cartesian without TNet}
\end{subfigure}
\begin{subfigure}{\linewidth}
\centering
\includegraphics[width=0.32\linewidth]{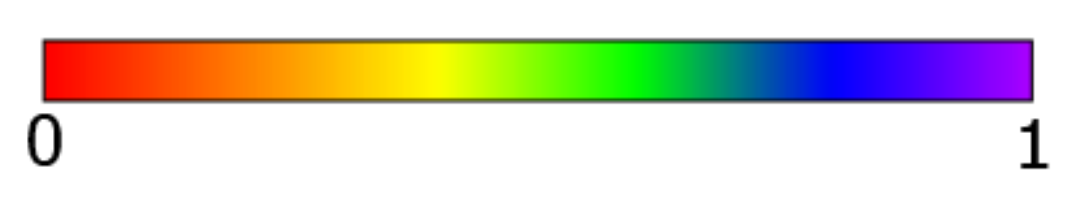}
\end{subfigure}
\captionsetup{justification=raggedright,singlelinecheck=false}
\caption{Example of evidential mass values, obtained from the weights optimized after the training, from all the variants of RoadSeg that can be fused together. Better seen in color, and by zooming in the electronic version of the article.}
\label{exempleevimass}
\end{@twocolumnfalse}
\end{figure*}

\paragraph{}
RoadSeg has been designed to allow for the generation of evidential mass functions from its outputs. A straighforward way to fuse several RoadSeg-like networks could then be to rely on an evidential fusion. We thus propose to use the model described in Equation~\ref{mlrfull} to generate evidential mass functions for each LIDAR point, from a set of neural networks. The resulting mass functions can then be fused, for each point, thanks to Dempster's rule of combination that is described in Equation~\ref{dsoperator}. Figure~\ref{exempleevimass} displays exemples of evidential mass functions that can be obtained from all the variants of RoadSeg that are expected to be independent.  Those mass functions are directly obtained after the training, from the biases of the final Instance-Normalization layer, and without any optimization of the $\alpha$ vector.

The plausibility transformation in Equation~\ref{plaustransfo} can then be used to compute the fused probability that each points belongs to the road, and the results can then be evaluated again on our test set. Given to the properties of plausibility transformation~\cite{cobb2006plausibility}, fusing the evidential mass functions obtained from the different networks via Dempster's rule, and then applying plausibility transformation on the resulting mass function, is equivalent to a Bayesian independent opinion poll fusion among the probabilities obtained from each network. This means that a classification which purely relies on a Bayesian interpretation of the probabilities, obtained from the networks, only depends on the sum of the values of the $\alpha$ vector in Equation~\ref{wjold}, which is supposed to be equal for all the possible $\alpha$ vectors. As we have chosen to consider that LIDAR points are classified as belonging to the road when the probabilities that are predicted by the network are higher than 0.5, we are in this case. We thus report in Table~\ref{fusiontestsetres} recomputed F1-score, Precision, Recall and IoU obtained after the fusion of different sets of networks. To prevent data incest, only RoadSeg-Intensity, RoadSeg-Spherical, RoadSeg-Cartesian and RoadSeg-Cartesian without TNet are considered. We also do not fuse RoadSeg-Cartesian and RoadSeg-Cartesian without TNet together, again to prevent data incest. Since the networks are then not trained jointly, a set of parameters has to be chosen for each one, among the ten that are available. For a fair comparison, the parameters that were selected for each network correspond to the best ones in terms of F1-score on the validation set. Indeed, they are not necessarily the best performing ones on the test set. We thus only report one Precision, Recall, F1-score and IoU for each combination.

\begin{figure*}[h!]
\centering
\begin{subfigure}{.5\textwidth}
  \centering
  \includegraphics[width=0.8\linewidth]{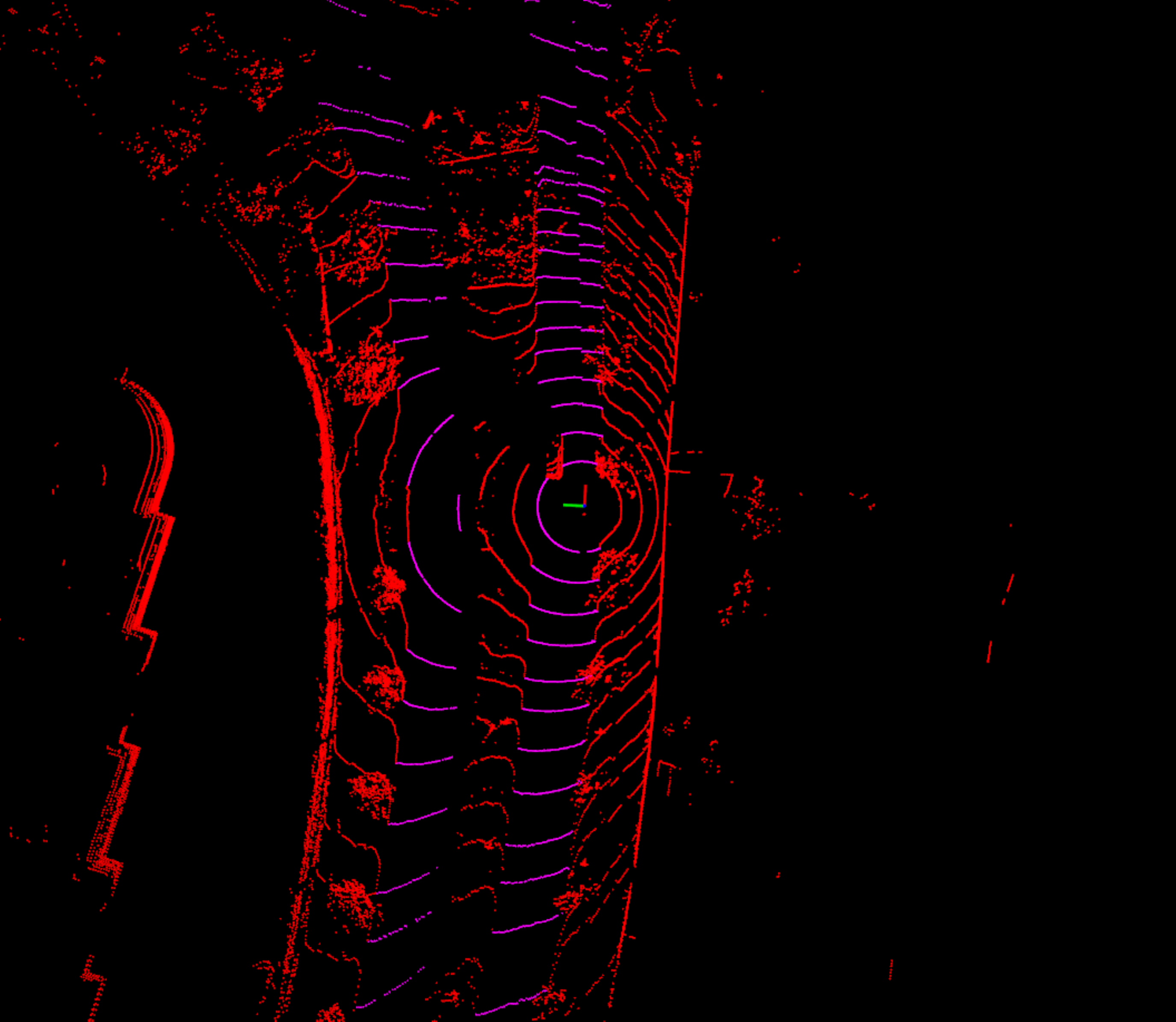}
  \caption{Original test scan}
  %\label{fig:sub1}
\end{subfigure}%
\begin{subfigure}{.5\textwidth}
  \centering
  \includegraphics[width=0.8\linewidth]{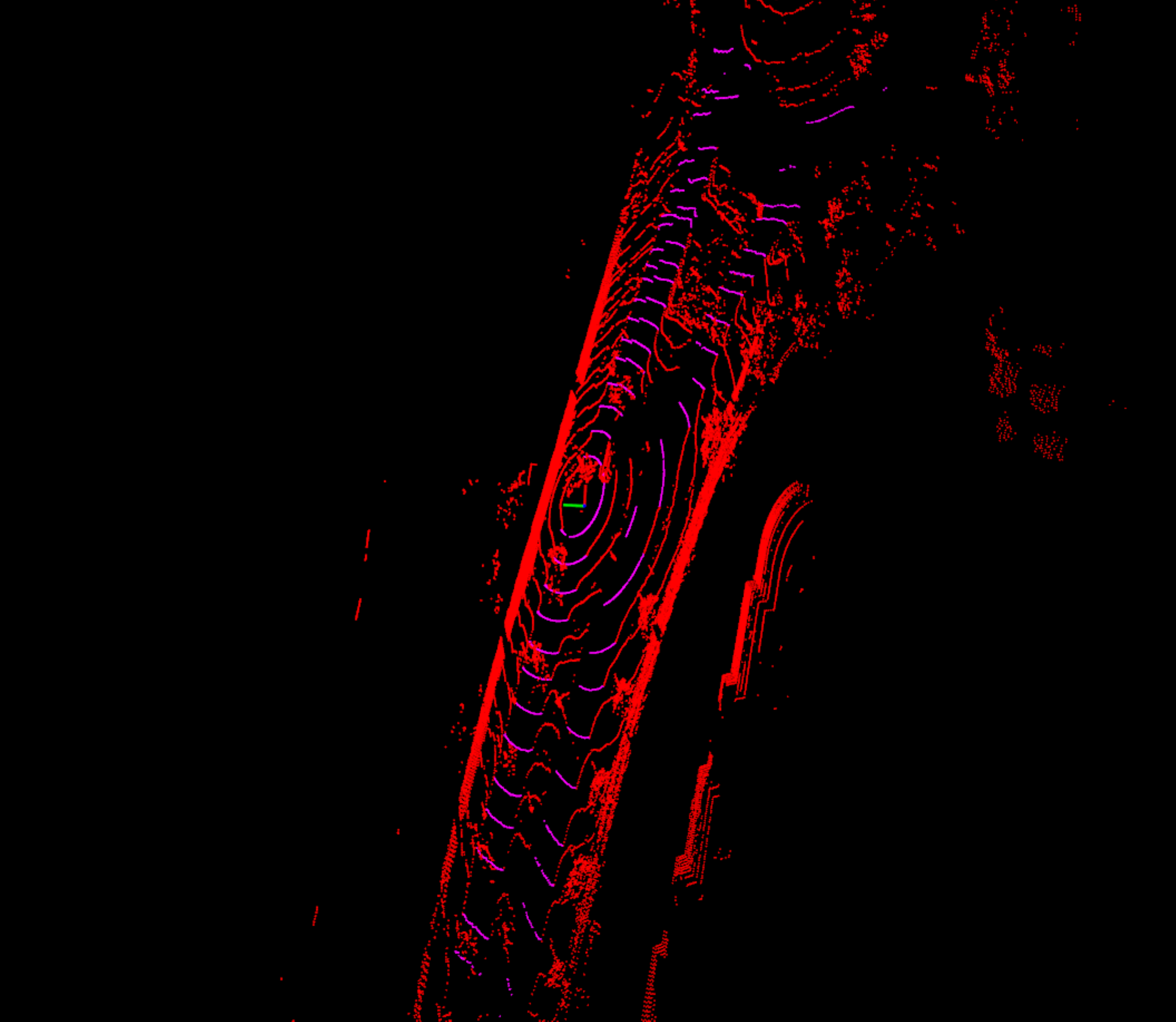}
  \caption{Test scan transformed by the TNet}
  %\label{fig:sub2}
\end{subfigure}
\caption{Effect of the TNet on a LIDAR scan. Purple points were manually labelled as road points, red ones as not belonging to the road. The two pictures correspond to the same field of view and scale.}
\label{tneteffect}
\end{figure*}

\begin{table*}[h]
\centering
\resizebox{\textwidth}{!}{%
\begin{tabular}{|c|c|c|c|c|c|c|c|c|c|c|c|}
\hline
    %\multicolumn{4}{|c|}{Models fused} & \multicolumn{4}{c|}{Evidential fusion}\\
   %\hline
   %\multicolumn{14}{c}{}\\
   %\hline
   Intensity & Spherical & Cartesian & Cartesian (w/o TNet) &Precision & Recall & F1-score & IoU\\
   \hline
   \checkmark&\checkmark&&&0.8986&0.8588&0.8782&0.7829\\%&0.9661&0.6943&0.8079&0.6777\\
   \hline
    &\checkmark&\checkmark&&0.8932&\textbf{0.8991}&0.8962&0.8119\\%&0.9532&0.8163&0.8795 &0.7849\\
   \hline
    &\checkmark&&\checkmark&0.9114&0.8740&0.8923&0.8055\\%&0.9626&0.7791&0.8612&0.7562\\
   \hline
    \checkmark&&\checkmark&&0.8869&0.8830&0.8849&0.7936\\%&0.9534&0.7208&0.8209&0.6963\\
   \hline
   \checkmark&&&\checkmark&0.8920&0.8501&0.8705&0.7707\\%&0.9568&0.6882&0.8006&0.6675\\
   \hline
   \checkmark&\checkmark&&\checkmark&\textbf{0.9178}&0.8694&0.8929&0.8066\\%&\textbf{0.9816}&0.6374&0.7729&0.6298\\
   \hline
   \checkmark&\checkmark&\checkmark&&0.9098&0.8904&\textbf{0.9000}&\textbf{0.8182}\\%&0.9809&0.6621&0.7906&0.6537\\
    \hline
\end{tabular}}
\caption{Evaluation of the results for several fusion schemes among RoadSeg networks}
\label{fusiontestsetres}
\end{table*}
\paragraph{}
%We can observe that, in general, Bayesian fusion leads to extremely high Precesion scores, but low Recall scores. This means that when all the fused networks are agreeing on a point, they most of the time are right. However, the conflict among the fused networks is not properly handled by a naïve Bayesian fusion, which leads to low Recall scores, and thus many false negatives. On the other hand, evidential f
Fusion leads to significantly better Precision and F1-scores and IoU scores, with Recall scores that in par with the best ones that are obtained from single networks. Fusion then prevents false-positives from happening, without significantly increasing the false-negative rate. Especially, the fusion of RoadSeg-Intensity, RoadSeg-Spherical and RoadSeg-Cartesian leads to the best results. The obtained road detection is satisfactory, as the F1-score is equal to 0.9, and the IoU is higher than 0.8. This is significantly better than the performances of each individual network, which seems to confirm that, originally, RoadSeg does not have enough parameters to properly detect the road from all the available features. This result moreover is particularly satisfactory because the training and validation labels were obtained automatically. 
\paragraph{}
The use of a TNet improves the results, but that comes at the cost of an increased inference time, as observed in Table~\ref{testsetres}. We however consider that the use of the TNet is relevant, when using evidential fusion. Indeed, the risk of data incest among RoadSeg-Spherical and RoadSeg-Cartesian exists, because the Cartesian coordinates can be obtained from the Spherical ones, and vice-versa. As Dempster's rule of combination is designed to fuse independent mass functions, this risk of data incest should better be limitted when operating in real-life conditions. We however point out that the fact that,  when being trained, all the networks were initialized randomly, and trained on random batches. Although these elements seem to be enough to ensure a certain level of independance~\cite{lakshminarayanan2017simple}, we preferred to enforce further this independance, by fusing networks with different inputs. Moreover, as the TNet transforms the input to the fire modules by an affine transformation, this dependence between RoadSeg-Cartesian and RoadSeg-Spherical is less likely to happen. To support this claim, we observe that just fusing RoadSeg-Spherical and RoadSeg-Cartesian leads to the second best results in terms of F1-score and IoU, while the fusion of RoadSeg-Spherical and RoadSeg-Cartesian without any TNet is significantly less performant. We also show, in Figure~\ref{tneteffect}, a LIDAR scan belonging to the test set, and the pointcloud obtained after transformation by the TNet used in the evidential fusion. The transformation predicted by the TNet is normally applied to a point-cloud that was normalized by batch-normalization, but for the sake of clarity, we apply it on the original point-cloud. Indeed, the normalized point-cloud is, by definition, more compact, and harder to visualize. We can see that, in this case, the TNet mainly applies rotations around the x and z axes. The global alignment of the scan is thus modified. The initial Batch-Normalization layer of RoadSeg-Spherical could also be seen as a global affine transformation applied to the scan. Yet, it is extremely unlikely that both a RoadSeg-Cartesian using a TNet and a RoadSeg-Spherical actually apply the same affine transformation to their input. The TNet can then be considered as a way to enforce independence among RoadSeg-Cartesian and RoadSeg-Spherical. 
\subsection{Comparison of the evidential mass functions obtained from the fused RoadSeg networks}
\paragraph{}
As a reminder, RoadSeg networks are trained as SqueezeSegV2 was originally on the KITTI object dataset, and thus use a weight decay of 0.0001, which was also needed by the results displayed in Equation~\ref{wdres}. We propose to compare the different evidential mass functions that can be obtained from the fusion of RoadSeg-Cartesian, RoadSeg-Spherical and RoadSeg-Intensity, as this is our best performing model, so as to verify their behavior depending on how the $\alpha$ vector is obtained. Especially, we focus on the points of the test set that are misclassified by this fusion of networks. A desirable behavior would be to have uncertain mass values generated for those falsely classified points. We chose to rely on the decomposable entropy for the evidential theory, defined in~\cite{jirouvsek2018decomposable}, to quantify the uncertainty level of the mass functions obtained from the fused RoadSeg networks. This entropy is similar to Shannon's entropy, especially in the sense that an uncertain evidential mass function will lead to a high entropy. In our case, this entropy measure $H$ on a mass function $m$ is computed as:
\raggedbottom
\begin{equation}
\begin{split}
H(m) = &-(m(\{R\})+m(\{R,\neg R\}))log_2(m(\{R\})\\
&+m(\{R,\neg R\})) \\
&-(m(\{\neg R\})+m(\{R,\neg R\}))log_2(m(\{\neg R\})\\
&+m(\{R,\neg R\})) \\
&+ m(\{R,\neg R\})log_2(m(\{R,\neg R\}))
\end{split}
\end{equation}
\raggedbottom
\begin{table*}
\centering
\resizebox{\textwidth}{!}{%
\begin{tabular}{|c|c|c|c|c|c|}
\hline
\multicolumn{5}{|c|}{Set(s) used for post-processing} & \cellcolor{black}\\
\hline
None & train & validation & additional\textunderscore train & additional\textunderscore test&Mean Entropy on misclassified points\\
\hline
\checkmark&&&&&0.3537\\
\hline
&\checkmark&&&&0.3663\\
\hline
&\checkmark&\checkmark&&&0.3664\\
\hline
&&&\checkmark&&0.3676\\
\hline
&&&&\checkmark&0.3668\\
\hline
&\checkmark&\checkmark&\checkmark&&0.3669\\
\hline
&\checkmark&\checkmark&&\checkmark&0.3664\\
\hline
&\checkmark&\checkmark&\checkmark&\checkmark&0.3668\\
\hline
\end{tabular}}
\caption{Comparison of the mean evidential entropies that were generated, for misclassified points, by RoadSeg-Intensity, RoadSeg-Spherical and RoadSeg-Cartesian}
\label{evientro}
\end{table*}
\paragraph{}
We report in Table~\ref{evientro} the mean entropy on the misclassified points, from evidential mass functions obtained by solving the minimization problem described in Equation~\ref{min} on several sets. We compare those values with their equivalent obtained without post-processing of the weights. The considered sets for the post processing were the training set, the training and validation sets, a collection of 2221 unlabelled random LIDAR scans that were acquired at the same locations as the the training and validation sets, and a collection of 695 unlabelled scans acquired in Guyancourt alongside the test set. The latest sets correspond to a tenth of the whole squences that were recorded to make the training, validation and test set. To ensure variety among the scans, the difference between the timestamps of those unlabelled scans is at least of one second. Those sets are denoted respectively as \textit{additional\textunderscore train} and \textit{additional\textunderscore test}.
\raggedbottom
\begin{figure*}[h!]
\centering
\begin{subfigure}{.5\textwidth}
  \centering
  \includegraphics[width=0.8\linewidth]{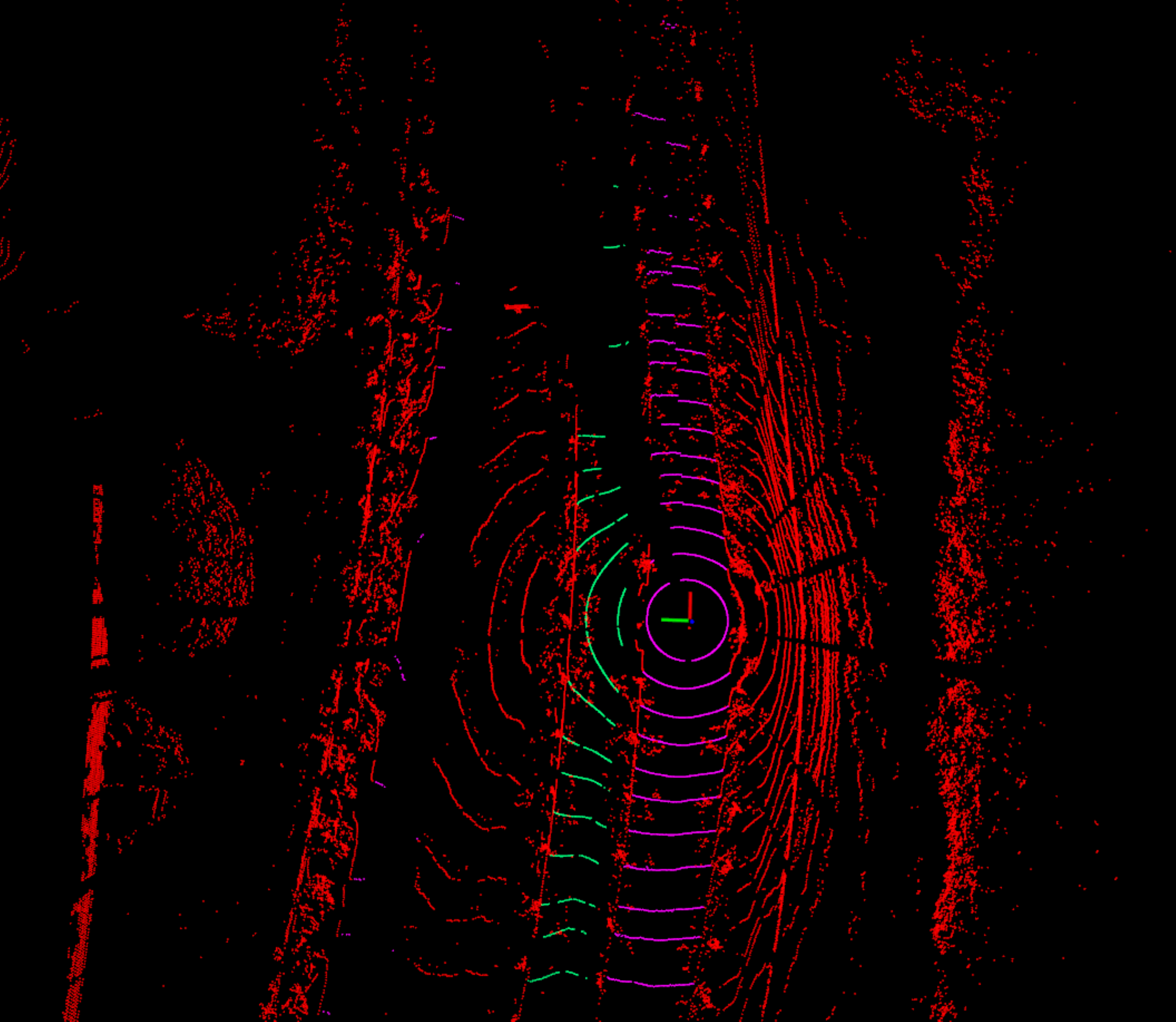}
  \caption{Labels}
  %\label{fig:sub1}
\end{subfigure}%
\begin{subfigure}{.5\textwidth}
  \centering
  \includegraphics[width=0.8\linewidth]{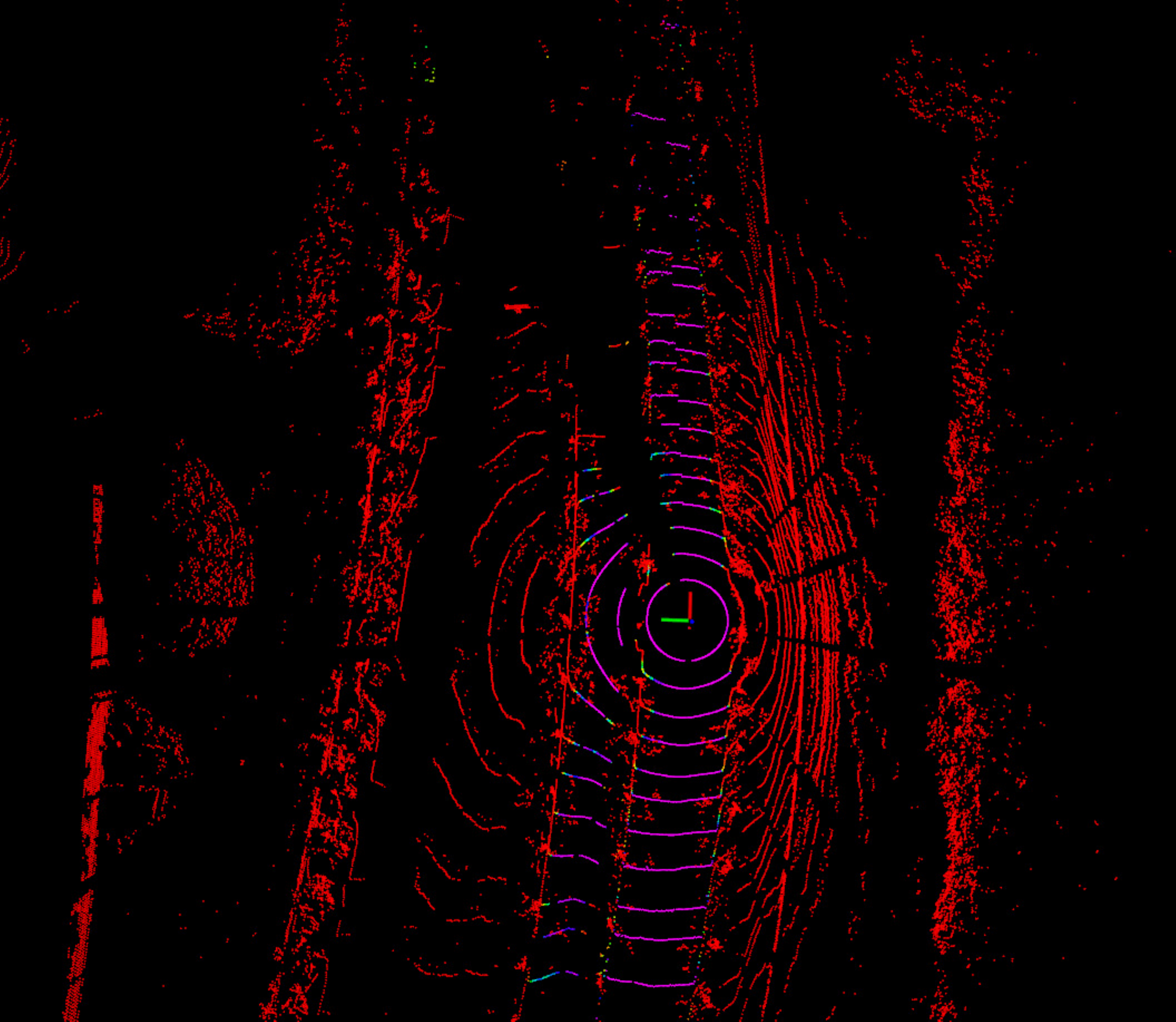}
  \caption{Predicted road probabilities}
  %\label{fig:sub2}
\end{subfigure}
\caption{Scan For which the fused networks achieve the best F1-score. On the left side, purple points were manually labelled as road; red ones as obstacles; green ones as do not care, as they correspond to a reserved bus lane. On the right, the purpler a point is, the higher the probability of being a road point is high, according to the fused networks.}
\label{bestf1}
\end{figure*}
\paragraph{}
The lowest mean entropy on the misclassified points corresponds to the vanilla results, when no post-processing is used on the weights of the networks. This is thus the most over-confident case. However, all the values of mean entropy are extremely close among each other. Interestingly the maximum entropy is achieved when only \textit{additional\textunderscore} train is used for post-processing. The use of data similar to the test set does not seem particularily useful and, counterintuitively, the use of bigger sets did not necessarily lead to more cautious evidential mass functions. As a conclusion, evidential mass functions can be generated only using the training and validation sets, and potentially additonal similar and unlabelled data. The use of Instance Normalization and weight decay is confirmed as a way to obtain near-optimal evidential mass functions during the training, as the differences in terms of entropy are barely visible. It could even be considered that no post-processing should be used, since the dataset on which it should be computed is not clear, but this should be confirmed by further experimental and theoretical analyses.
\subsection{Examples of results}
We now present some segmentation results from the test set, that were generated from the fusion of RoadSeg-Cartesian, RoadSeg-Spherical and RoadSeg-Intensity.
\subsubsection{Best F1 score}
We first show the scan for which the fused networks reach the best results, in terms of per-scan F1-score. On this situation, the fused networks achieved an F1-score of 0.9569, and an IoU of 0.9173. As displayed in Figure~\ref{bestf1}, this is a very simple situation with a straight road, and no other vehicles, although a small part of the scan is missing. The fact that the system can handle such simple situations is reassuring. We also point out the fact that the reserved bus lane, labelled in green, was fully classified as road. Following the policy we previously exposed, those points corresponding to the bus lane were not considered in the evaluation.
\begin{figure*}[h!]
\centering
\begin{subfigure}{.5\textwidth}
  \centering
  \includegraphics[width=0.8\linewidth]{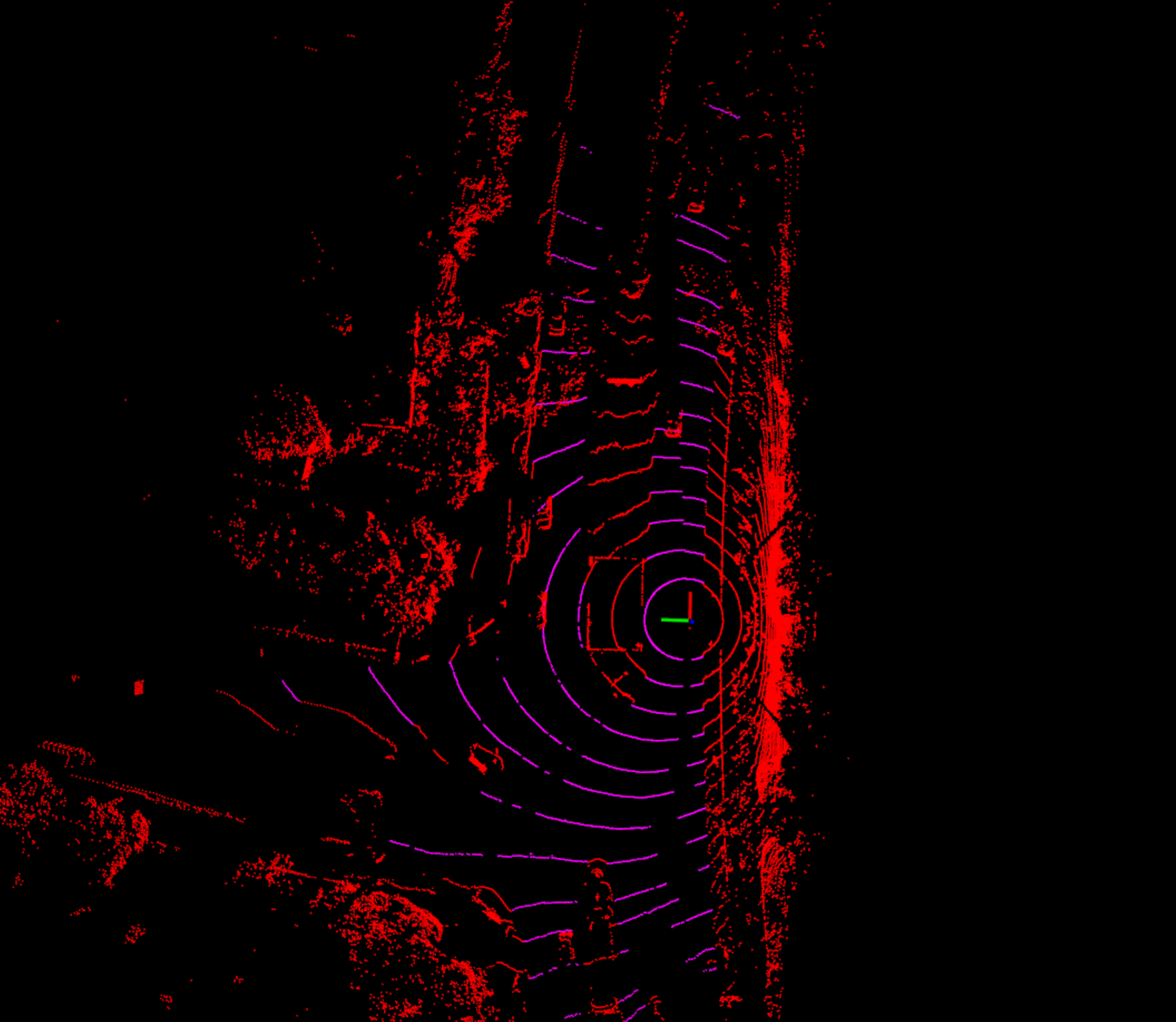}
  \caption{Labels}
  %\label{fig:sub1}
\end{subfigure}%
\begin{subfigure}{.5\textwidth}
  \centering
  \includegraphics[width=0.8\linewidth]{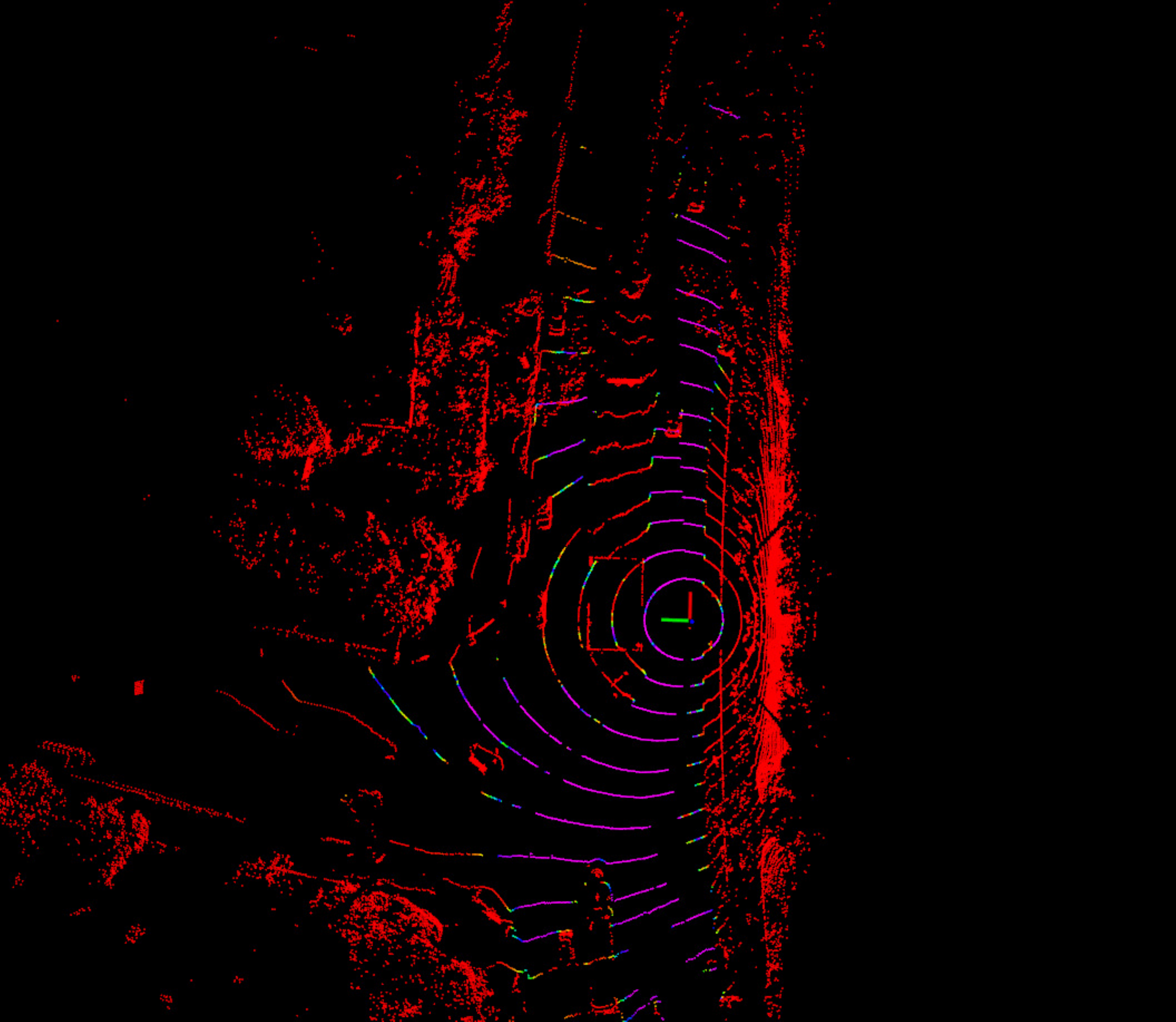}
  \caption{Predicted road probabilities}
  %\label{fig:sub2}
\end{subfigure}
\caption{Scan for which the fused networks achieve the worst F1-score.}
\label{worstF1}
\end{figure*}
\subsubsection{Worst F1 score}
We show, in Figure~\ref{worstF1}, the scan for which the fused networks achieve the worst result in terms of F1-score and IoU. The F1-score for this scan is equal to 0.8971, and the IoU is equal to 0.8133. The errors are mainly localized on the left side, probably because the central median partially occluded this area. The results on the ego-lane are still satisfactory.
\subsubsection{Crossroad}
We present a result at a crossroad. Most of the actual road surface is classified as road. However, remote, narrow roads on top-right and bottom-left are undetected. This is because those roads are hard to distinguish when projected into the range images processed by the RoadSeg networks, due to their narrowness. The network achieves an F1-score of 0.9271 and an IoU of 0.8641 on this scan.
\begin{figure*}
\centering
\begin{subfigure}{.5\textwidth}
  \centering
  \includegraphics[width=0.8\linewidth]{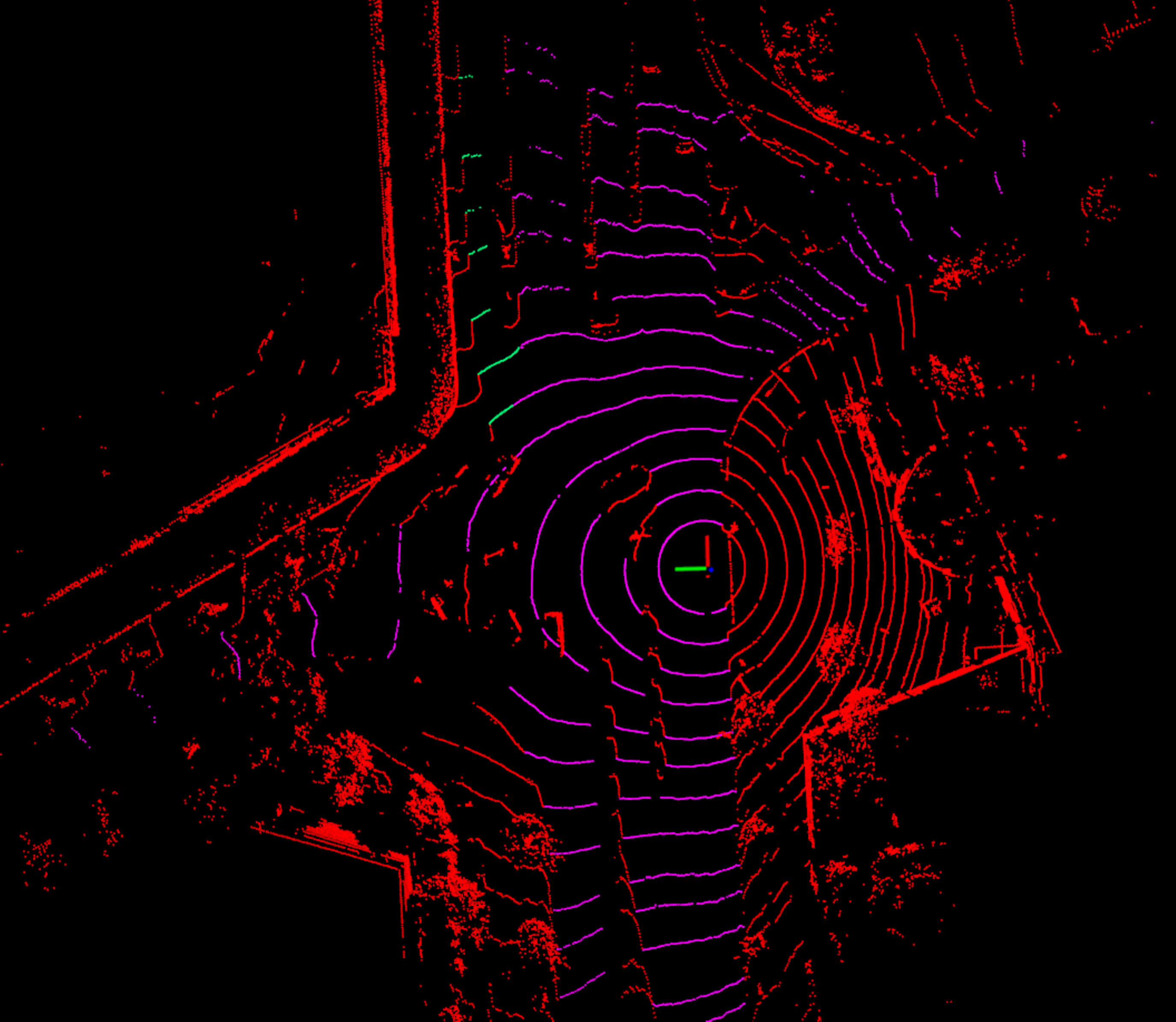}
  \caption{Labels}
  %\label{fig:sub1}
\end{subfigure}%
\begin{subfigure}{.5\textwidth}
  \centering
  \includegraphics[width=0.8\linewidth]{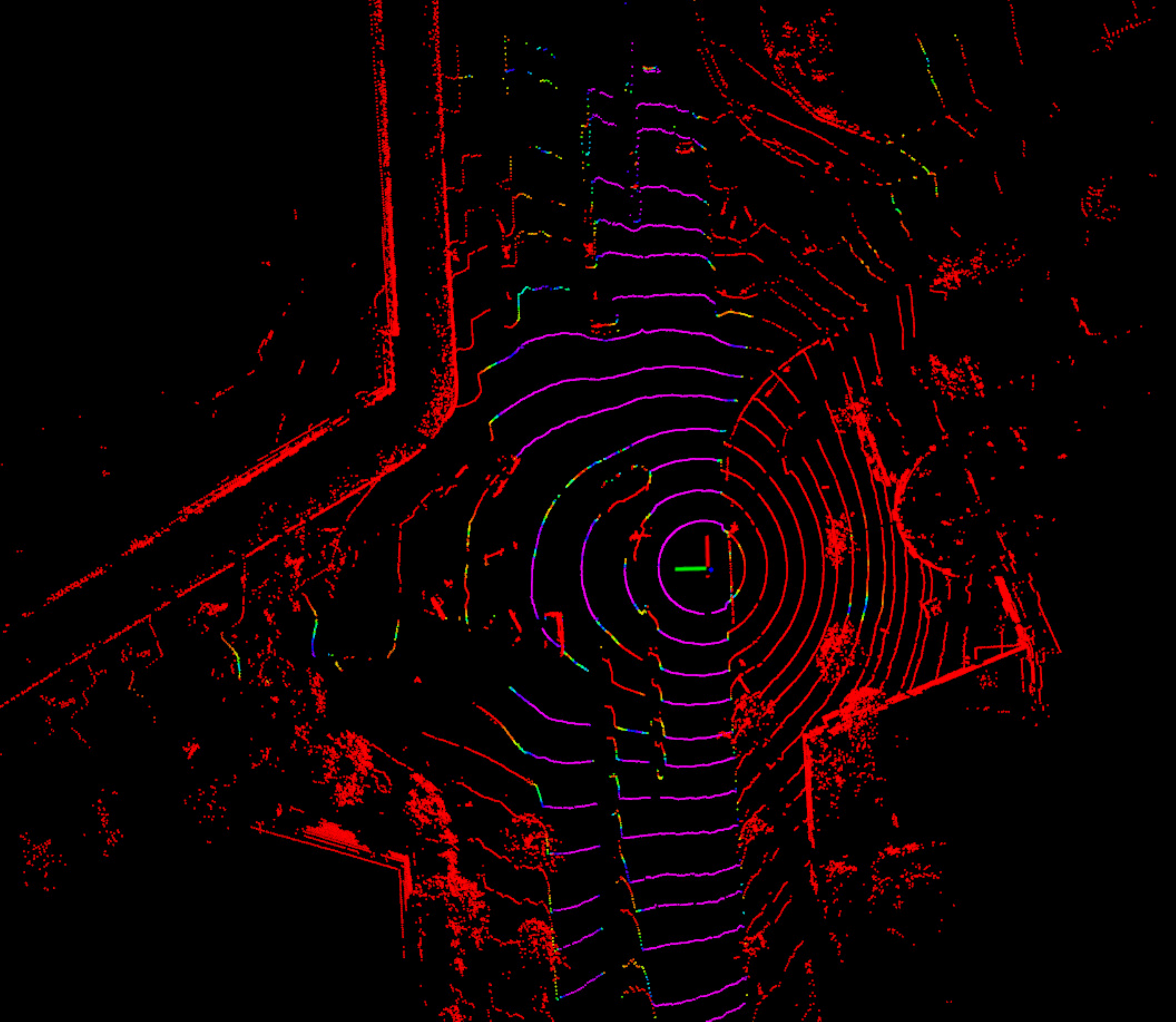}
  \caption{Predicted road probabilities}
  %\label{fig:sub2}
\end{subfigure}
\caption{Example at a crossroad}
\end{figure*}
\begin{figure*}
\centering
\begin{subfigure}{.5\textwidth}
  \centering
  \includegraphics[width=0.8\linewidth]{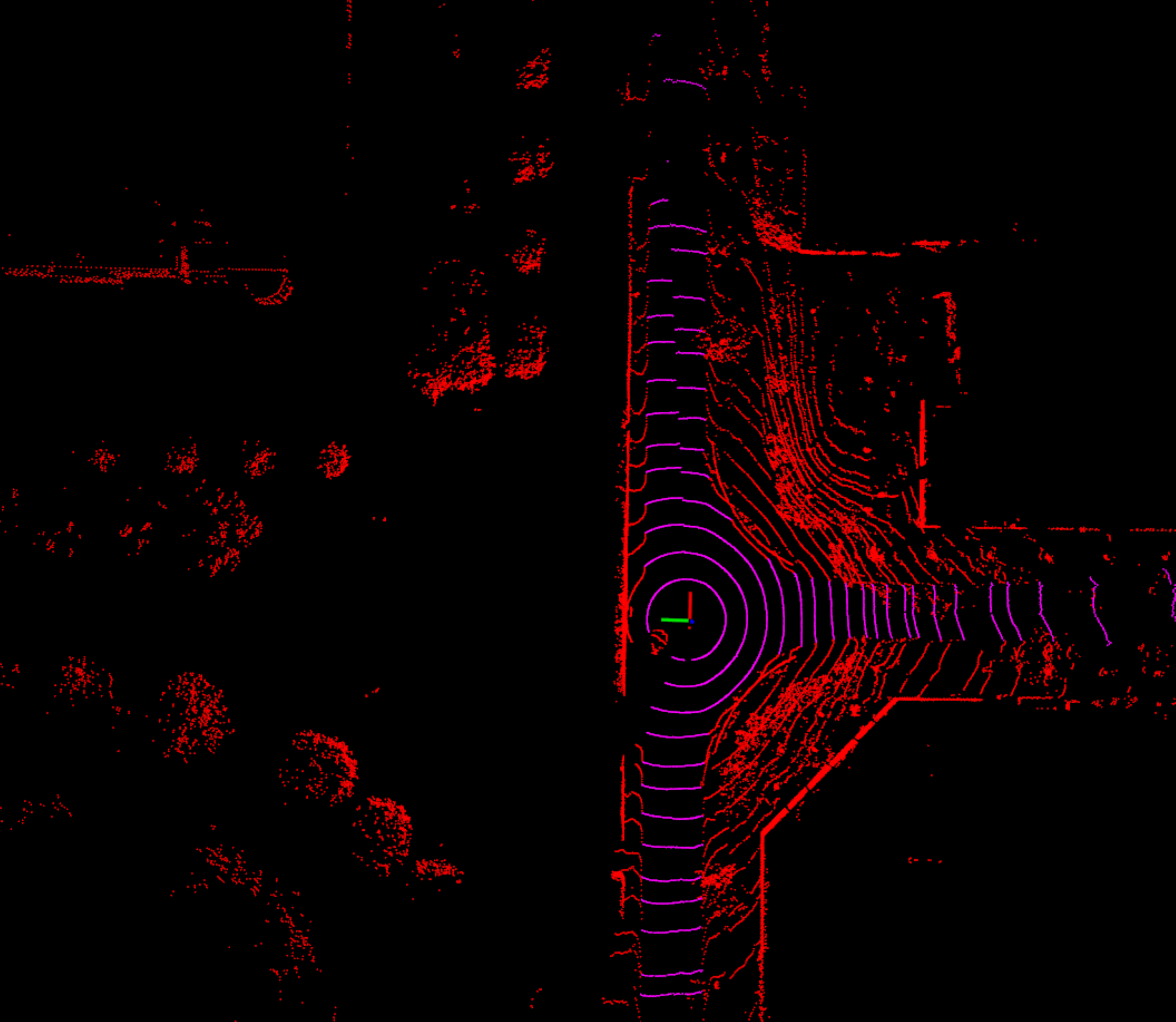}
  \caption{Labels}
  %\label{fig:sub1}
\end{subfigure}%
\begin{subfigure}{.5\textwidth}
  \centering
  \includegraphics[width=0.8\linewidth]{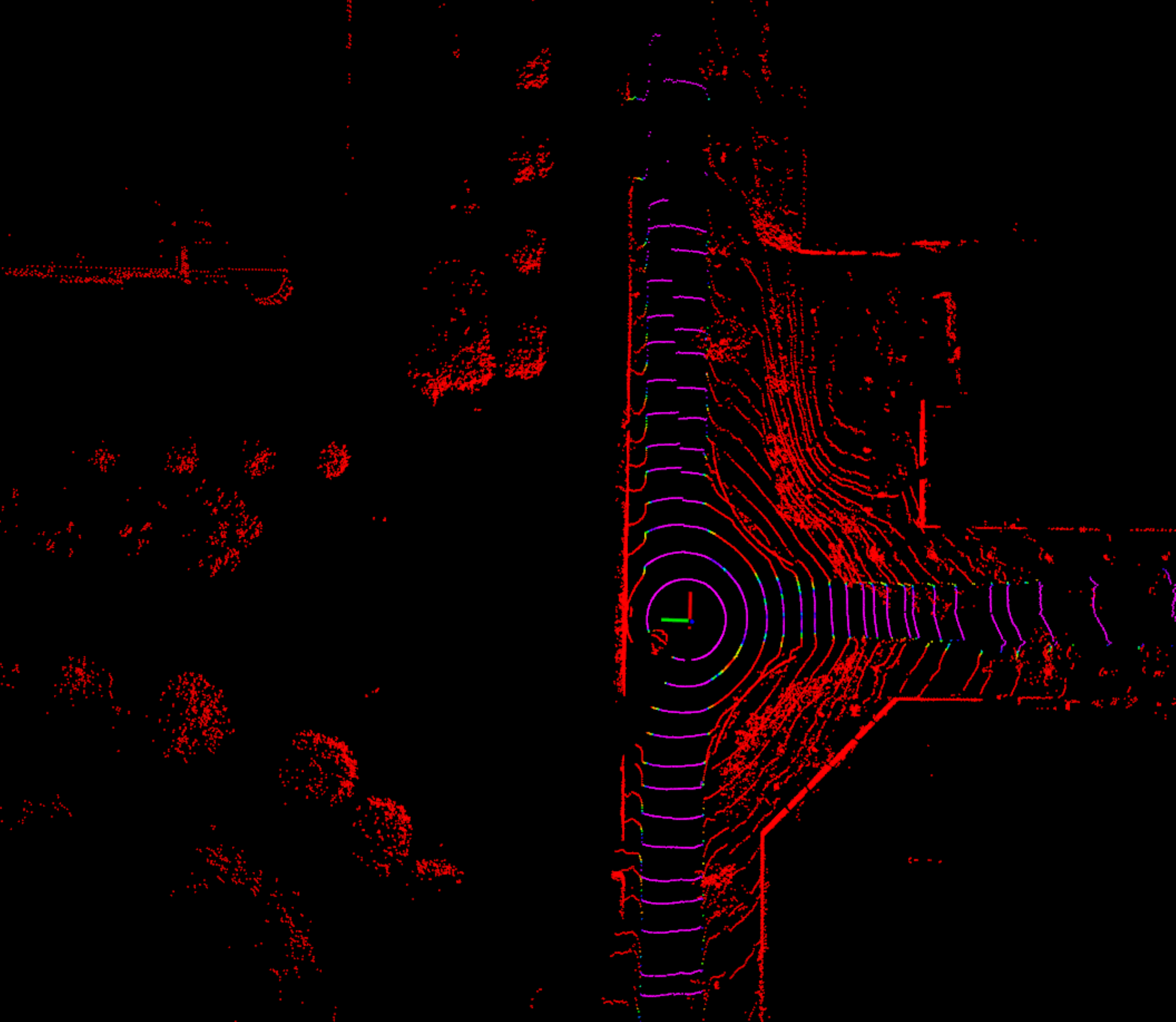}
  \caption{Predicted road probabilities}
  %\label{fig:sub2}
\end{subfigure}
\caption{Exemple at a junction}
\end{figure*}
\subsubsection{Junction}
To counterbalance the results on the previous use case, we present results obtained at a junction. Again most of the road surface is properly detected. However, the networks consider that the entrance to the road on the right is narrower than what it actually is. This is because of our label generation procedure, which relied on a map. Indeed, many of the junctions in those maps were mapped similarly, which under-estimated entrance width that were considered to be equal to the roads' length. The fused networks still achieve an F1-score of 0.9323 and IoU of 0.8732 on this scan.
\subsubsection{Roundabout}
We conclude with a roundabout. The results are very satisfying, as most of the actual road surface is properly detected. The remote vehicle is also properly considered as an obstacle. The central median in front of the vehicle if however partially considered as belonging to the road. The fused networks achieve an F1-score of 0.9010 and an IoU of 0.8199 on this scan. 
\paragraph{}
\begin{figure*}
\centering
\begin{subfigure}{.5\textwidth}
  \centering
  \includegraphics[width=0.8\linewidth]{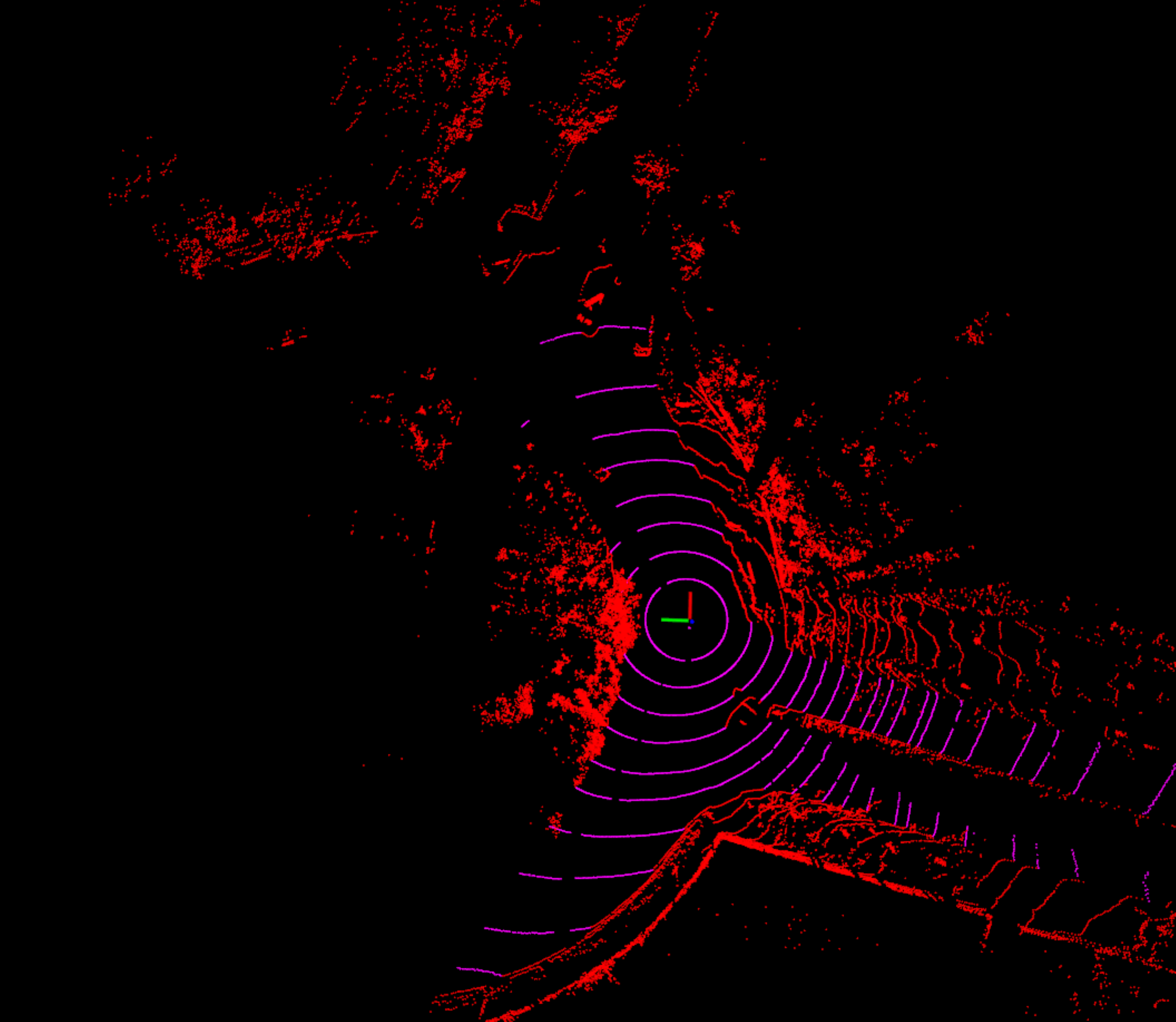}
  \caption{Labels}
  %\label{fig:sub1}
\end{subfigure}%
\begin{subfigure}{.5\textwidth}
  \centering
  \includegraphics[width=0.8\linewidth]{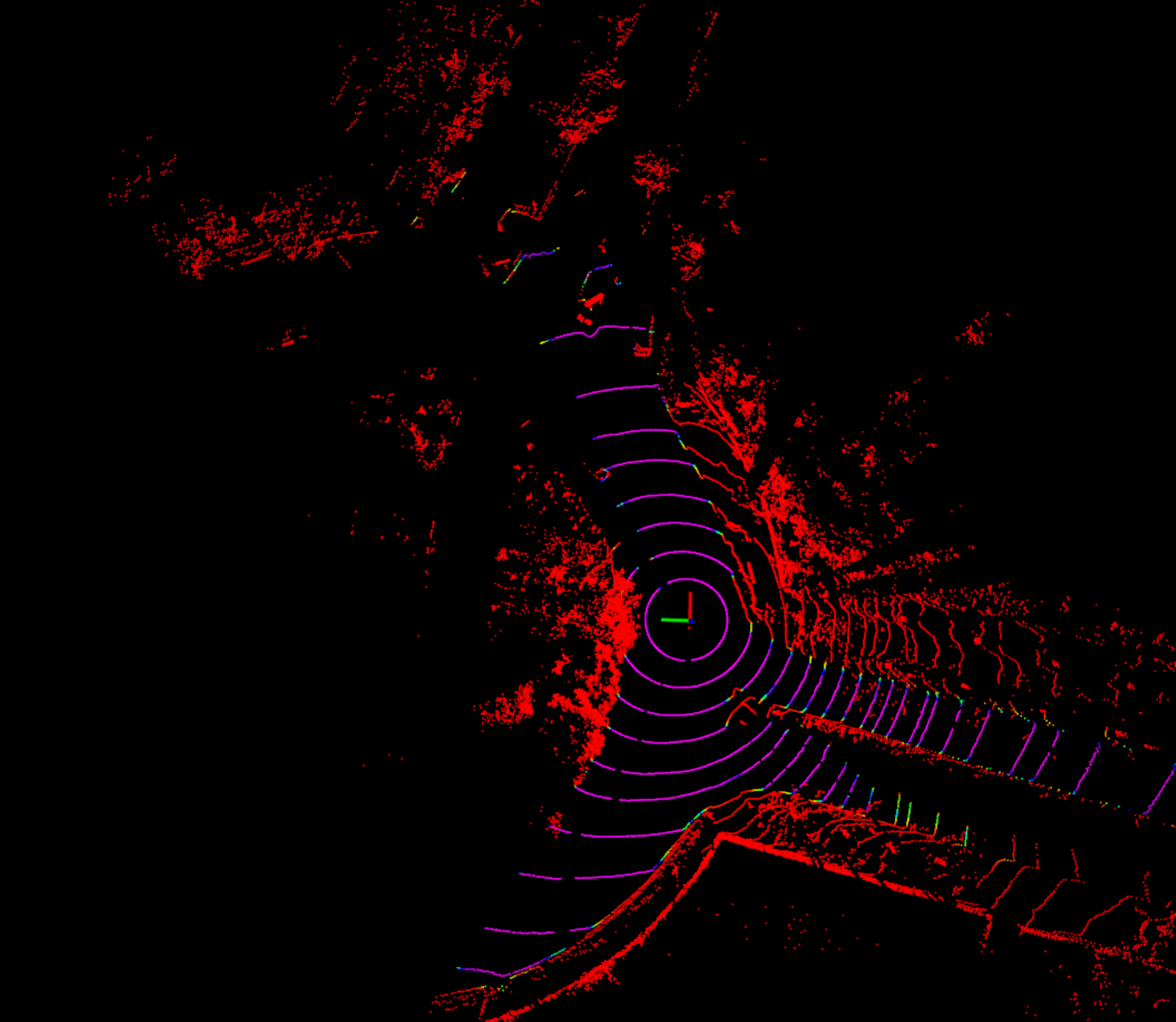}
  \caption{Predicted road probabilities}
  %\label{fig:sub2}
\end{subfigure}
\caption{Example at a roundabout}
\end{figure*}
The fusion of RoadSeg-Cartesian, RoadSeg-Spherical and RoadSeg-Intensity leads to very promising results, especially in straight roads and roundabouts. Junctions and crossroads can also be processed with a certain efficiency, although the network is limited by what was present in the training set. Indeed, very few crossroads were automatically labelled, and the approximative representation of the junctions in the maps that we used influence the final results of the network. Hopefully, the fact that these results have been achieved with a relatively small training set, that was automatically labelled, indicates that there is probably room for improvement. Additional and finer training data would certainly correct the behavior of the fused networks.
\section{Evidential road mapping and road object detection from the fused RoadSeg networks}%
\begin{figure*}
\centering
\includegraphics[width=0.3\textwidth, angle=90]{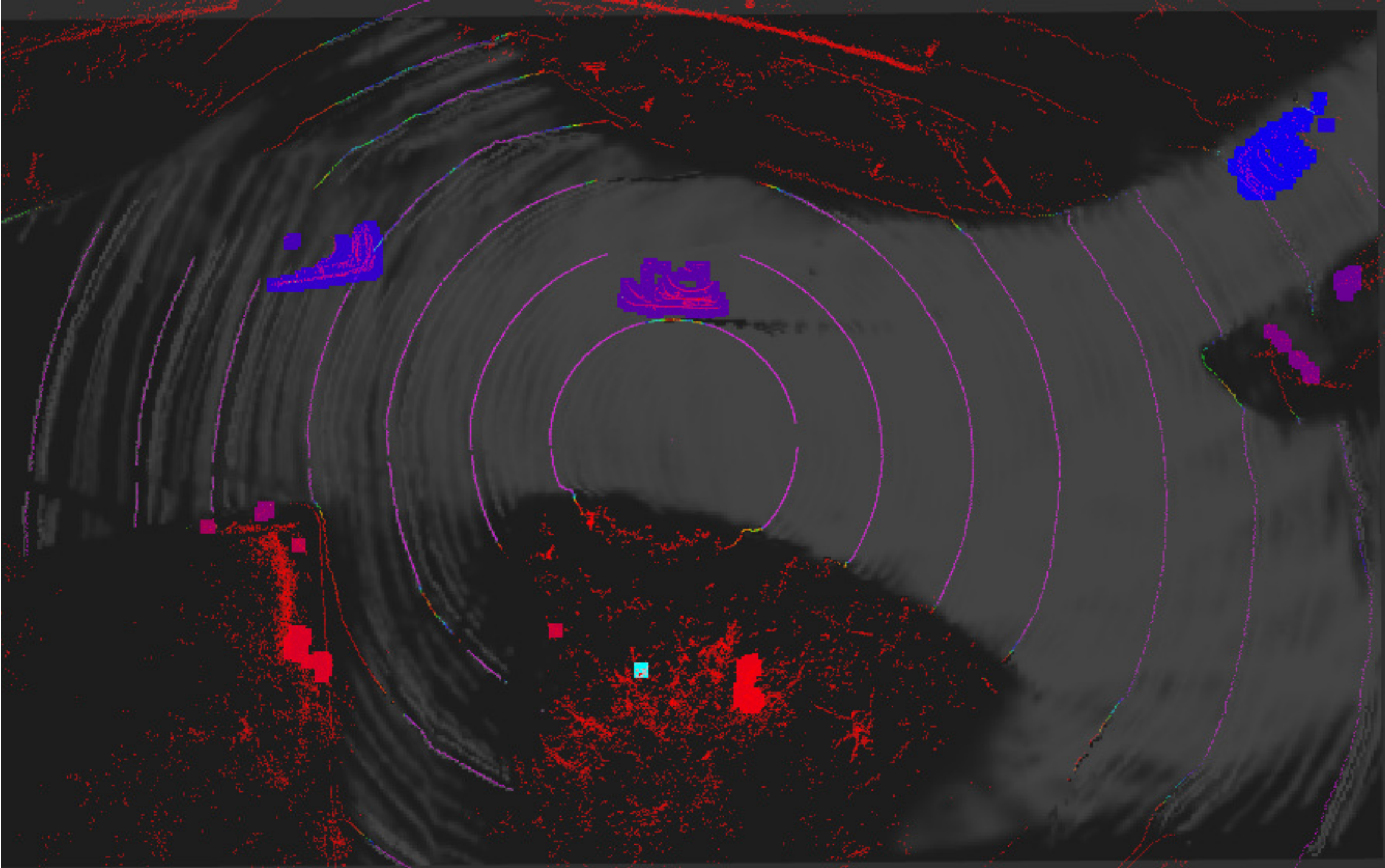}
\caption{Results obtained from the road mapping and object detection pipeline. The LIDAR scan classified by the three RoadSeg network is visible. Below, a greyscale RoadGrid represents the belief for $m(\{R\})$ in each cell. The clustered objects (mainly vehicles on the road) are colored according to their cluster id.}
\label{outex}
\end{figure*}

\label{gridmapalgo} 
\begin{figure*}[h!]
\centering
\includegraphics[width=0.75\textwidth]{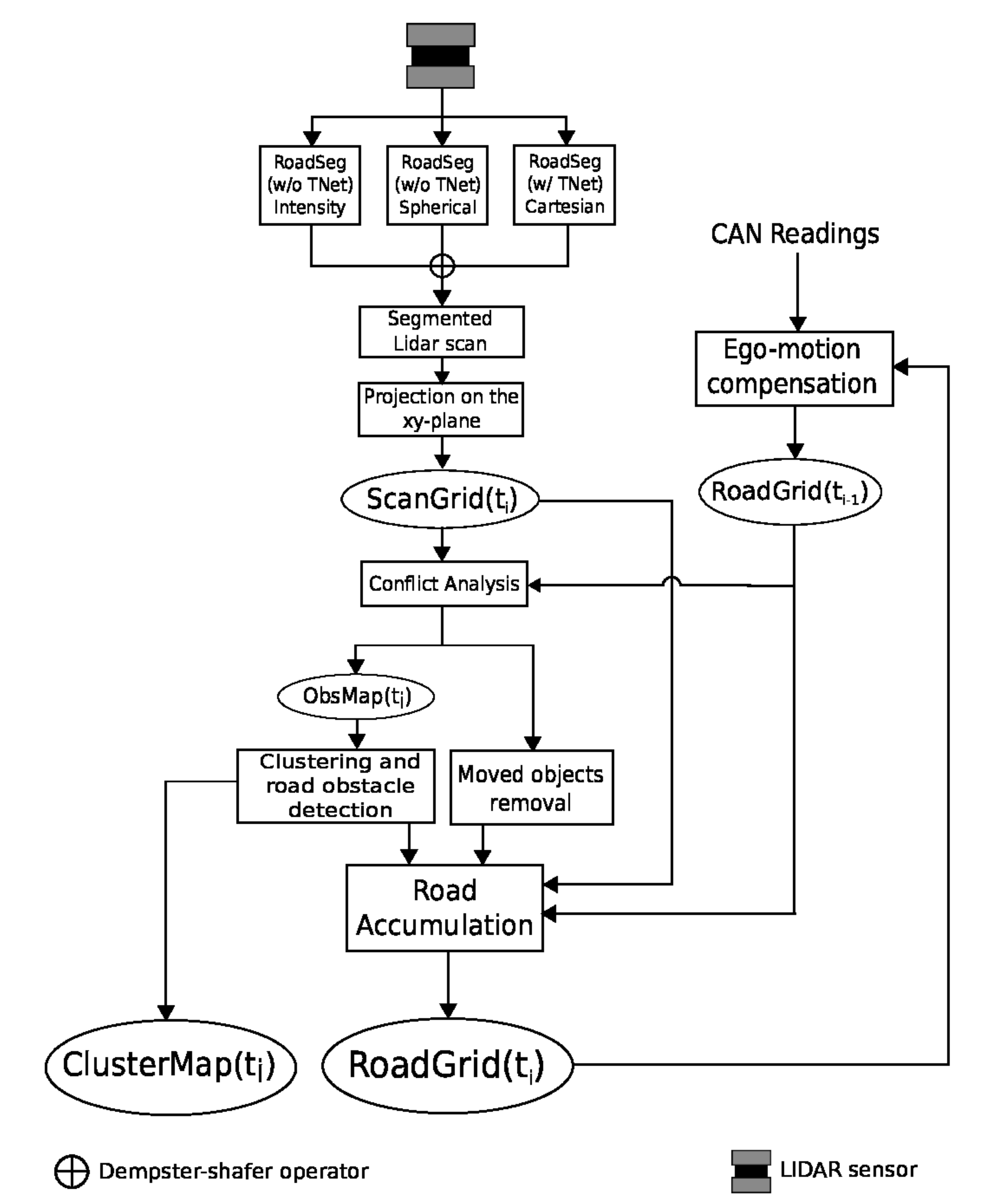}
\caption{General evidential road mapping and road object detection algorithm}
\label{mapobsalgo}
\end{figure*}
From the satisfactory results obtained by fusing several RoadSeg networks, and evidential mass functions that can be generated from their outputs, an usable representation of the road for a navigation algorithm can be created, by fusing consecutive road detection results. We chose to rely on an evidential grid mapping framework. A naive approach would be to extend the evidential grid mapping algorithm depicted in~\cite{yu2014evidential}, by replacing the original geometrical model by the evidential mass functions generated from RoadSeg. Such an approach however cannot handle moving objects properly, as it only relies on the fusion of consecutive observations. Inspired by the work in~\cite{moras2011moving}, in which it is observed that the conflict induced by the fusion of evidential grids can correspond to moving objects, we propose an evidential road mapping algorithm, to generate both a grid depicting the actual road surface, and a list of moving road obstacles. We consider that the area below a moving road obstacle should be considered as road, and that the road grid should only depict the reality of the road, independently of the presence of obstacles. Figure~\ref{mapobsalgo} depicts the whole algorithm, and Figure~\ref{outex} presents a possible output of the system. In the next sections, we present its different steps in details.%cite ppniv here
\subsection{Projection of the segmentation on the xy-plane}
\paragraph{}
As an evidential fusion of RoadSeg-Intensity, RoadSeg-Spherical and RoadSeg-Cartesian leads to the best classification performances, the algorithm processes the segmentation results obtained after evidential fusion of the three networks. The generation of evidential mass functions can be performed either from the original weights, that are directly obtained after the training of the networks, or from post-processed weights. We arbitrarily chose to make the road grid correspond to the xy-plane, in the reference coordinate system used by the LIDAR. This plane is split into equally sized grid cells, which cover a pre-defined area around the sensor. The state of each cell of index $i$ can be represented by three evidential mass values $m_i(\{R\})$ (road), $m_i(\{\neg R\})$ (not road) and $m_i(\{R,\neg R\})$ (unknown). Similarly to what is done at the LIDAR point level, those evidential mass values respectively quantify the evidence towards the fact that the $i^{th}$ cell belongs to the road, does not belong to the road, or is in an unknown state. A straightforward way to compute $m_i(\{R\}), m_i(\{\neg R\})$ and $m_i(\{R,\neg R\})$ is to project, into the xy-plane, all the LIDAR points, and the evidential mass values that are obtained after the fusion of the RoadSeg networks. As previously stated, the RoadSeg networks process scans in which the motion of the vehicle was not compensated. As such, this motion has to be compensated, in the segmented point-cloud that we obtain from the RoadSeg networks, before projecting the points into the grid map. We thus reuse the same approach and assumptions as for the soft-labelling procedure, which we described in Section~\ref{subsec:softlabelprocedure}, except that we use an identity Rotation/Translation matrix, because the grid map and the LIDAR sensor share the same coordinate system. Then, $m_i(\{R\}), m_i(\{\neg R\})$ and $m_i(\{R,\neg R\})$ can be obtained by fusing the mass values of the points projected into the grid-cell $i$, thanks to Dempster's rule of combination. To reduce the computational complexity of this projection and fusion step, each grid cell is processed in parallel. For the sake of clarity, we drop the cell-index $i$. The number of points projected into each grid-cell is unknown, and varries over time and for each cell of the grid. To solve this issue, we rely on the rewriting of the Dempster-Shafer operator in terms of \textit{commonality functions}~\cite{shafer1976mathematical}.
\paragraph{}
For $\Omega=\{R,\neg R\}$ our binary frame of discernment, a commonality value $Q(A)$ can be computed from a mass function $m$ for each element $A \in 2^{\Omega}$, as follows:
\begin{equation}
Q(A) = \sum_{B\supseteq A}m(B)
\label{m2q}
\end{equation}
The evidential mass function $m$ can be recovered from the commonality values, as follows:
\begin{equation}
m(A) = \sum_{B\supseteq A}(-1)^{|B|-|A|}Q(B)
\label{q2m}
\end{equation}
Commonality functions can be used to fuse $n$ evidential mass functions into a fused mass function $m_{res}$ as follows:
%%%%à enlever ?
\begin{enumerate}
\item Compute $Q_1$,...,$Q_n$ from the $n$ mass functions, using Equation~\ref{m2q}
\item For each $A \in 2^{\Omega}$, $Q_{res}(A) = \exp(\sum_{j=1}^n ln(Q_j(A)))$
\item Compute $m^*_{res}$ from $Q_{res}$, the unnormalized version of $m_{res}$ using Equation~\ref{q2m}
\item Normalize $m_{res}$ as follows: $\forall A \in 2^\Omega \setminus \{\emptyset\}$, $m_{res}(A) = Km^*_{res}(A)$ witk $K = 1/(1-m(\emptyset))$ ; $m(\emptyset) = 0$ 
\end{enumerate}
%à enlever ?
This procedure is equivalent to consecutively applying the Dempster's rule of combination among the $n$ evidential mass functions. However, this formulation enables the projection and fusion operations to be reinterpreted as an operation on a 2D histogram. 
\paragraph{}
The log-commonalities associated to each point can trivially be computed in parallel, after the fusion of the results generated by the three RoadSeg networks. If $n$ corresponds to the number of points that are projected into a grid cell, and whose evidential mass functions have to be fused, $\sum_{j=1}^n ln(Q_j(A))$ can be computed by histogramming the x and y coordinates of each point, and by weighting the samples by the corresponding log-commonalities. The evidential mass values associated to each cell can then be recovered for each cell. We call the resulting evidential grid, which was only generated from a single scan, a \textit{ScanGrid}.

\subsection{Conflict analysis}
\begin{figure*}[h]
\centering
\includegraphics[width=0.7\textwidth]{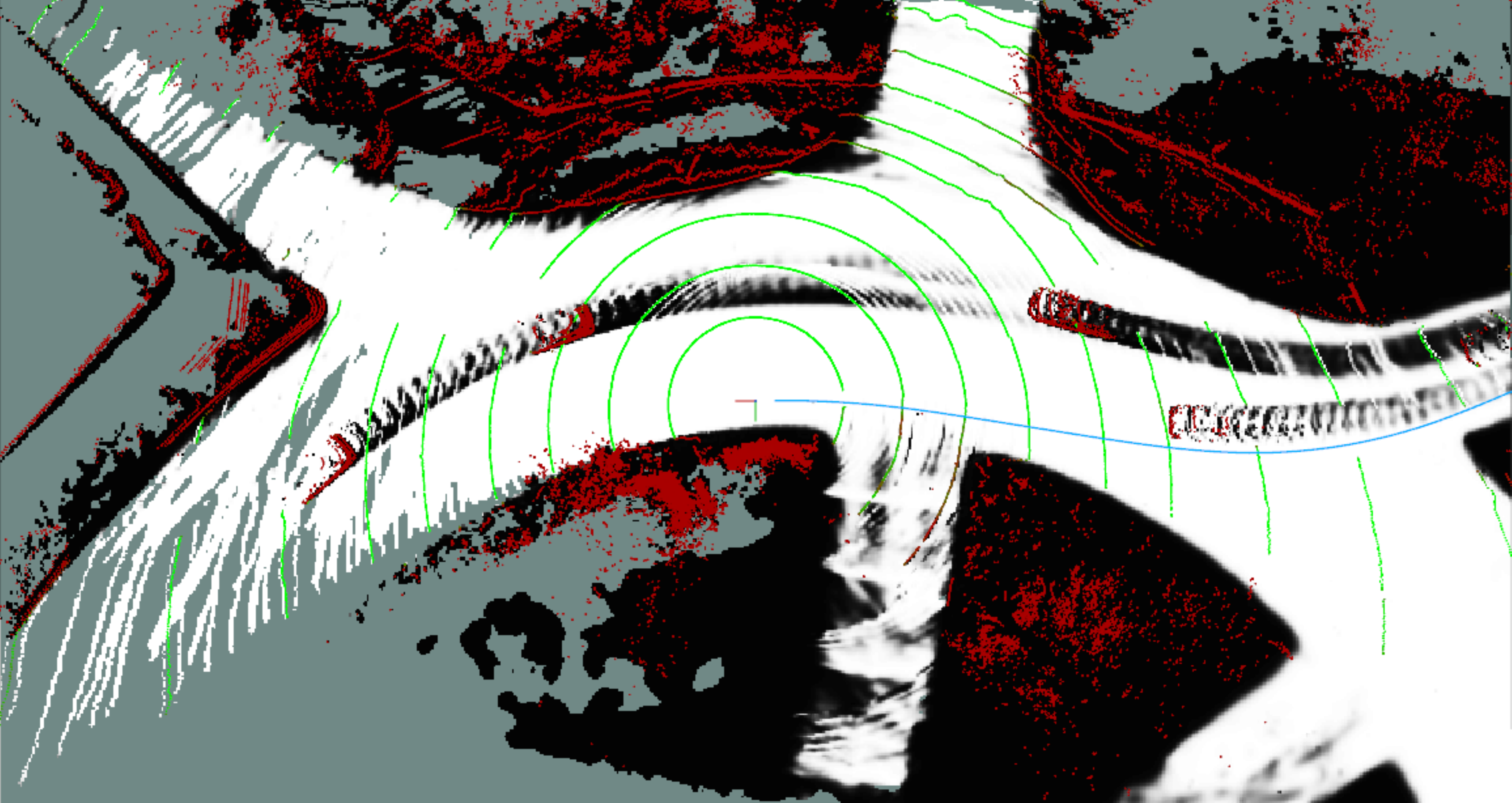}
\caption{Exemple of RoadGrid obtained by accumulating ScanGrids without considering the objects that stand on the road. White cells have an $m(\{R\})$ value higher than 0.5, black cells have  an $m(\{\neg R\})$ value higher than 0.5, grey cells have an $m(\{R,\neg R\})$ value higher than 0.5.}
\label{accuroad}
\end{figure*}

\paragraph{}
In order to generate a dense representation of the road, ScanGrids have to be fused over time. Let a \textit{RoadGrid} be an evidential grid that has been obtained by accumulating several previous ScanGrids. A RoadGrid is supposed to only represent the road surface, without considering objects that might stand on the road. The latest ScanGrid is noted ScanGrid($t_i$), and the latest RoadGrid available is noted RoadGrid($t_{i-1}$). Let $m_{t_i}$ be the evidential mass function that correspond to a given cell of ScanGrid($t_i$), and $m_{t_{i-1}}$ the evidential mass function of the cell of RoadGrid($t_{i-1}$) that is at the same position. A naive way to accumulate ScanGrids would be to use, again, the Dempster-Shafer operator to fuse all the $m_{t_i}$s with the corresponding $m_{t_{i-1}}$.  However, this could lead to a catastrophic accumulation of objects over time, that would affect the estimated road surface, without corresponding to actual objects. This case is depicted in the Figure~\ref{accuroad}. The current LIDAR scan is depicted in green and red. Green points are classified as road points, and red points as obstacles (not road). Under the scan, an evidential grid corresponding to the simple accumulation of ScanGrids is depicted. White cells are classified as road cells ($m(\{R\})>0.5$), black cells as obstacle cells ($m(\{\neg R\})>0.5$), and the grey ones are in an unknown state ($m(\{R,\neg R\})>0.5$). We can observe that the vehicles that are driving on the road create artifacts, and cells that intersect their trajectories are falsely considered as not belonging to the road surface. Consequently, it must be ensured that the objects that stand on the road, and are potentially moving, are not falsely fused with the RoadGrid. We introduce two frame of discernments: $\Omega_{obs}=\{O,\neg O\}$ and $\Omega_{displaced}=\{D,\neg D\}$. The first one models the presence of road obstacles that do not belong to the road surface, as the $O$ proposition. The second one models the fact that a previously present road obstacles is no longer present, as the $D$ proposiion. This case typically corresponds to a vehicle that was static on the road, while the RoadGrid was being generated, and started moving. 
\paragraph{}
$\Omega_{obs}$ can be used to detect which cells of the ScanGrid($t_i$) should not be fused with RoadGrid($t_{i-1}$). The evidential mass functions in this frame of discernment can be computed, for each cell, from the conflict between the $m_{t_{i-1}}$ and $m_{t_{i}}$ mass functions. Indeed, a high value for both $m_{t_{i}}(\{\neg R\})$ and $m_{t_{i-1}}(\{R\})$ can indicate that a moving road obstacle is currently observed in the corresponding ScanGrid($t_i$) cell, and thus should not be fused with RoadGrid($t_{i-1}$). However, it could also indicate that the neural network has trouble detecting a given road, meaning that the corresponding cells should instead be fused. Let $m_{obs}$ be the evidential mass function associated to a cell of ScanGrid($t_i$) under the $\Omega_{obs}$ frame of discernment. We propose to compute $m_{obs}$ as follows:
\raggedbottom
\begin{subequations}
\small
\begin{align}
&m_{obs}(\{O\}) = \alpha(\overline{Z}) m_{t_{i-1}}(\{R\}) m_{t_{i}}(\{\neg R\})\\
&m_{obs}(\{\neg O\}) = 0\\
&m_{obs}(\{O, \neg O\}) = 1 - m_{obs}(\{O\})
\end{align}
\end{subequations}
This formulation supposes that only $m_{t_{i-1}}(\{R\})$ and $m_{t_{i}}(\{\neg R\})$ can indicate the presence of a road obstacle. The $\alpha$ function computes a discounting factor, which depends on the mean z coordinates of the points that have been projected, while creating the ScanGrid($t_i$), into the considered grid cell. This mean elevation is noted $\overline{Z}$.  As the LIDAR used by the ZoeCab systems is put on the roof of the vehicles, $\overline{Z}$ is typically negative when only ground points have been projected into a grid cell. We define $\alpha$ as follows:
\begin{equation}
\alpha(z) = min(exp(\nu(z+\xi)),1)
\end{equation}
This function only generates discounting factors in the ]0,1] range. The $\xi$ parameter indicates the absolute value of the height from which the conflict does not have to be discounted. The $\nu$ factor monitors the growth of $\alpha(z)$.
\paragraph{}
Similarly, we can define $m_{displaced}$ as the evidential mass function associated to a cell of ScanGrid($t_i$) under the $\Omega_{displaced}$ frame of discernment. We propose to compute $m_{displaced}$ as follows:
\begin{subequations}
\small
\begin{align}
&m_{displaced}(\{D\}) = (1-\alpha(\overline{Z})) m_{t_{i}}(\{R\}) m_{t_{i-1}}(\{\neg R\})\\
&m_{displaced}(\{\neg D\}) = 0\\
&m_{displaced}(\{D, \neg D\})= 1 - m_{displaced}(\{D\})
\end{align}
\end{subequations}

\subsection{Moved objects removal}
\paragraph{}

From $m_{displaced}$, grids that should not be considered anymore as occupied in RoadGrid($t_{i-1}$) can easily be detected. The grids in RoadGrid($t_{i-1}$) for which  $m_{displaced}(D)$ is higher than 0.5 are reinitialized to a fully unknown state: $m_{t_{i-1}}(\{R\}) = 0$, $m_{t_{i-1}}(\{\neg R\}) = 0$, $m_{t_{i-1}}(\{R, \neg R\}) = 1$. 

\subsection{Clustering and road object detection}
\paragraph{}
Similarly, the $m_{obs}$ mass can be used to detect grid cells that belong, in a ScanGrid, to a road obstacle, and should not be fused with RoadGrid($t_{i-1}$). A binary ObsMap($t_i$) map is created from ScanGrid($t_i$) and the $m_{obs}$ values. This binary map represents, for each cell, the presence of a road obstacle. A binary cell is set to 1 if the corresponding cell in ScanGrid($t_i$) has an $m_{obs}(O)$ value higher than 0.5. Otherwise, it is set to 0.
\begin{figure*}[h]
\centering
\begin{subfigure}{.45\textwidth}
\centering
  \includegraphics[height=\textwidth, angle=90]{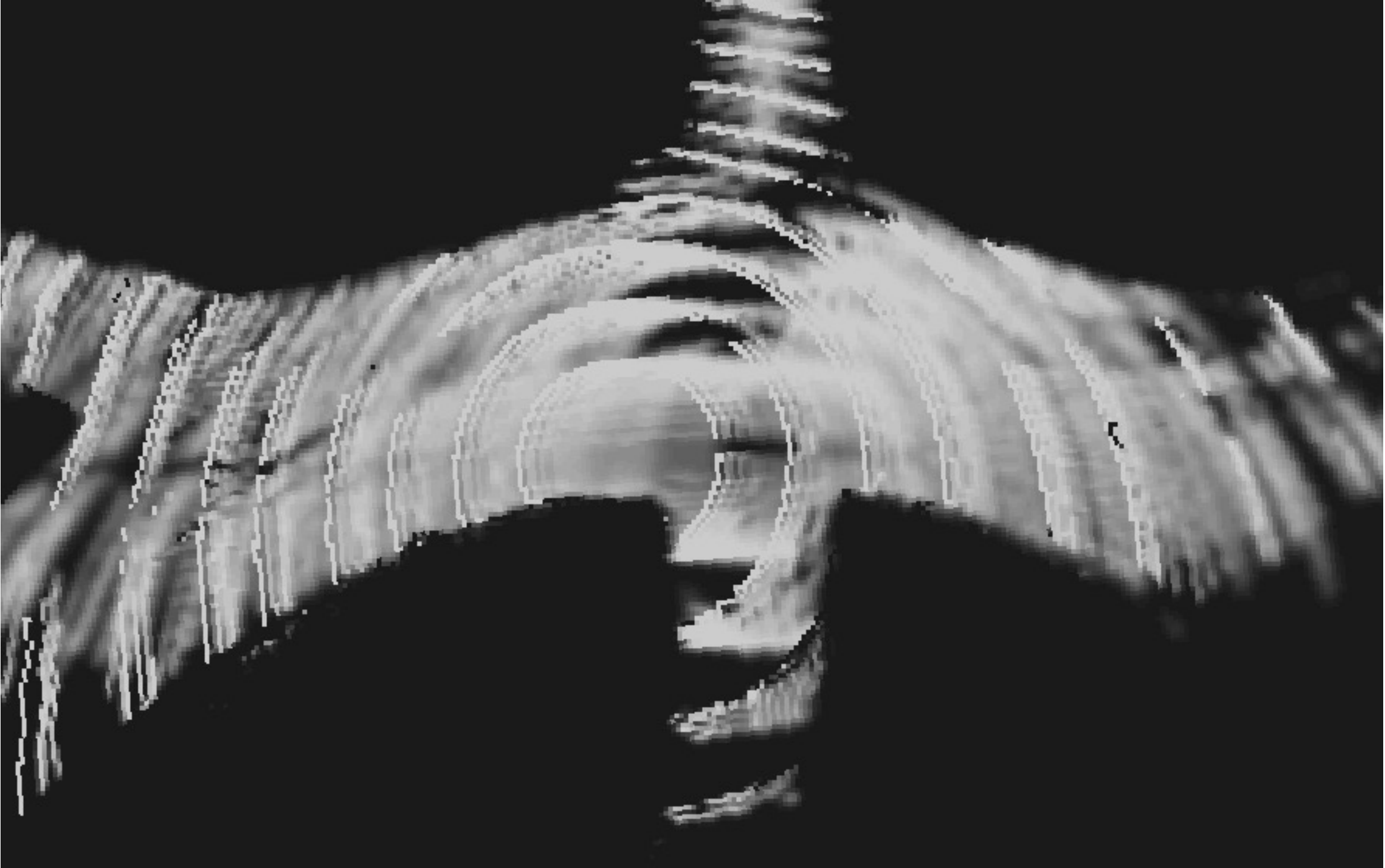}
  \caption{$m(\{R\})$ in RoadGrid($t_{i-1}$)}
  %\label{fig:sub1}
\end{subfigure}
\begin{subfigure}{.45\textwidth}
\centering
  \includegraphics[height=\textwidth, angle=90]{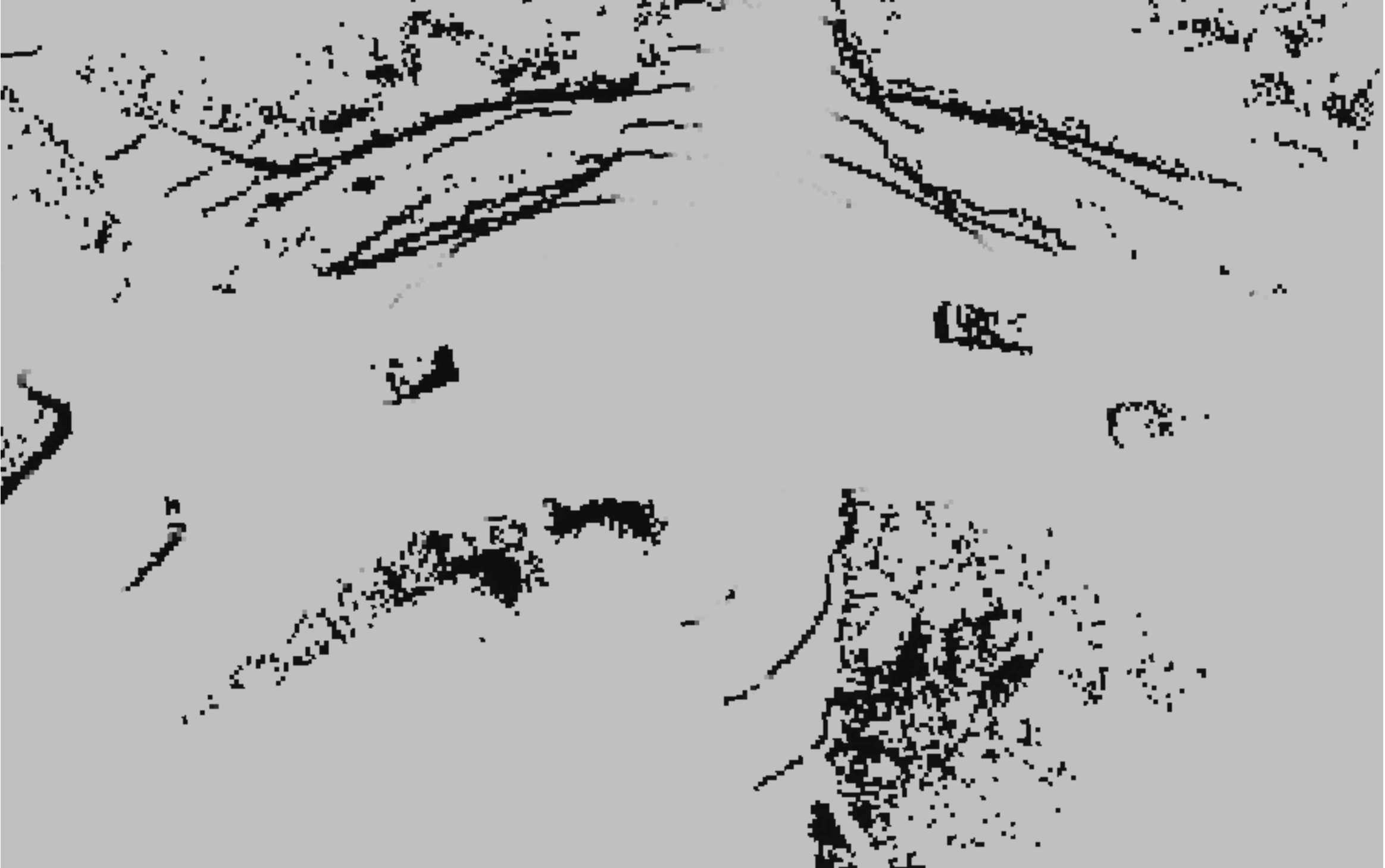}
  \caption{$m(\{\neg R\})$ in ScanGrid($t_i$)}
  %\label{fig:sub2}
\end{subfigure}
\begin{subfigure}{.45\textwidth}
\centering
  \includegraphics[height=\textwidth, angle=90]{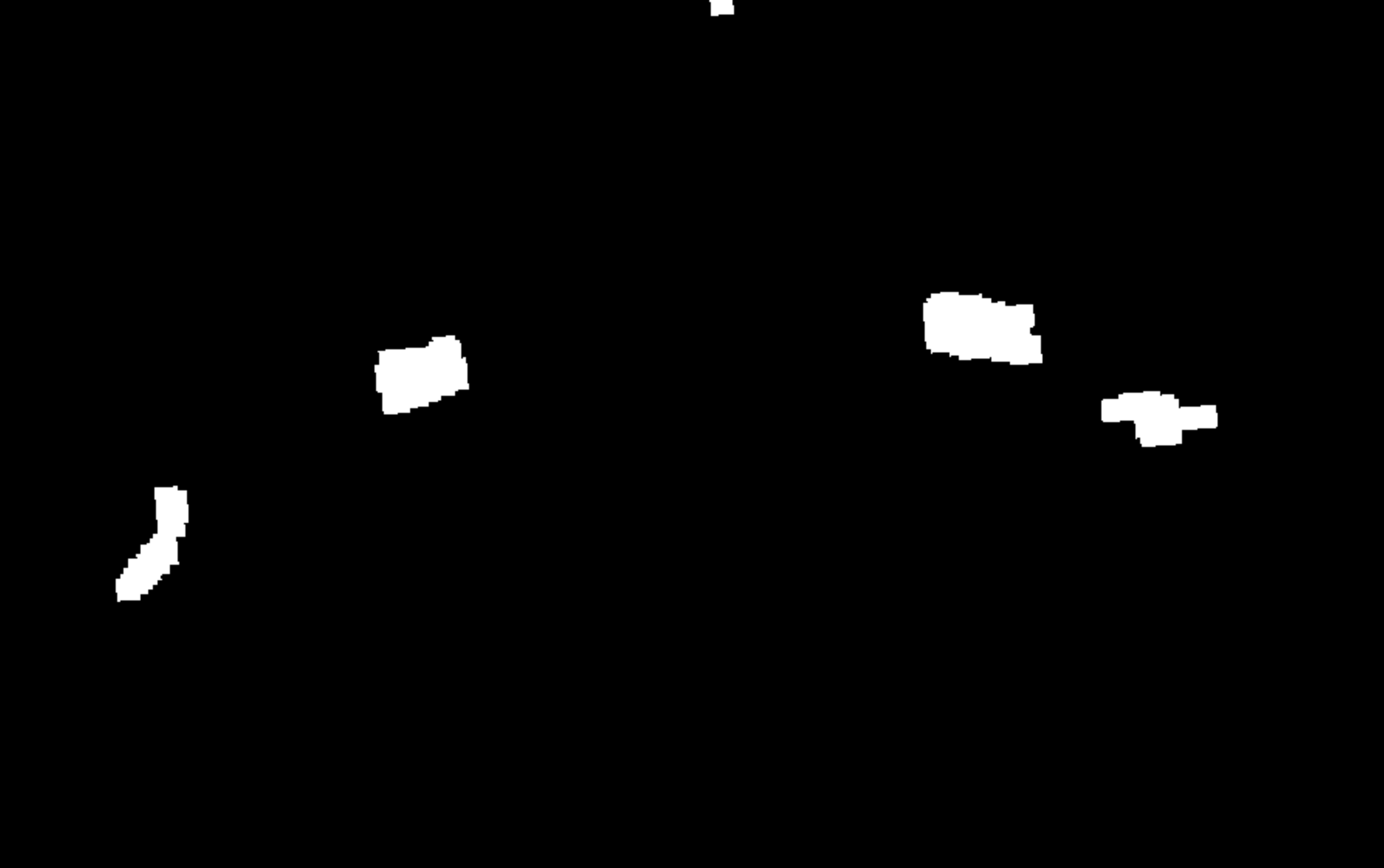}
  \caption{Maximum filtered ObsGrid($t_i$)}
  %\label{fig:sub2}
\end{subfigure}
\begin{subfigure}{.45\textwidth}
\centering
  \includegraphics[height=\textwidth, angle=90]{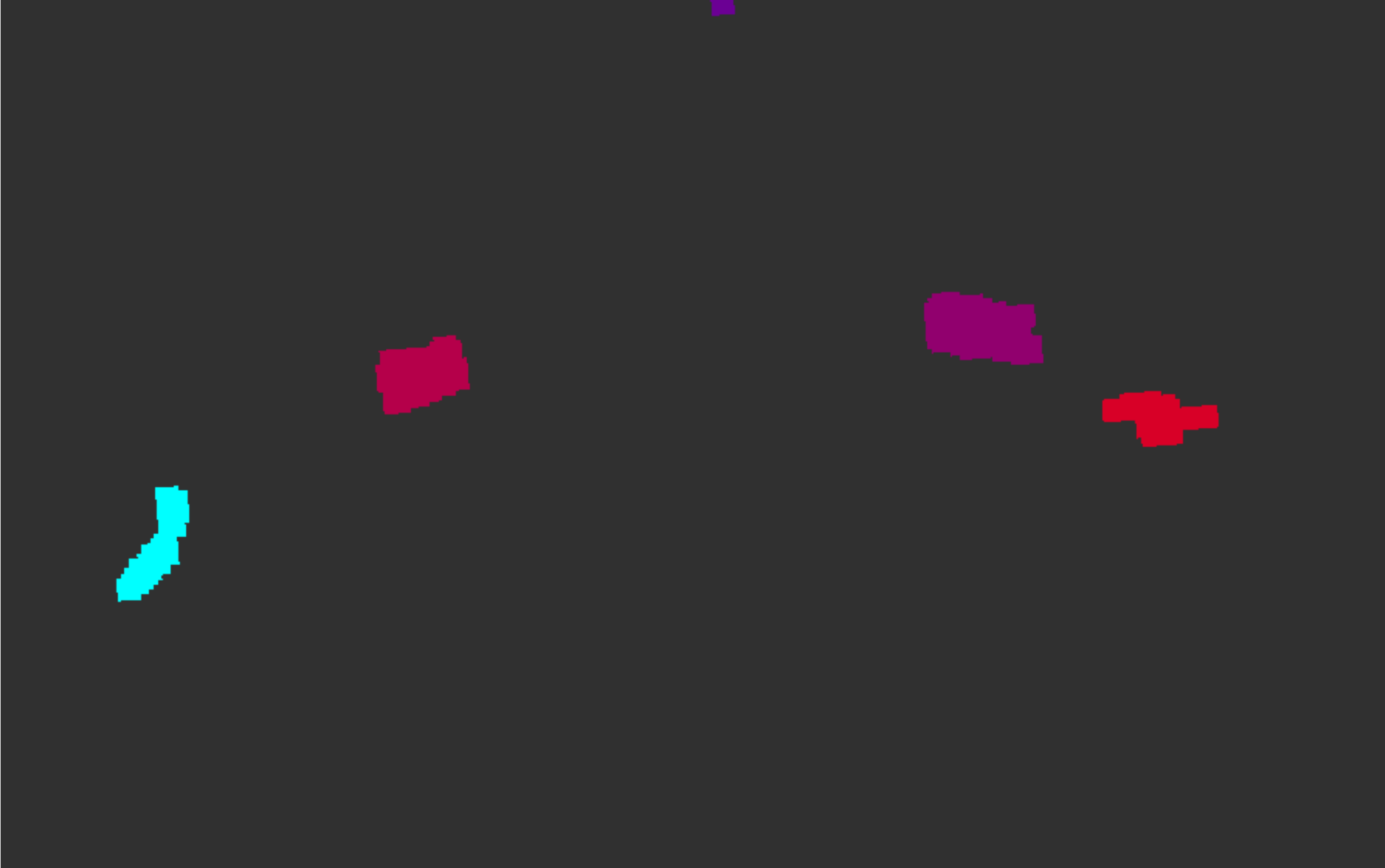}
  \caption{ClusterMap($t_i$)}
  %\label{fig:sub2}
\end{subfigure}
\caption{Grids used for clustering and road object detection. The evidential mass values were generated from the weights obtained after the training, without any post-processing.}
\label{gridsroadobj}
\end{figure*}

\paragraph{}
ObsMap($t_i$) can be used to generate a list of detected road obstacles. First of all, a 5$\times$5 maximum filter is applied to ObsMap($t_i$), to inflate the detected objects. This pessimistic behavior is justified by the need of taking into account the fact that the LIDAR points at the edges of those objects, when having been projected into the grid cells, might have been projected into cells where road points were also present. The $\alpha$ function might then be affected, an return an under-confident discounting factor. This maximum filtering is also used to connect grid cells that belong to the same physical obstacle, which might not be the case because of the sparsity of the LIDAR scans. Finally, ObsGrid($t_i$) is converted into a grid of cluster ids, noted ClusterMap($t_i$), by connected component labelling, with an 8-connectivity. In each cell of ClusterMap($t_i$) is indicated the id of the cluster to which the cell belongs, or 0 if the cell does not correspond to a clustered object. This ClusterMap($t_i$) can be seen as a list of localized road obstacles. Afterwards, the cells of ScanGrid($t_i$) for which a cluster id has been returned are also reinitialized to a fully unknown state: $m_{t_{i}}(\{R\}) = 0$, $m_{t_{i}}(\{\neg R\}) = 0$, $m_{t_{i}}(\{R, \neg R\}) = 1$. Each grid used in this step is presented in Figure~\ref{gridsroadobj}.

\subsection{Road accumulation and ego-motion compensation}
As potential road objects and displaced objects have been removed, ScanGrid($t_{i}$) and RoadGrid($t_{i-1}$) can trivially be fused, by simply using the Dempster-Shafer operator on $m_{t_{i}}$ and $m_{t_{i-1}}$ for each grid cell. The resulting RoadGrid($t_{i}$) is then available for a navigation system, and can be fused with new incoming LIDAR scans. However, when a new ScanGrid will be generated, the displacement of the vehicle over time will have to be considered, before fusing it with a RoadGrid. An odometry model is thus needed to reproject the RoadGrid. A CAN odometry model can be used to track the movement of the acquisition platform, and reproject the RoadGrid when a new ScanGrid will be available. Cells of the RoadGrid that are not projected into the area covered by the new ScanGrid are dropped. New cells that cover previously unobserved areas are initialized to a fully unknown state, with a mas value of 1 for $\{R,\neg R\}$.

\subsection{Implementation and example of output}
\begin{figure*}
\begin{@twocolumnfalse}
\centering
\begin{subfigure}{.3\linewidth}
\centering
  \includegraphics[width=0.8\linewidth]{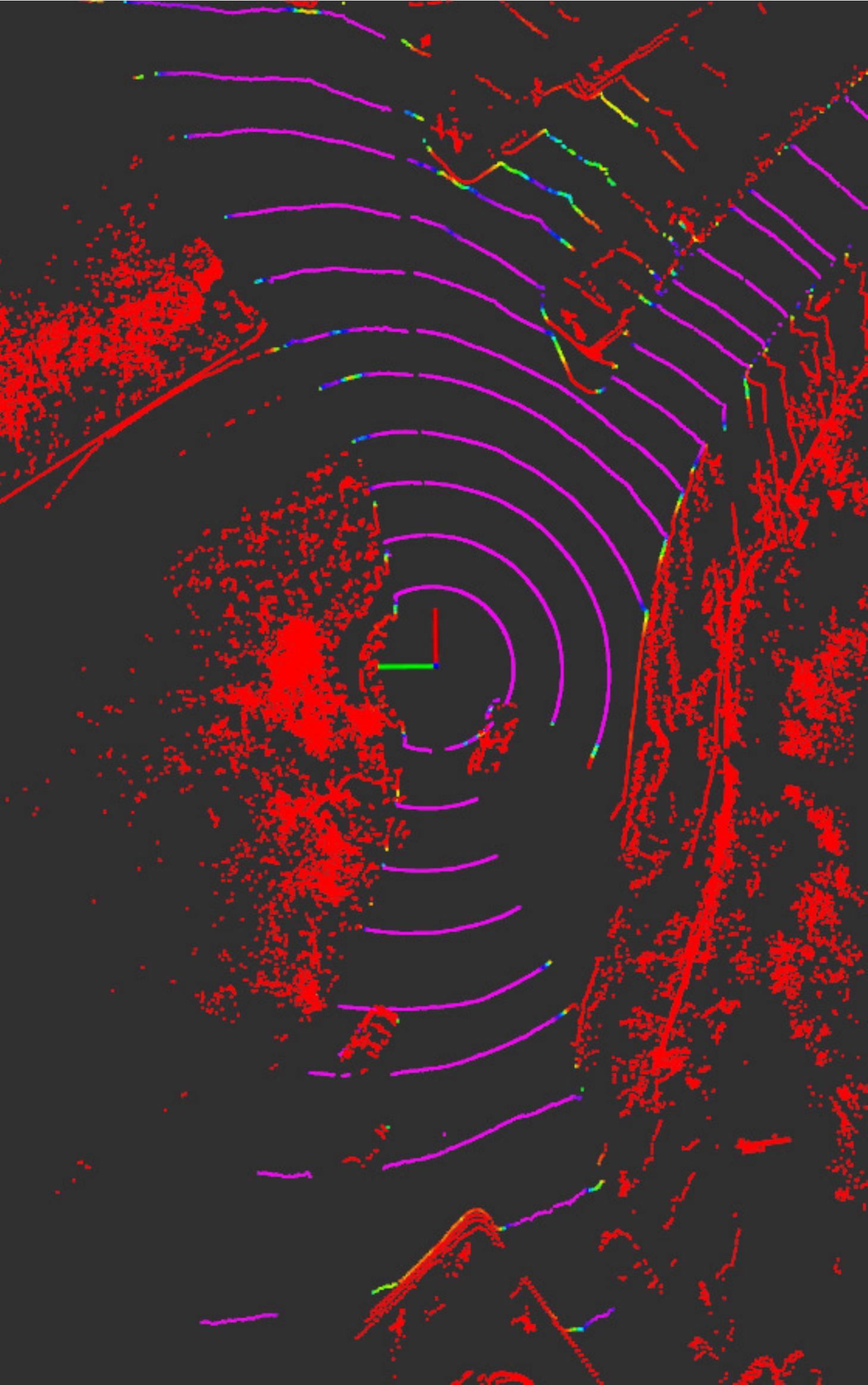}
  \caption{$m(\{R\})$ for each point}
  %\label{fig:sub2}
\end{subfigure}
\begin{subfigure}{.3\linewidth}
\centering
  \includegraphics[width=0.8\linewidth]{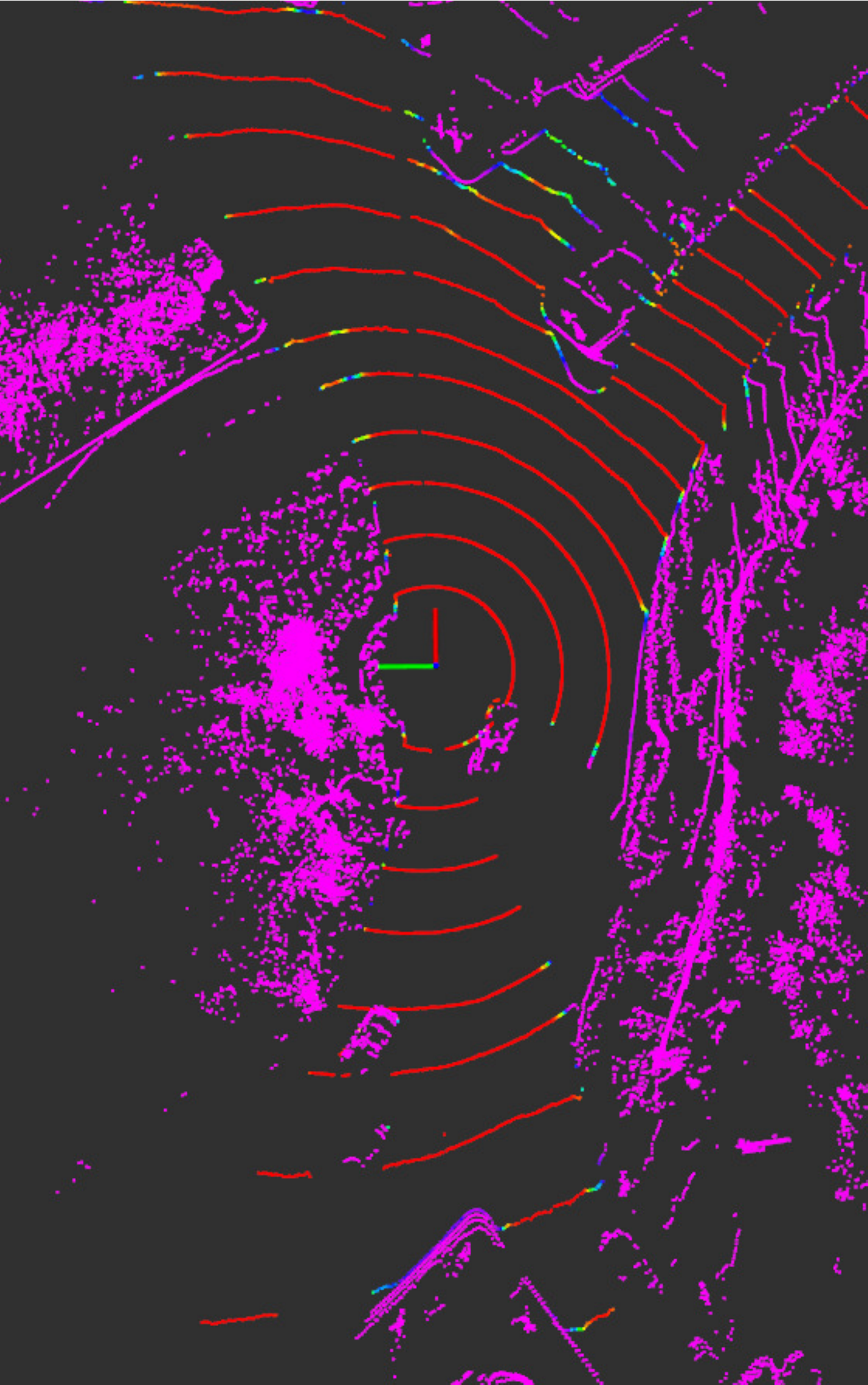}
  \caption{$m(\{\neg R\})$ for each point}
  %\label{fig:sub2}
\end{subfigure}
\begin{subfigure}{.3\linewidth}
\centering
  \includegraphics[width=0.8\linewidth]{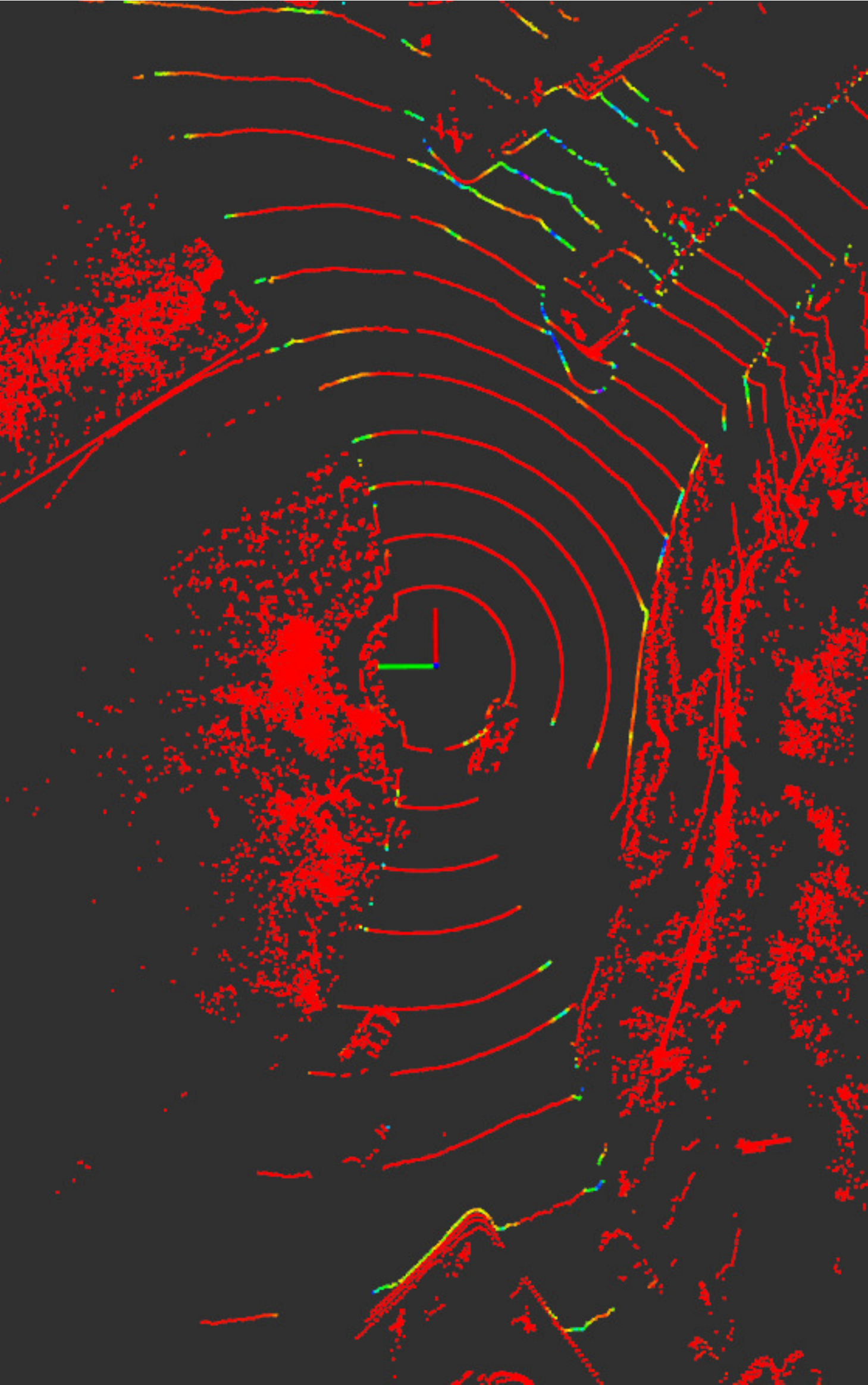}
  \caption{$m(\{R, \neg R\})$ for each point}
  %\label{fig:sub2}
\end{subfigure}
\begin{subfigure}{.3\linewidth}
\centering
  \includegraphics[width=0.8\linewidth]{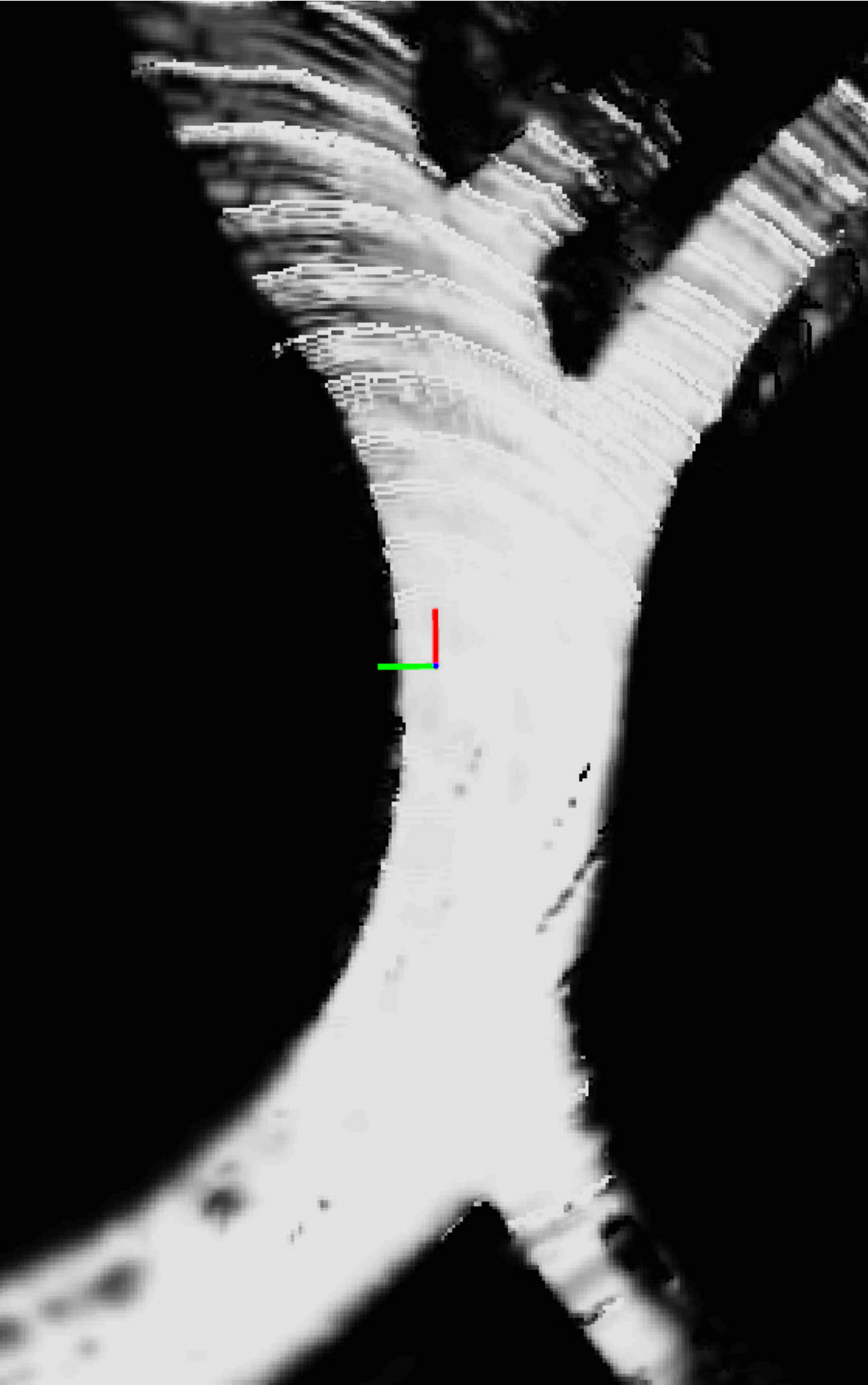}
  \caption{$m(\{R\})$ in RoadGrid($t_i$)}
  %\label{fig:sub2}
\end{subfigure}
\begin{subfigure}{.3\linewidth}
\centering
  \includegraphics[width=0.8\linewidth]{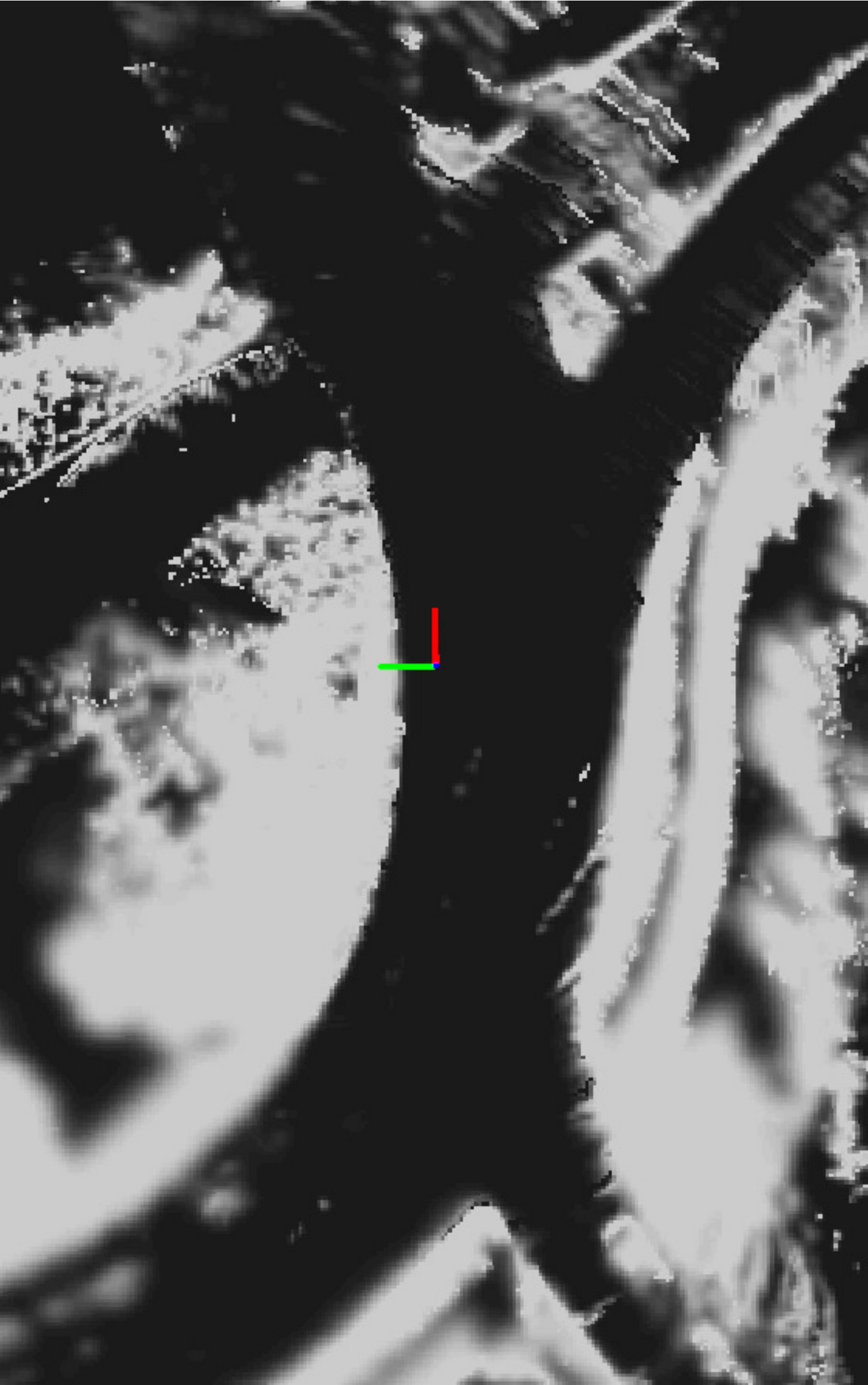}
  \caption{$m(\{\neg R\})$ in RoadGrid($t_i$)}
  %\label{fig:sub2}
\end{subfigure}
\begin{subfigure}{.3\linewidth}
\centering
  \includegraphics[width=0.8\linewidth]{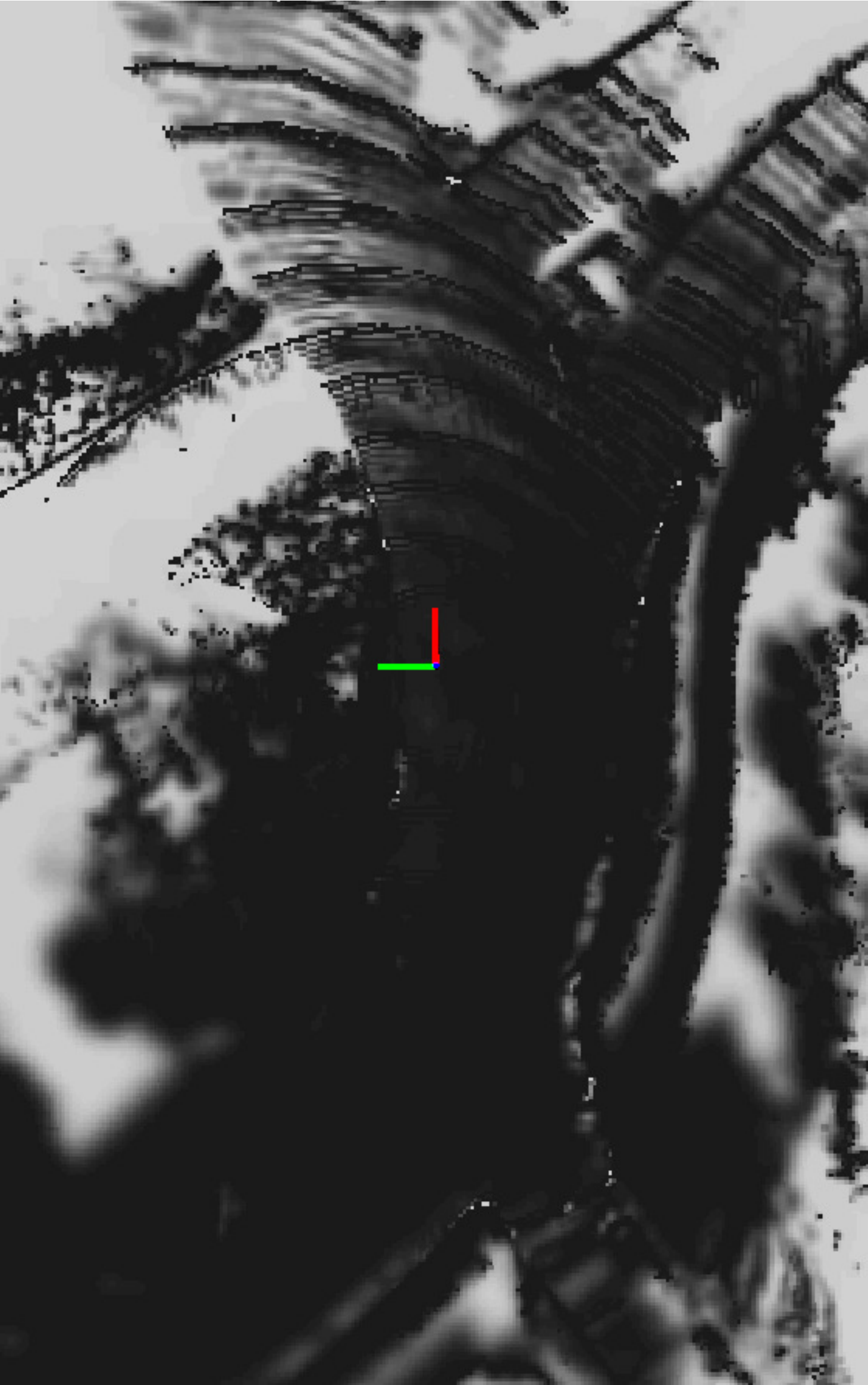}
  \caption{$m(\{R, \neg R\})$ in RoadGrid($t_i$)}
  %\label{fig:sub2}
\end{subfigure}
\begin{subfigure}{.3\linewidth}
\centering
  \includegraphics[width=0.8\linewidth]{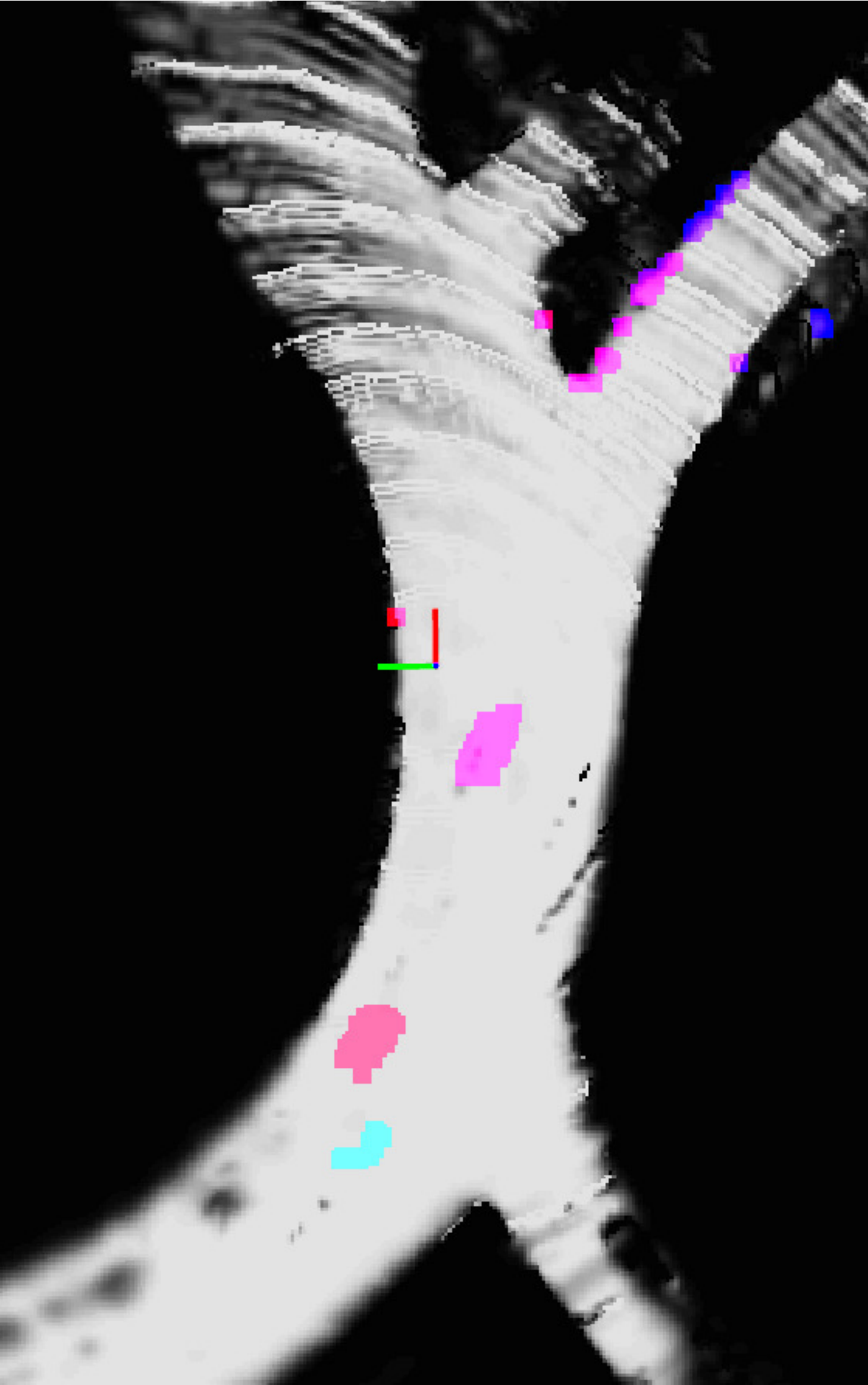}
  \caption{ClusterMap($t_i$) and (d)}
  %\label{fig:sub2}
\end{subfigure}
\caption{Outputs from the road mapping and object detection algorithm. The evidential mass values were generated from the weights obtained after the training, without any post-processing.}
\label{outexfull}
\end{@twocolumnfalse}
\end{figure*}
\paragraph{}
\begin{figure*}[h]
\centering
\includegraphics[width=\linewidth]{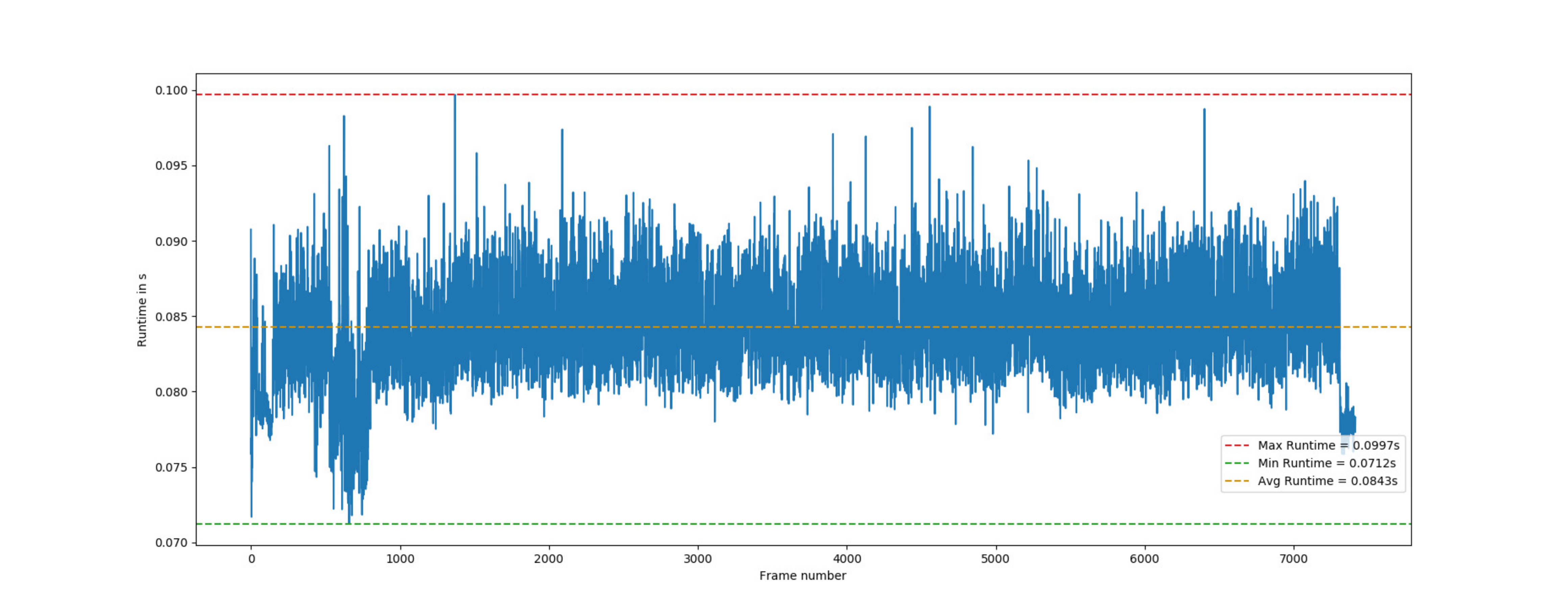}
\caption{Runtime of the road mapping and object detection algorithm relying on the fused RoadSeg networks}
\label{runtime}
\end{figure*}
The algorithm was implemented as a Python ROS node. The inference and evidential fusion of the neural networks is done via the PyTorch framework, and the operations on the grid are performed thanks to the Numpy, OpenCV and Scipy libraries. The TitanX GPU that was used for the training in reused for the inference of the neural networks, but all the grid operations are done on an Intel i7-6700K octacore CPU. An $80m\times50m$ grid is computed around the vehicle, with a cell size of $0.2m$. Only points that have a Z coordinate in the [-2.5,0] range are considered. The odometry is evaluated from an Extended Kalman filter relying on a classical Constant Turn Rate and Velocity (CTRV) model. The CAN network provides the system with speed an heading direction measurements at 10 Hz, and a yaw rate measurement at 100Hz. The CTRV model normally also fuses position measurements obtained from a GNSS sensor, but we chose to rely on a pure CAN odometry, so as to be agnostic to the localization system that is in use. The $\nu$ and $\xi$ values in the $\alpha$ function are empirically set to 4 and 1,5. Similarly to what was done for the training, validation and test sets, the LIDAR scans are obtained from a VLP32C running at 10Hz. We report in Figure~\ref{runtime} the temporal behavior of the algorithm over a 12-minute recording session in Guyancourt. The measured runtimes cover the unpacking of the LIDAR scans, the inference of the neural networks, their fusion, and all the steps of the grid-level mapping and detection algorithm. Our current implementation manages to match, on this recording session, the publication rate of the LIDAR, as the run time is always below 100ms. Yet, the processing time is sometimes very close to 100ms, due to significant jitter. We thus cannot guarantee that the LIDAR scans will always be processed at 10Hz with the current implementation. However, the fact that most of the current implementation relies on standard functions, without extensive use of the GPU, indicates that the performances will be improved by using a dedicated, pure GPU implementation of the functions used in this road mapping and object detection algorithm. We report an additional example of the outputs that are available from the algorithm, in Figure~\ref{outexfull}. This example highlights one limitation with using conflict analysis for object detection: false positives tend to happen at road edges. This can be explained by the fact that road edges are ambiguous by nature, especially because the system was trained on coarse labels. Errors while estimating the odometry from CAN readings, due to sensor noise, can also lead to false positives. A video depicting the whole sequence is available online.\footnote{\url{https://datasets.hds.utc.fr/tmp/automatic-and-manual-lidar-road-labels/road-mapping-demo/}}
\raggedbottom
\begin{figure*}[h!]
\centering
\includegraphics[width=0.8\textwidth]{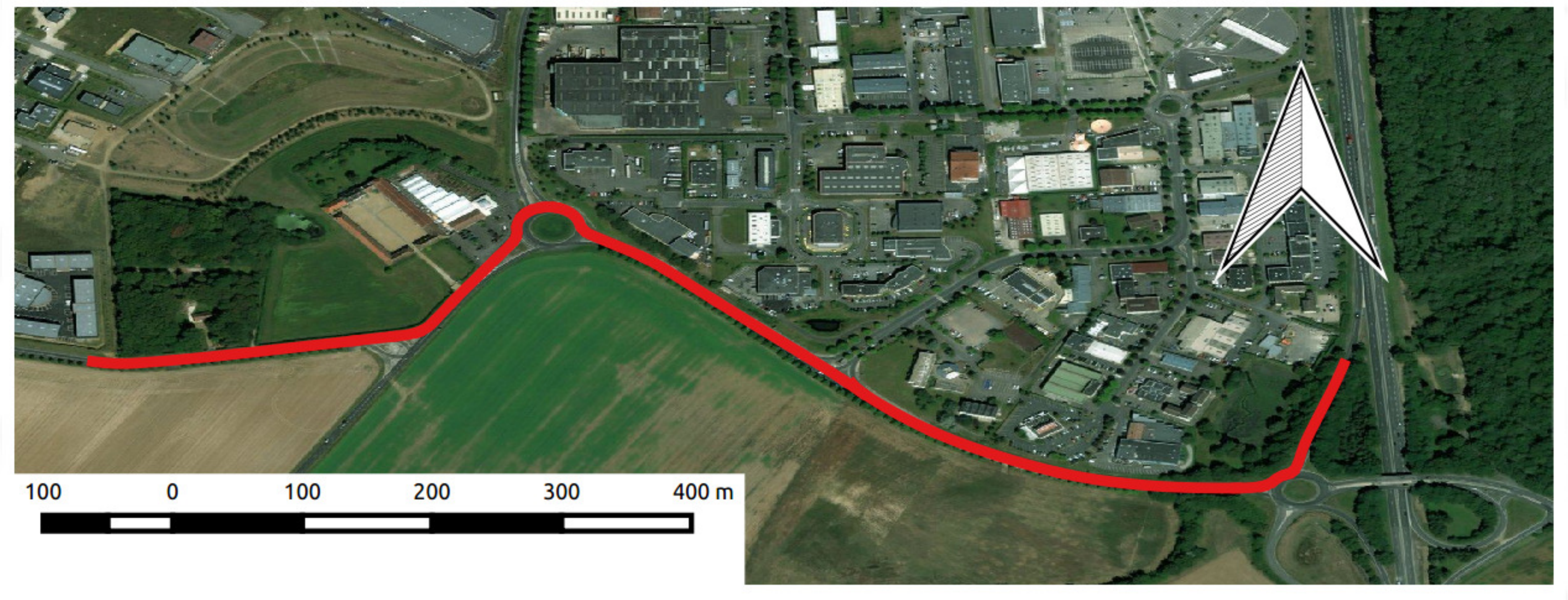}
\caption{Area of evaluation of the road mapping algorithm. The red line indicates the path of the vehicle over the driving sequence used for evaluation. The arrow indicates the north direction.}
\label{grid_gt_area}
\end{figure*}
\begin{figure}[h]
\centering
\includegraphics[width=0.38\textwidth]{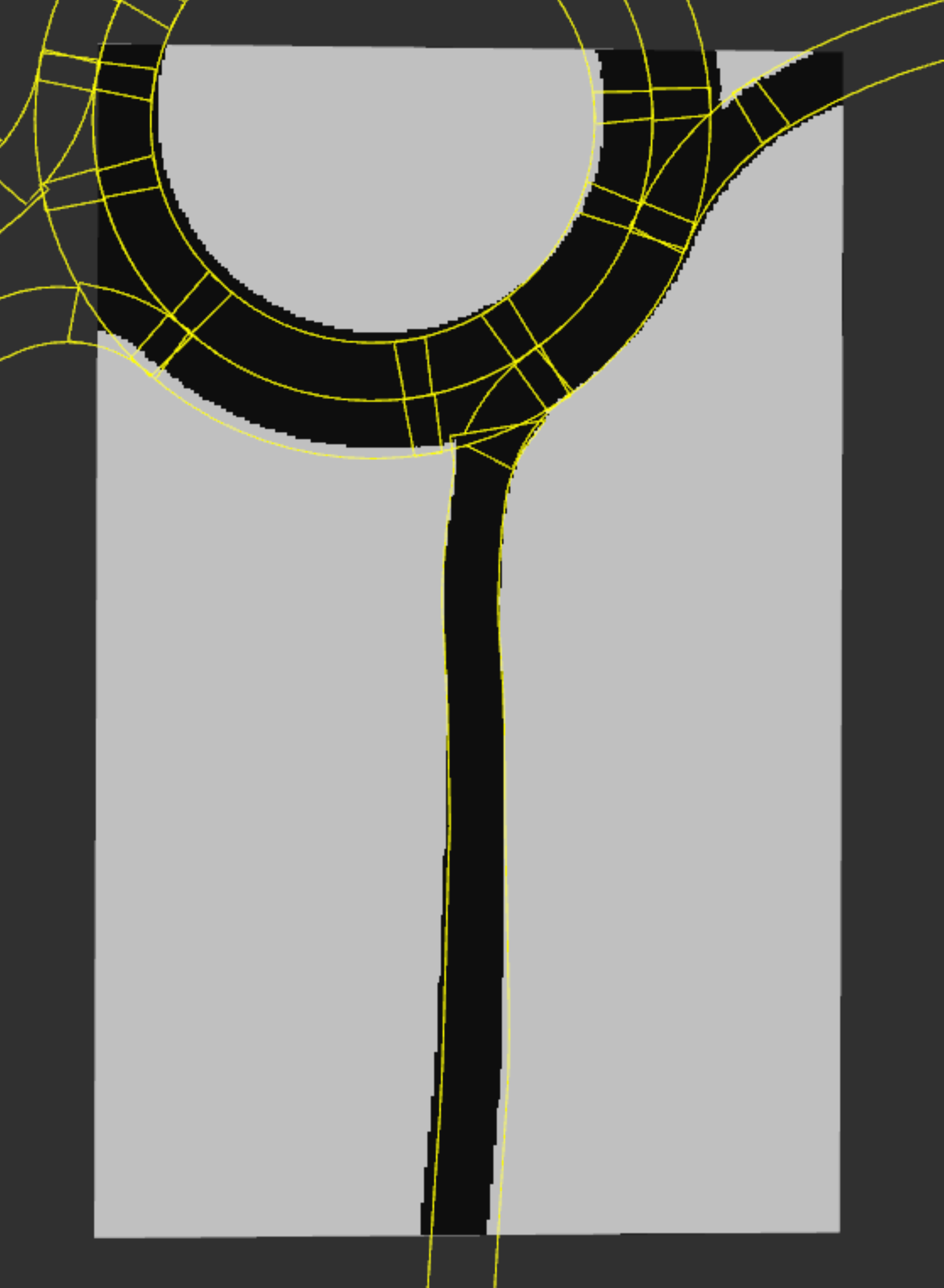}
\caption{Generation of ground-truth grid from HD Maps. The yellow lines represent the map skeleton ; black cells are considered as belonging to the road, and grey ones as not belonging to the road}
\label{gridfrommap}
\end{figure}

\subsection{Evaluation of the quality of the grids}

\paragraph{}
By using the HD Maps from which the training and validation sets were labelled, a ground truth to evaluate the quality of the road grids can be obtained. An empty grid around the vehicle, which follows the same dimensions as the grids obtained from our algorithm ($80m\times50m$, with a cell size of $0.2m$) is first created. The center of each cell is then projected into the corresponding map of the environment. Then, if the center of a cell is projected into a mapped road, the cell is considered as fully belonging to the road ; if not, it is supposed not to belong to the road. We assume a perfect localization system ; otherwise, the resulting grids would be ambiguous, and could not be considered as a ground truth anymore. Figure~\ref{gridfrommap} shows an example of such a ground-truth grid. The grid mapping algorithm can then be evaluated over a driving sequence, provided that the localization remains accurate enough, and that all the roads are properly and unambiguously mapped. The Guyancourt area, where the test set was recorded, was thus not suitable, because of the reserved bus lanes that are present over the area. We chose to evaluate the road mapping algorithm over a driving sequence recorded in the Rambouillet area. In order to make this evaluation fair, the recording was done on roads that are perfectly mapped, and were not part of the training dataset. We selected a peri-urban section where the roads were bordered by fields and small buildings, so as ensure an accurate GNSS positioning, and reliable RTK corrections, over the whole sequence. The driving area is depicted in Figure~\ref{grid_gt_area}. The LIDAR grids are generated from the same LIDAR and vehicle than the test set. The vehicle was driving on open roads, and traffic was thus present during the recording.

\begin{figure*}[h!]
\begin{@twocolumnfalse}
\centering
\begin{subfigure}{\textwidth}
\centering
\includegraphics[width=0.8\textwidth]{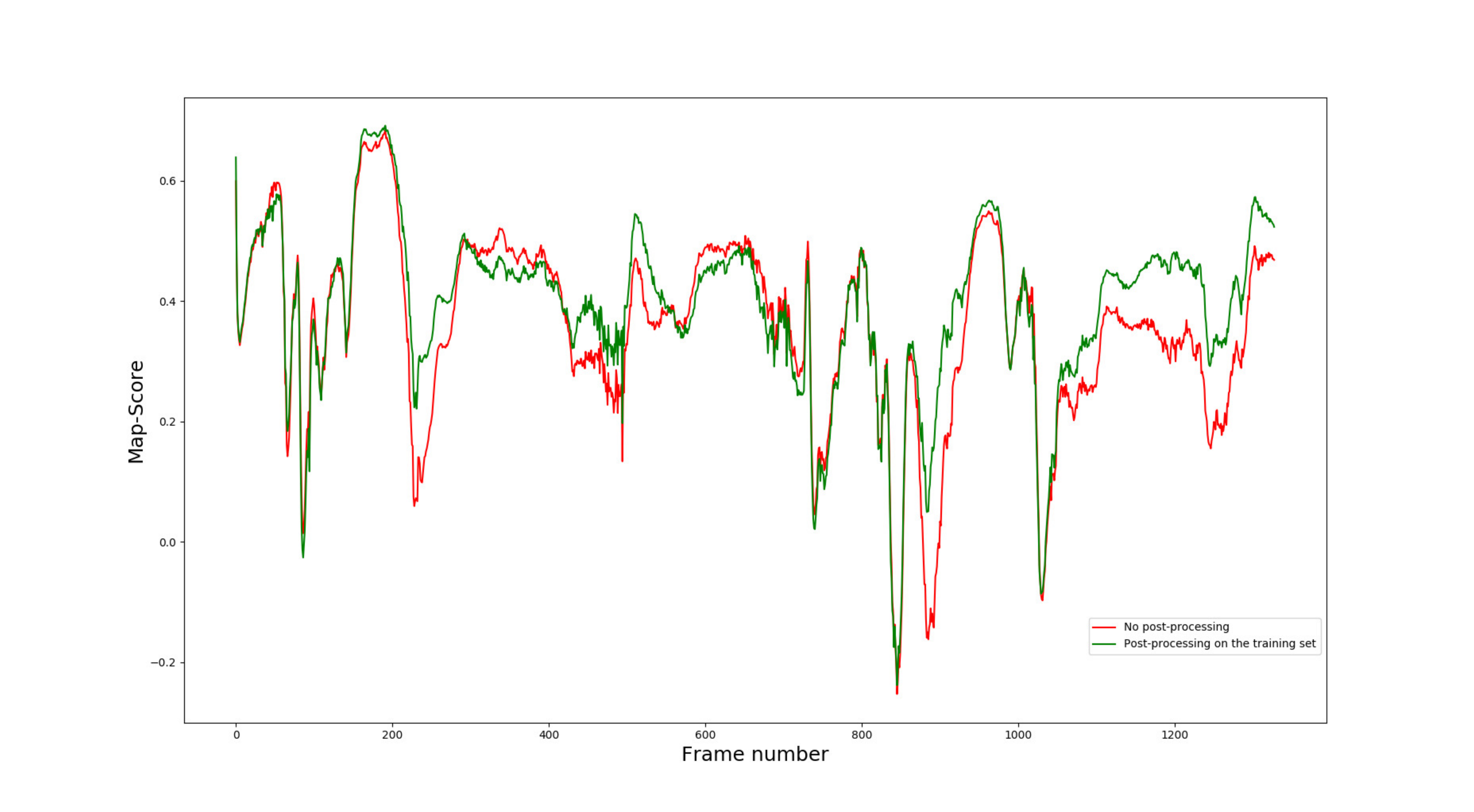}
\end{subfigure}
\begin{subfigure}{\textwidth}
\centering
\includegraphics[width=0.8\textwidth]{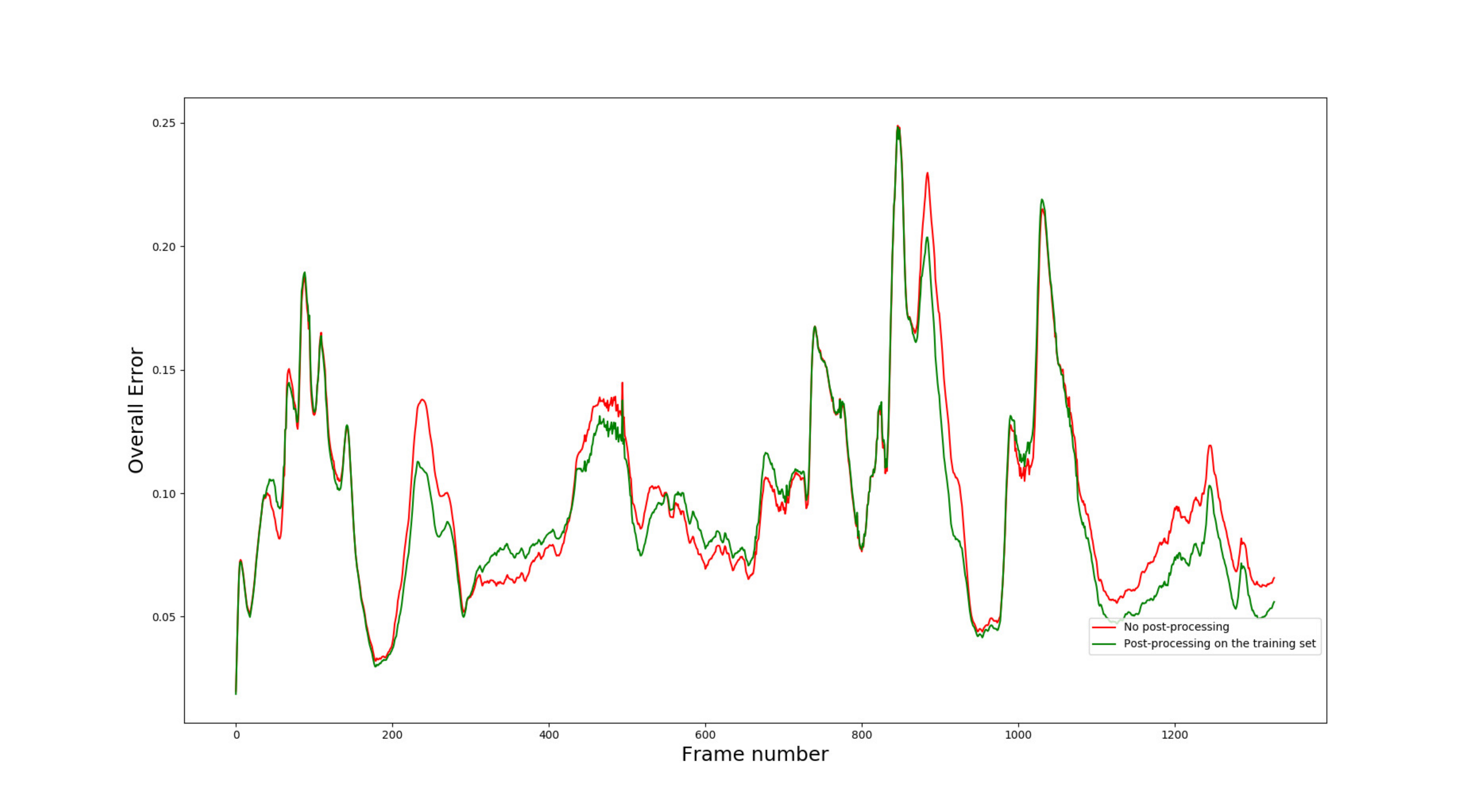}
\end{subfigure}
\begin{subfigure}{\textwidth}
\centering
\includegraphics[width=0.8\textwidth]{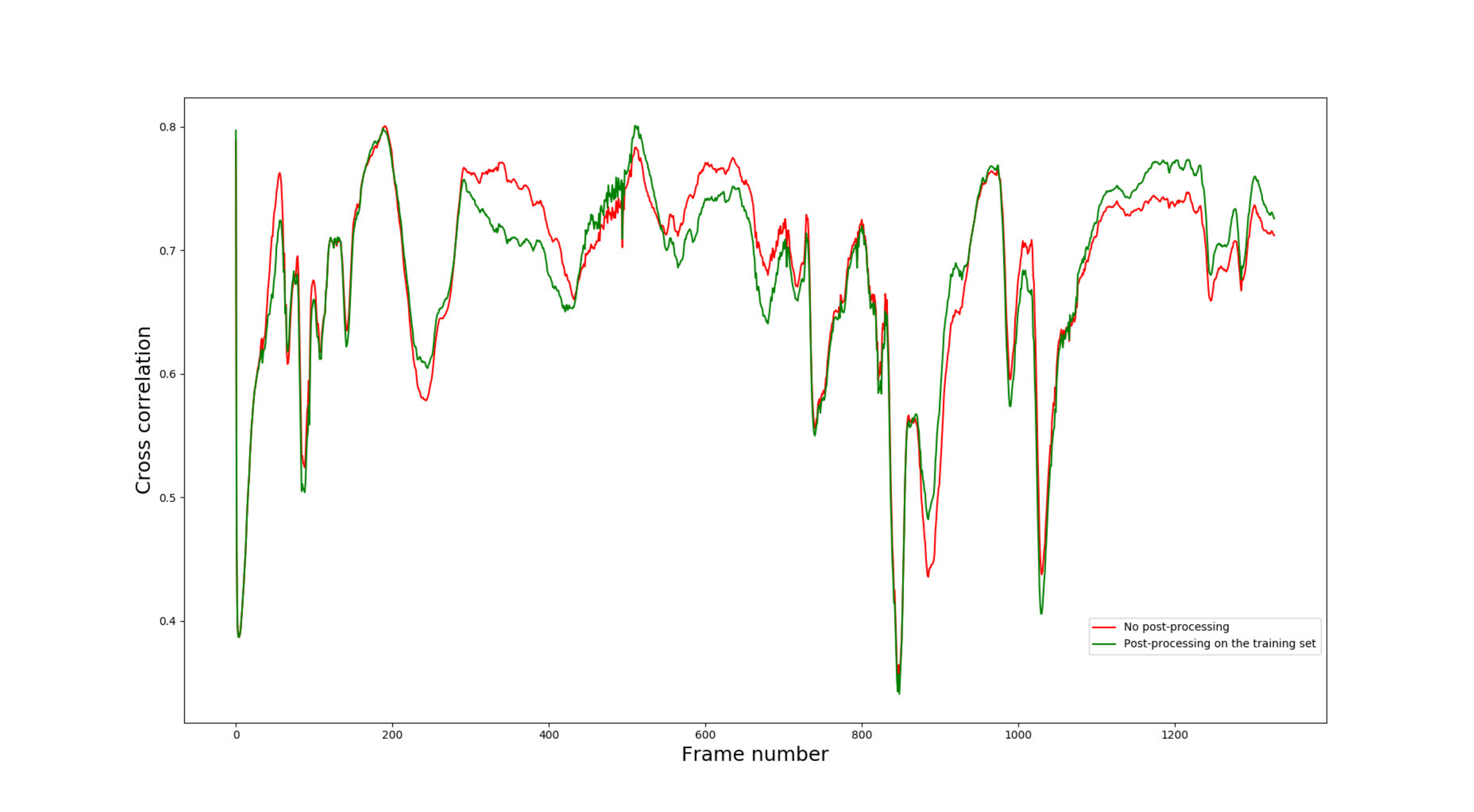}
\end{subfigure}
\caption{Map-Score, Overall Error and Cross correlation of our road grids, over a driving session.}
\label{metrics_grid}
\end{@twocolumnfalse}
\end{figure*}

\paragraph{}
To evaluate the quality of our evidential road grids, we propose to rely on three metrics, that are traditionnally used to compare occupancy grids with simulated ground truth, as in~\cite{balaguer2009evaluating}: the cross correlation coefficient~\cite{o2003empirical}, the Map-Score~\cite{martin1996robot}, and the Overall Error~\cite{carlson2005conflict}. At each update of the road grid, and at each new LIDAR scan, those metrics can be computed with regards to the ground truth grid obtained from the HD map. Those three metrics are complementary, since the cross correlation coefficient globally compares the statistics of the ground truth and estimated grids, the Overall Error estimates the error on the evidential mass values at the cell level, and the Map-Score compares how well probabilities estimated at the cell level match the ground truth. We however adapt those metrics to our system, as they were originally used to assess the performances of fully autonomous robotics systems, that relied on path planning and exploration algorithms. That is the reason why we only compute the Map-Score, the Overall Error, and the cross correlation coefficient from grid cells in which at least one LIDAR point has been projected. The Map-Score and Overall Error are also normalized with regards to the number of grid cells that were used to compute them. Let $i$ be the frame index, $N_c$ the number of cells in the ground truth and estimated road grids, $1_R(j)$ a binary indicator which indicates that the $j^{th}$ grid cell of the ground truth grid belongs to the road, $1_L(j)$ a binary indicator  which indicates that at least one LIDAR point was projected into the $j^{th}$ cell of our road grid, $m_j(R)$ the estimated mass value on $R$ evaluated on the $j^{th}$ cell from our algorithm, and $Pl\textunderscore Pm_j(R)$ the probability that the $j^{th}$ grid cell belongs to the road, computed from the corresponding evidential mass values and the plausibility transformation. We compute a Map-Score as follows:
\begin{equation}
\small
\begin{aligned}
MapScore_i &= \sum_{j=1}^{N_c}\frac{1_L(j)*\left[1+log_2(Pl\textunderscore Pm_j(R)*1_R(j))\right]}{1_L(j)} \\
&+\sum_{j=1}^{N_c}\frac{1_L(j)*\left[(1-Pl\textunderscore Pm_j(R))*(1-1_R(j))\right]}{1_L(j)}
\end{aligned}
\end{equation}
We compute an Overall Error on $m_j(R)$ as follows:
\begin{equation}
Overall\textunderscore Error_i = \frac{\sum_{j=1}^{N_c}1_L(j)*|m_j(R) - 1_R(j)|}{\sum_{j=1}^{N_c}1_L(j)}
\end{equation}
Finally, the cross-correlation coefficient is estimated as follows:
\begin{equation}
\scriptsize
Cross\textunderscore Correlation_i = \frac{\overline{Pl\textunderscore Pm_j(R)*1_R(j)}*\overline{Pl\textunderscore Pm_j(R)}*\overline{1_R(j)}}{\sigma(Pl\textunderscore Pm_j(R))*\sigma(1_R(j))}
\end{equation}
where the mean values and standard deviations are only computed from cells in which at least one LIDAR point has been projected.
\paragraph{}
Those metrics do not directly indicate whether a road grid can be used by a robotic system. They only indicate how well estimated grids match with ground truth grids. However, they can be used to compare several grid mapping algorithms, as approaches that match ground truth grids have high cross correlation coefficients and Map Scores, and low Overall Error rates. We thus propose to compare the evidential road grids obtained from RoadSeg networks using weights that were not post-processed, and grids obtained from RoadSeg networks using post-processed weights. We especially compare the results without post-processing of the weights with the results from weights that were post-processed on the training set, as originally proposed in~\cite{denoeux2019logistic}. Doing so, we evaluate the interest of this post-processing, with regards to the use of RoadSeg weights directly obtained after the training.
\paragraph{} 
Figure~\ref{metrics_grid} depicts the evolution of these metrics over the drving sequence used for evaluation. For the three metrics, both curbs obtained from the post-processed and original weights follow the same evolution. None of the two approaches consistently outperform the other over the whole sequence. The local minima of the curbs depicting the performances of the weights that were not post-processed, in terms of Map-Score and Cross correlation, are significantly lower than their counterpart in the curbs depicting the performances of the post-processed weights. The local maxima however are not necessarily lower for the post-processed weights, which indicates that the post-processed weights have led to more cautious results in instances where the road grids were wrong. For the Overall Error rates, the local maxima reached by the original weights are, often, significantly larger than their counterparts corresponding to the post-processed weights. The fact that the post-processed weights are more cautious, especially where classification errors have happened, is thus clearly visible. However, the original weights still perform relatively well, when compared to their post-processed counterparts. Indeed, in 37.5\% of the frames, the original weights lead to road maps that have strictly higher Map-Scores than the post-processed weights. In 40.7\% of the frames, the original weights lead to lower Overall Error rates than the post-processed weights. Finally, in 53.3\% of the frames, the grids obtained from the original weights even have strictly higher cross correlation coefficients than the grids obtained from the post-processed weights.
\paragraph{} 
The fact that the post-processed weights seem to lead to more cautious grids when analyzed at the cell-level, via Map-Scores and Overall Error rates, while the impact on a global cross correlation coefficient is not as clearly visible, indicates that overall, grids obtained from the post-processed weights and the original weights are very similar. The differences seem to mainly come from individual cells that might happen to be misclassified, and more uncertain when using post-processed weights, without significantly impacting the overall grid. This is compatible with our previous observations on misclassified points, for which the entropy was indeed higher when using post-processed weights, but still close to its counterpart calculated from the original weights. 

\section{Conclusion}
We presented a system that relies on the evidential fusion of three neural networks to detect the road in LIDAR scans. From a training set that was automatically labelled thanks to an HD Map, we achieve performant results on a manually labelled test set. The evidential framework thus seems to be a performant way to fuse neural networks, provided that their respective inputs are independent. We also presented an algorithm that uses this road detection system to map the road surface over time, and cluster road objects. A simple CPU/GPU implementation of this algorithm is able to process LIDAR scans at approximately 10 Hz, which fits the usual publication rate of state-of-the-art LIDAR sensors. Additional training data is likely to lead to even better results, which is easy to obtain from our automatic label generation procedure, provided that accurate maps are available. A refinement of the training procedure, to cope with the label noise in the automatic labels, is also a possible research direction. Moreover, a more accurate compensation mechanism, to handle the movement of the vehicle during the scanning process, would be beneficial. However, such a mechanism would have to be able to cope with sensor noise and timing errors, and to produce uncertainty estimates in the individual coordinates of each LIDAR, point while constructing a coherent scan during the sweeping process. This uncertainty information could even be directly used in a grid mapping algorithm, to further improve the representation of the environment. To the best of our knowledge, such an approach has not been proposed yet. Another interesting research direction, to improve our road mapping algorithm, would be to force the networks to generate even more cautious evidential mass values for misclassified points, as the current post-processing procedure proposed in~\cite{denoeux2019logistic} does not lead to road grids that are significantly better than grids obtained from the directly obtained after the training. Multi-class classification would also be valuable, especially for road object detection, but this would come at the cost of, either, semantically enhanced maps, or intensive manual labellisation. This would however be very useful to detect more objects, as conflicts analysis only allows us to detect objects on the road, but not on the sidewalks for instance. Huge LIDAR datasets for semantic segmentation are emerging~\cite{behley2019semantickitti}, but the burden of manual labellisation is still a reality. Furthermore, LIDAR scans produced from different sensors can be very different, in terms of granularity and resolution. This makes the use of external data way more complex than for image segmentation. Finally, even if we manage to detect the road from a neural network, without relying on a explicit model, we still rely on hyper-parameters for conflict analysis and object detection. We thus produce false positives at the road edges. A neural network, trained to detect objects, could potentially replace this conflict analysis step. Nevertheless, a mechanism to ensure the absence of false negatives would then have to be implemented. We instead believe that knowledge-based approaches could properly handle false positives. For instance, it has been proposed to ensure the consistence of the perception information via rule-based systems, so as to build a coherent World Model from pre-existing knowledge~\cite{albus20024d}. Adequate rules could potentially be used to cope with the false positives, depending on their dimensions and location for instance. For example, flat objects that are localized at road borders could be easily considered as false positives.

\subsubsection*{Acknowledgments}
This work is supported by a CIFRE fellowship from Renault S.A.S, and realized within the SIVALab joint laboratory between Renault S.A.S, and Heudiasyc (UMR 7253 UTC/CNRS). We also benefited from fundings from the Equipex ROBOTEX (ANR-10- EQPX-44-01). We thank Nicolas Caddart who, in the context of an internship at Renault S.A.S, to complete his engineering cursus at the ENSTA school of engineering, participated in the labelling of the test dataset, and in the software developments.

\bibliographystyle{ieeetr}
\bibliography{references}
%\end{multicols}
\end{document}